\documentclass[journal]{IEEEtran}
\pdfoutput=1
\usepackage{amsmath,amsfonts}
\usepackage{algorithmic}
\usepackage{algorithm}
\usepackage{array}
\usepackage[caption=false,font=footnotesize,labelfont=rm,textfont=rm]{subfig}
\usepackage{textcomp}
\usepackage{stfloats}
\usepackage{url}
\usepackage{bm}
\usepackage{verbatim}
\usepackage{graphicx}
\usepackage{cite}
\usepackage{pifont}
\usepackage{tikz}
\usetikzlibrary{spy}
\usepackage{xspace}
\usepackage{booktabs}
\usepackage{multirow}
\usepackage{float}
\usepackage{subfig}
\usepackage{overpic}
\usepackage[pagebackref,breaklinks,colorlinks]{hyperref}
\usepackage[capitalize]{cleveref}
\crefname{section}{Sec.}{Secs.}
\Crefname{section}{Section}{Sections}
\Crefname{table}{Table}{Tables}
\crefname{table}{Tab.}{Tabs.}

\newcommand{\ie}{{\emph{i.e.}}\xspace}

\newcommand{\eg}{{\emph{e.g.}}\xspace}

\newcommand{\etal}{{\emph{et al.}}}

\newcommand{\myname}[0]{SE-VFI} %
\newcommand{\mydataset}[0]{SEID} %
\newcommand{\mynetwork}[0]{SEVFI-Net} %

\begin{document}

\title{Video Frame Interpolation with Stereo Event and Intensity Cameras}

\author{Chao~Ding, Mingyuan~Lin, Haijian~Zhang, Jianzhuang Liu, and Lei~Yu
\thanks{Chao~Ding, Mingyuan~Lin, Haijian~Zhang, and Lei~Yu are with the School of Electronic Information, Wuhan University, Wuhan 430072, China. Email:\{dingchao, linmingyuan, haijian.zhang, ly.wd\}@whu.edu.cn.}
\thanks{Jianzhuang Liu is with the Huawei Noah’s Ark Lab, Shenzhen 518000, China. Email: liu.jianzhuang@huawei.com.}
\thanks{The research was partially supported by the National Natural Science Foundation of China under Grants 62271354 and 61871297.}

\thanks{Corresponding authors: Lei~Yu and Haijian~Zhang.}
}

\markboth{SUBMITTED TO IEEE Transactions on Multimedia}%
{Shell \MakeLowercase{\textit{et al.}}: A Sample Article Using IEEEtran.cls for IEEE Journals}

\maketitle

\begin{abstract}
The stereo event-intensity camera setup is widely applied to leverage the advantages of both event cameras with low latency and intensity cameras that capture accurate brightness and texture information. However, such a setup commonly encounters cross-modality parallax that is difficult to be eliminated solely with stereo rectification especially for real-world scenes with complex motions and varying depths, posing artifacts and distortion for existing Event-based Video Frame Interpolation (E-VFI) approaches.
To tackle this problem, we propose a novel Stereo Event-based VFI (\myname) network (\mynetwork) to generate high-quality intermediate frames and corresponding disparities from misaligned inputs consisting of two consecutive keyframes and event streams emitted between them.
Specifically, we propose a Feature Aggregation Module (FAM) to alleviate the parallax and achieve spatial alignment in the feature domain. We then exploit the fused features accomplishing accurate optical flow and disparity estimation, and achieving better interpolated results through flow-based and synthesis-based ways.
We also build a stereo visual acquisition system composed of an event camera and an RGB-D camera to collect a new Stereo Event-Intensity Dataset (\mydataset) containing diverse scenes with complex motions and varying depths.
Experiments on public real-world stereo datasets, \ie, DSEC and MVSEC, and our \mydataset\ dataset demonstrate that our proposed \mynetwork\ outperforms state-of-the-art methods by a large margin. The code and dataset are available at \href{https://dingchao1214.github.io/web_sevfi/}{https://dingchao1214.github.io/web\_sevfi/}.
\end{abstract}


\section{Introduction}
\IEEEPARstart{S}{tereo} event-intensity camera setup allows us to fully perceive the dynamic contents in the scene and has been widely applied in existing depth estimation and stereo matching algorithms\cite{gu2022self,zuo2021accurate,kim2022real,wang2021stereo}. In this setup, the event camera records continuous motion information with extremely high temporal resolution and low power consumption, while the intensity camera captures precise scene brightness and texture information.
However, this setup suffers from the cross-modality parallax issue, especially in real-world scenes with complex non-linear motions and varying depths~\cite{gehrig2021dsec,zhu2018multivehicle}. As a result, it can significantly degenerate the performance of existing Event-based Video Frame Interpolation (E-VFI) approaches, most of which rely on simulation datasets and require per-pixel spatial alignment between events and frames~\cite{tulyakov2021time,tulyakov2022time,he2022timereplayer,paikin2021efi,gao2022superfast,xiao2022eva}, leading to artifacts and distortions with the stereo event-intensity camera setup as shown in~\cref{fig-1}.
\def\cimwidth{0.185}
\def\zuoxia{(-2.25,1.8)}
\def\youshang{(-0.95,2.8)}

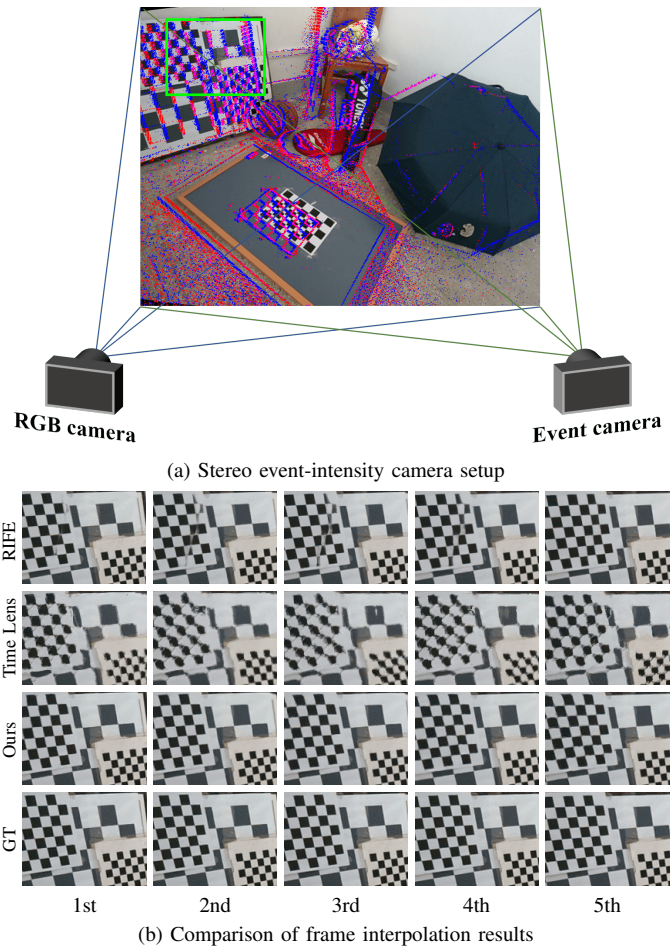
\begin{figure}[t]
\footnotesize
	\centering
 \begin{minipage}[t]{\linewidth}
    		\centering
      \vspace{1mm}
    		\begin{tikzpicture}[spy using outlines={rectangle,green,magnification=\ssmag,size=\ssizz},inner sep=0]
				\node {\includegraphics[width=\linewidth]{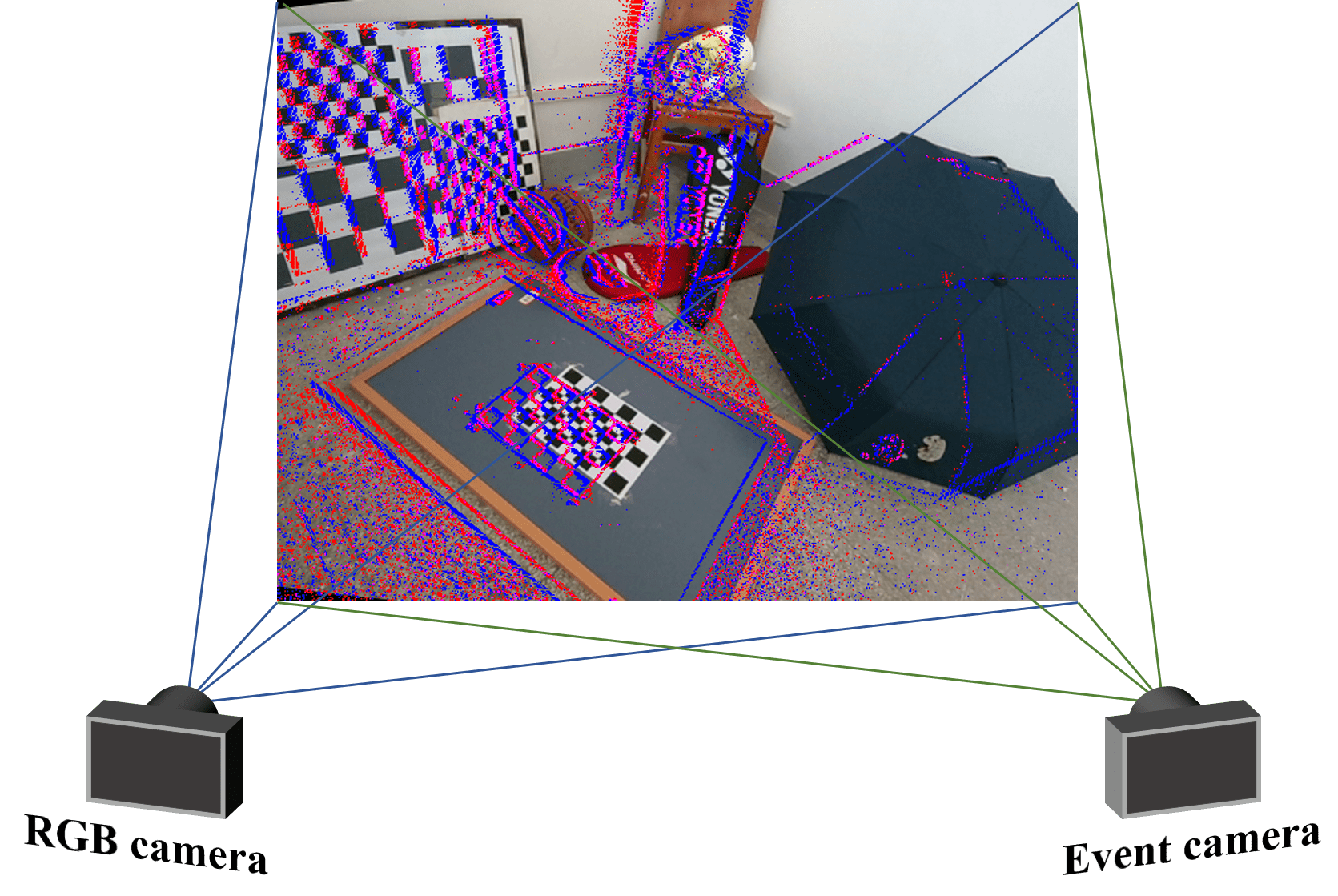}};
				\draw[line width=1pt,green] \zuoxia rectangle \youshang;
				\end{tikzpicture}	
			(a) Stereo event-intensity camera setup\vspace{0.3mm}
    	\end{minipage}
\vspace{-7mm}
\begin{minipage}[]{\linewidth}
            \vspace{-2.5mm}
    		\centering
      \captionsetup[subfloat]{labelsep=none,format=plain,labelformat=empty}
      \begin{tikzpicture}[inner sep=0]
            \node [label={[label distance=0.27cm,text depth=-1ex,rotate=90]right: \textcolor{black}{\scriptsize {RIFE}}}] at (15,15) {};
            \end{tikzpicture}
      \subfloat[]{\includegraphics[width=\cimwidth\textwidth,trim={40 340 440 20},clip]{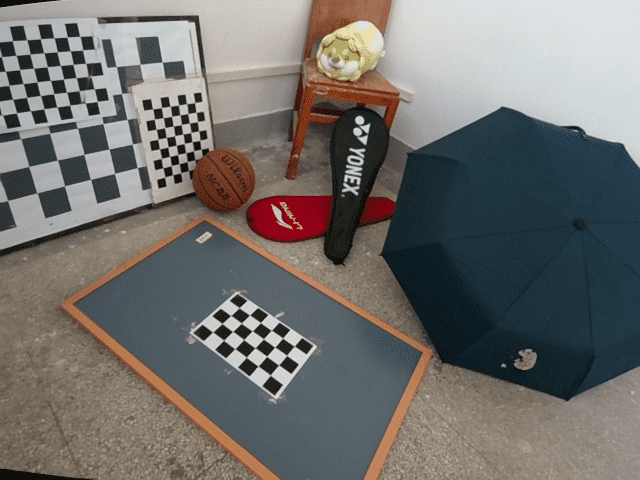}}\hfill
      \subfloat[]{\includegraphics[width=\cimwidth\textwidth,trim={40 340 440 20},clip]{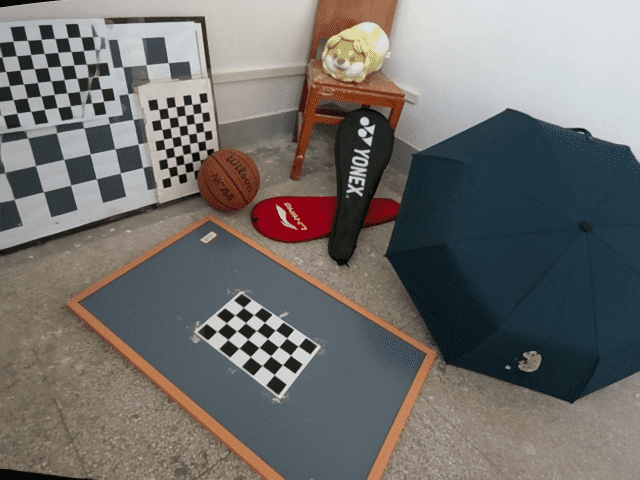}}\hfill
	   \subfloat[]{\includegraphics[width=\cimwidth\textwidth,trim={40 340 440 20},clip]{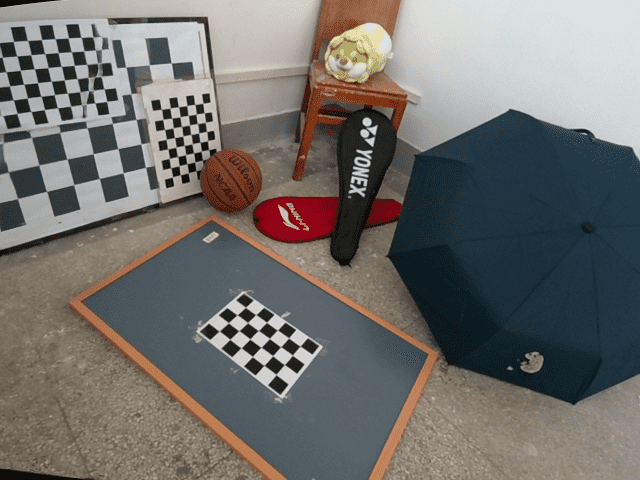}}\hfill
	   \subfloat[]{\includegraphics[width=\cimwidth\textwidth,trim={40 340 440 20},clip]{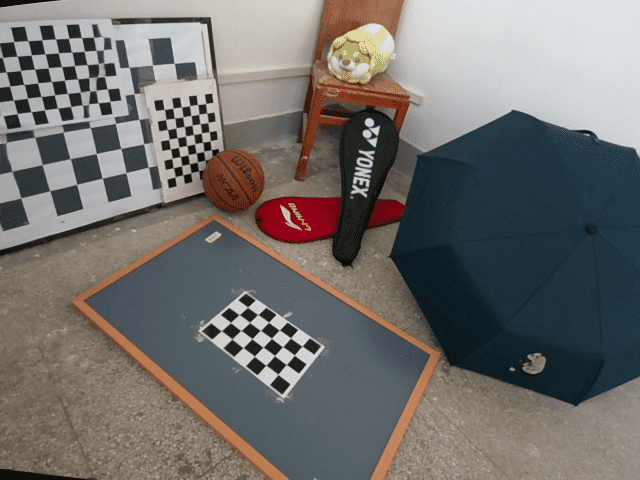}}\hfill
	   \subfloat[]{\includegraphics[width=\cimwidth\textwidth,trim={40 340 440 20},clip]{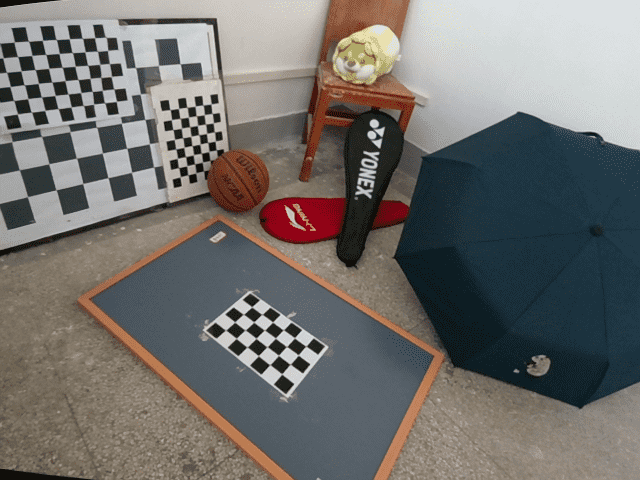}}\hfill
    \end{minipage}
    \vspace{-7mm}
    \begin{minipage}[]{\linewidth}
    		\centering
      \captionsetup[subfloat]{labelsep=none,format=plain,labelformat=empty}
      \begin{tikzpicture}[inner sep=0]
            \node [label={[label distance=0.05cm,text depth=-1ex,rotate=90]right: \textcolor{black}{\scriptsize {Time Lens}}}] at (15,15) {};
            \end{tikzpicture}
      \subfloat[]{\includegraphics[width=\cimwidth\textwidth,trim={40 340 440 20},clip]{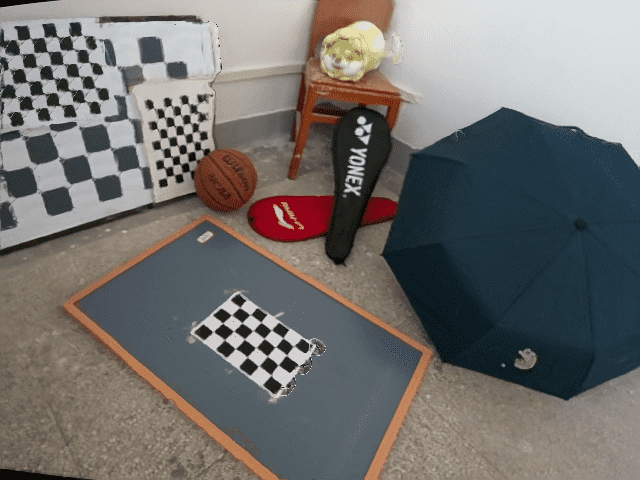}}\hfill
      \subfloat[]{\includegraphics[width=\cimwidth\textwidth,trim={40 340 440 20},clip]{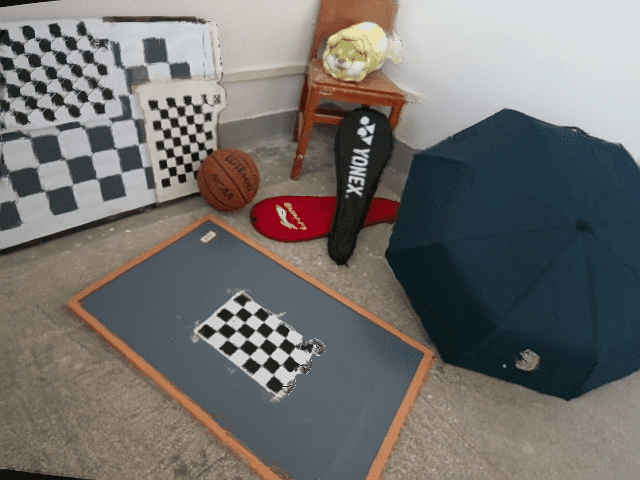}}\hfill
	   \subfloat[]{\includegraphics[width=\cimwidth\textwidth,trim={40 340 440 20},clip]{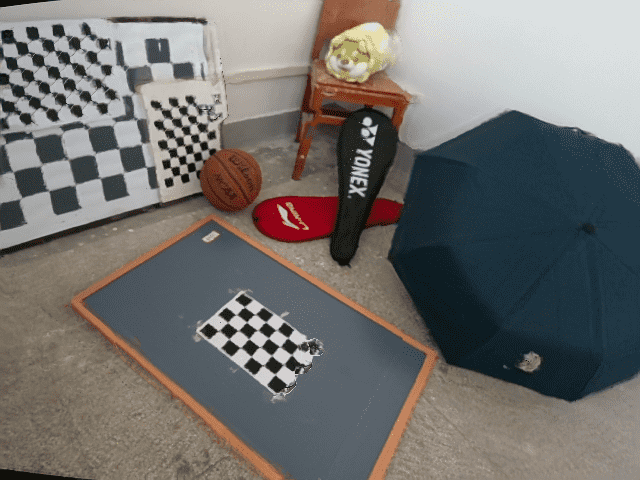}}\hfill
	   \subfloat[]{\includegraphics[width=\cimwidth\textwidth,trim={40 340 440 20},clip]{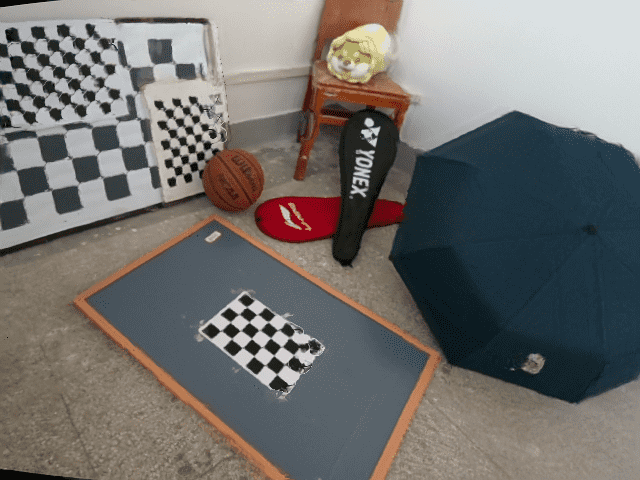}}\hfill
	   \subfloat[]{\includegraphics[width=\cimwidth\textwidth,trim={40 340 440 20},clip]{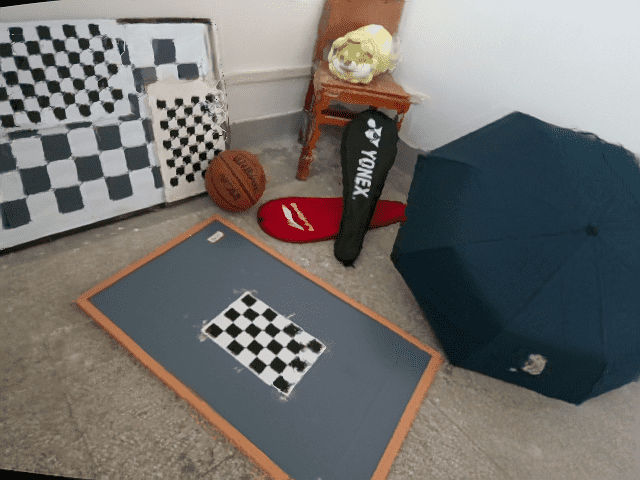}}\hfill
    \end{minipage}
    \vspace{-7mm}
    \begin{minipage}[]{\linewidth}
    		\centering
      \captionsetup[subfloat]{labelsep=none,format=plain,labelformat=empty}
      \begin{tikzpicture}[inner sep=0]
            \node [label={[label distance=0.33cm,text depth=-1ex,rotate=90]right: \textcolor{black}{\scriptsize {Ours}}}] at (15,15) {};
            \end{tikzpicture}
      \subfloat[]{\includegraphics[width=\cimwidth\textwidth,trim={40 340 440 20},clip]{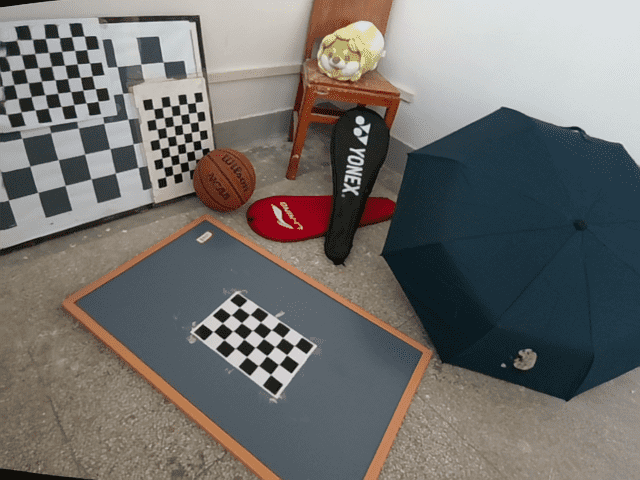}}\hfill
      \subfloat[]{\includegraphics[width=\cimwidth\textwidth,trim={40 340 440 20},clip]{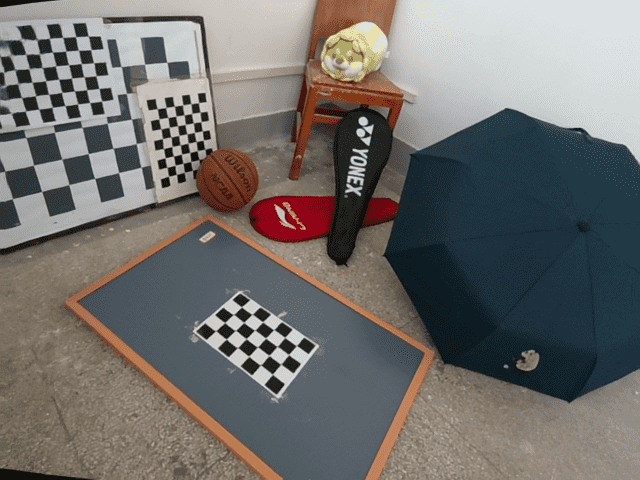}}\hfill
	   \subfloat[]{\includegraphics[width=\cimwidth\textwidth,trim={40 340 440 20},clip]{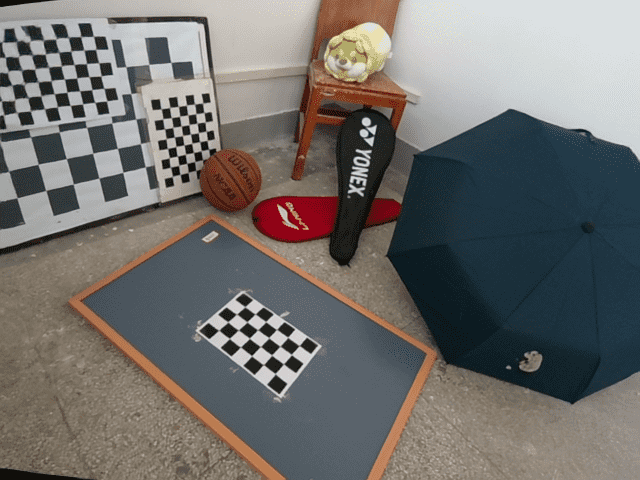}}\hfill
	   \subfloat[]{\includegraphics[width=\cimwidth\textwidth,trim={40 340 440 20},clip]{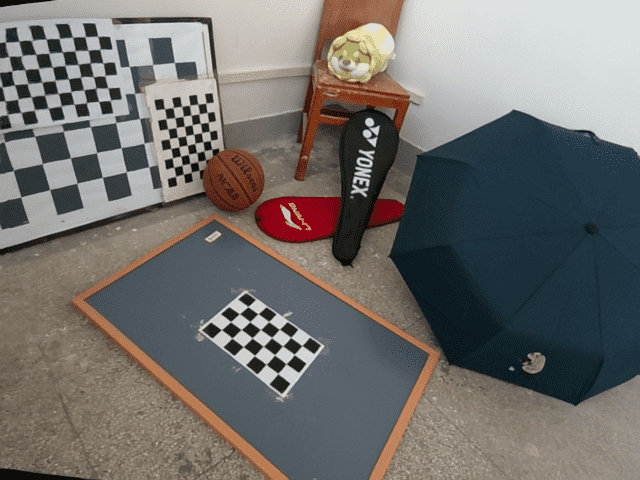}}\hfill
	   \subfloat[]{\includegraphics[width=\cimwidth\textwidth,trim={40 340 440 20},clip]{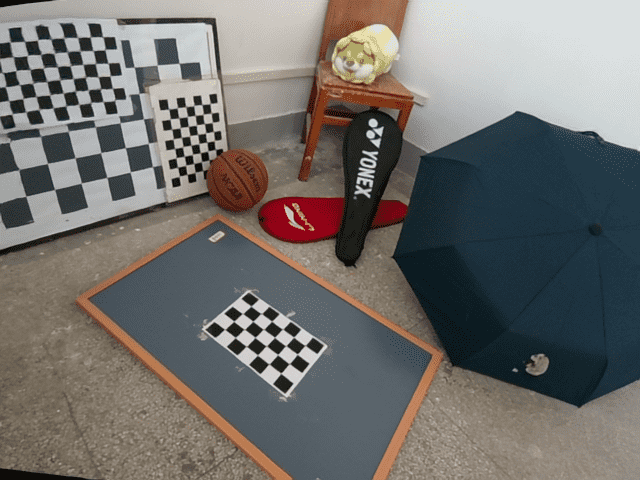}}\hfill
    \end{minipage}
    \vspace{-7mm}\begin{minipage}[]{\linewidth}
    		\centering
      \captionsetup[subfloat]{labelsep=none,format=plain,labelformat=empty}
      \begin{tikzpicture}[inner sep=0]
            \node [label={[label distance=0.42cm,text depth=-1ex,rotate=90]right: \textcolor{black}{\scriptsize {GT}}}] at (15,15) {};
            \end{tikzpicture}
      \subfloat[1st]{\includegraphics[width=\cimwidth\textwidth,trim={40 340 440 20},clip]{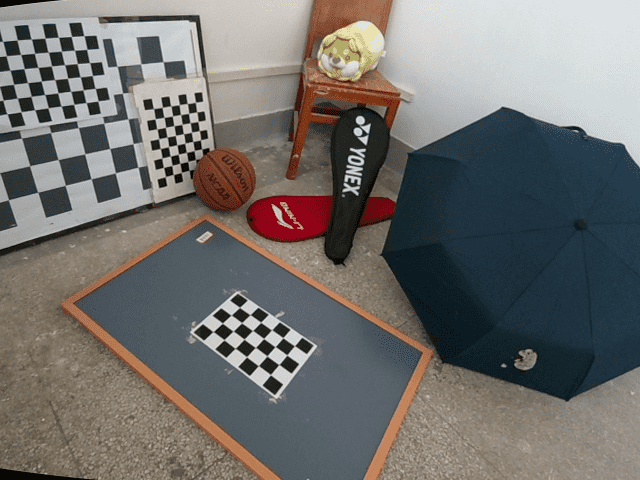}}\hfill
      \subfloat[2nd]{\includegraphics[width=\cimwidth\textwidth,trim={40 340 440 20},clip]{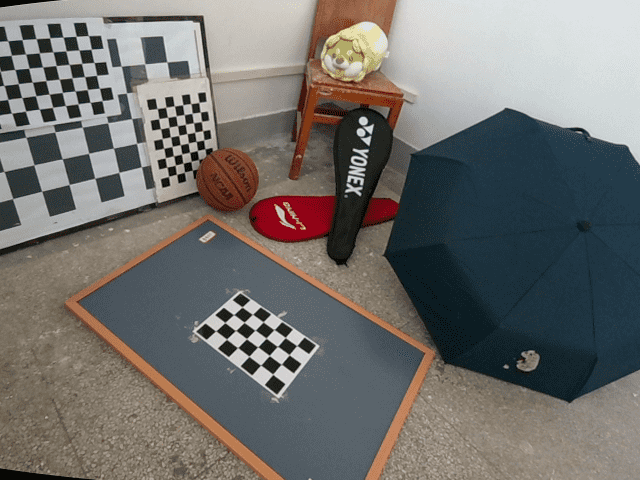}}\hfill
	   \subfloat[3rd]{\includegraphics[width=\cimwidth\textwidth,trim={40 340 440 20},clip]{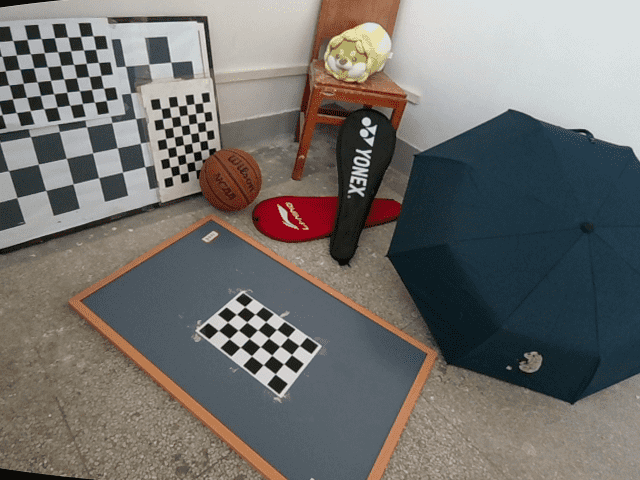}}\hfill
	   \subfloat[4th]{\includegraphics[width=\cimwidth\textwidth,trim={40 340 440 20},clip]{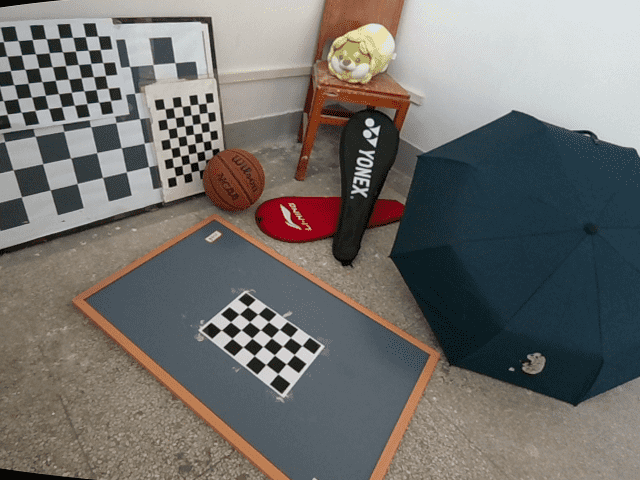}}\hfill
	   \subfloat[5th]{\includegraphics[width=\cimwidth\textwidth,trim={40 340 440 20},clip]{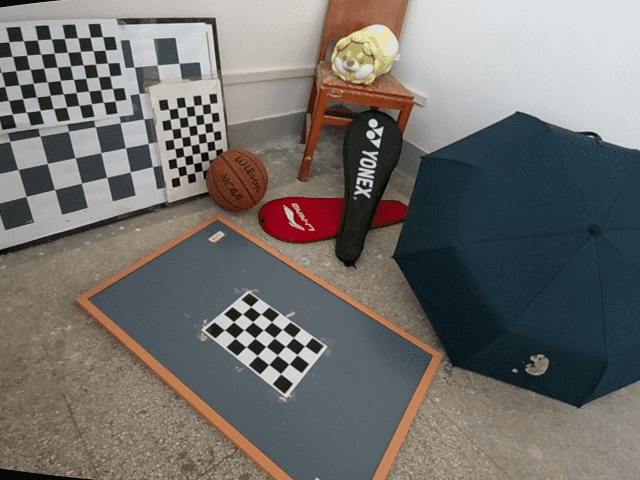}}\hfill
    
        \vspace{1mm}
        (b) Comparison of frame interpolation results\vspace{7mm}
    \end{minipage}

	\caption{Illustrative examples of the impact of the misaligned event and frame data. (a) The events and frames captured by a stereo camera setup share the same field of view but are still misaligned at the pixel level. (b) Comparison of video frame interpolation results of RIFE~\cite{huang2022real}, Time Lens~\cite{tulyakov2021time}, and our proposed \mynetwork. Our method generates the fewest artifacts and achieves the best visualization performance.}
	\label{fig-1}
\end{figure}

Achieving the per-pixel cross-modal alignments is extremely challenging for the Stereo E-VFI (SE-VFI) task due to the missing inter-frame intensities and modality differences. Even though existing works~\cite{tulyakov2021time, gehrig2021dsec} directly apply the global homography and stereo rectification, it is only valid for scenes either with a large depth~\cite{gehrig2021dsec} or within a plane~\cite{tulyakov2021time}. Complex motions in real-world dynamic scenes often lead to temporally varying depths, violating the planary homography and thus bringing cross-modal misalignments. Thus the correspondence between events and frames is necessary to fulfill the SE-VFI, which is however an ill-posed problem since the missing inter-frame intensities.  On the other hand, event cameras work in a completely different mechanism from intensity cameras, only responding to brightness changes and asynchronously emitting binary events. Despite the implicit inclusion of scene structure and texture information in events, they are often triggered at the edges of objects with high-intensity contrast. These modality differences bring another challenge to establishing spatial data correspondence between events and frames.

To handle these problems, we propose a network called \mynetwork\ for the Stereo Event-based Video Frame Interpolation (\myname) task to fully exploit the potential of E-VFI in real-world scenarios. \myname\ aims to generate high-quality intermediate frames and cross-modal disparities using spatially misaligned events and frames.
Our \mynetwork\ consists of four main modules, \ie, \textit{AlignNet}, \textit{SynNet}, \textit{FusionNet}, and \textit{RefineNet}.
Specifically, the AlignNet is built by composing a weighted dual encoder and a Feature Aggregation Module (FAM) to mitigate the modality gap and spatial misalignment between events and frames.
The dual encoder enables us to extract multi-scale features from different modalities separately. To mitigate the spatial misalignment, we employ deformable convolution networks\cite{dai2017deformable} in FAM to estimate the spatial correspondence of different data and obtain aligned features. These features and correspondence enable us to achieve a coarse estimation of motion flows and disparities.
Then, we employ motion flows to warp the boundary frames, resulting in flow-based results. Similarly, we utilize disparities to warp the original events, providing contour constraints for synthesis-based results.
However, it is important to note that these results are only approximations due to the lack of precise inter-frame intensities and the differences in modalities. As a result, the fused result of them may contain overlaps and distortions.
To address these issues, we incorporate a RefineNet to eliminate the defects in the fused results and further optimize the estimations from motion flows and disparities. By doing so, we can effectively handle the challenges posed by modality differences and spatial misalignments.

Additionally, due to the limited diversity of scenes captured by current stereo datasets such as the Stereo Event Camera Dataset for Driving Scenarios (DSEC)~\cite{gehrig2021dsec} and the Multi-Vehicle Stereo Event Camera Dataset (MVSEC)~\cite{zhu2018multivehicle},
they are unable to effectively evaluate the performance of algorithms in various scenes and across a wide range of depth variations.
Thus we build a stereo visual acquisition system containing a SilkyEvCam event camera together with an Intel Realsense D455 RGB-D camera and collect a new Stereo Event-Intensity Dataset (\mydataset). Our \mydataset\ captures dynamic scenes with complex non-linear motions and depths variation that pose a challenge for \myname. It provides events with a higher resolution of $640\times 480$ and high-quality RGB frames at the same resolution. Besides, it also provides depth maps synchronized with RGB frames, which can be used for the stereo-matching task.

The main contributions of this paper are three-fold:
\begin{itemize}
    \item We propose a novel \mynetwork\ framework for the task of E-VFI when frames and events are captured with stereo camera settings in spatially misaligned scenes.
    \item We collect a new stereo event-intensity dataset, containing high-resolution events, high-quality RGB frames, and synchronous depth maps captured in various scenes with complex motions and varying depths.
    \item Extensive experiments show that our \mynetwork\ yields high-quality frames and accurate disparities, and achieves state-of-the-art results on real-world stereo event-intensity datasets.
\end{itemize}

\section{Related Work}
\subsection{Frame-based VFI (F-VFI)}
Video Frame Interpolation (VFI) is a widely investigated low-level task in computer vision, aiming to restore intermediate frames from the neighboring images in the video sequence~\cite{niklaus2017cvpr}. However, it is severely ill-posed and the primary challenge is caused by the missing inter-frame information in terms of motions and textures. To relieve the burden, existing VFI approaches commonly rely on inter-frame motion prediction from neighboring frames~\cite{li2020video}, and thus can be roughly categorized into {\it warping-based} and {\it kernel-based}.

{\it Warping-based} methods~\cite{niklaus2020softmax,Lu_2022_CVPR,Niklaus_2018_CVPR,bao2019depth,zhang2022optical} utilize optical flow that perceives motion information between consecutive frames and captures dense correspondences.
High-quality VFI results often depend on precise motion estimation.
Due to the lack of inter-frame motion information, these methods are typically under the assumption of linear motion and brightness constancy between keyframes.
Some techniques and information have been utilized to enhance the interpolation performance, \eg, forward warping\cite{niklaus2020softmax}, transformer\cite{Lu_2022_CVPR}, context\cite{Niklaus_2018_CVPR}, depth\cite{bao2019depth}, patch-based~\cite{kaviani2015frame,zhang2022optical} and deformable convolution\cite{shi2021video,lei2023flow}.
But due to the assumption of linear motion, these methods encounter challenges when dealing with complex non-linear motions in real-world scenarios. Some approaches design complex high-order motion models, such as cubic\cite{chi2020all,chun2012mri} and quadratic\cite{xu2019quadratic}, to address non-linear motions.
However, they need more neighboring frames as inputs to estimate motion models and still struggle in tackling real-world complex motions.

{ \it Kernel-based} methods\cite{niklaus2017cvpr,niklaus2017iccv,zhang2009spatio} incorporate both motion estimation and frame reconstruction. They utilize the input frames to estimate convolution kernels, which encode local motions across the input keyframes and generate intermediate frames based on these kernels.
Compared to warping-based methods, kernel-based ones are more effective in dealing with occlusion, sudden brightness change, and blurry inputs.
But they need expensive costs for computing per-pixel convolution kernels and are hard to capture large motions due to the fixed and small kernel sizes.

The common problem of Frame-based VFI (F-VFI) methods is the missing information between input frames.
Thus motion models or convolution kernels can only be estimated by consecutive neighboring frames, which limits the types of scenes that can be coped with by them in the real world.

\subsection{Event-based VFI (E-VFI)}
\begin{figure*}[ht]
  \centering
   \includegraphics[width=\textwidth]{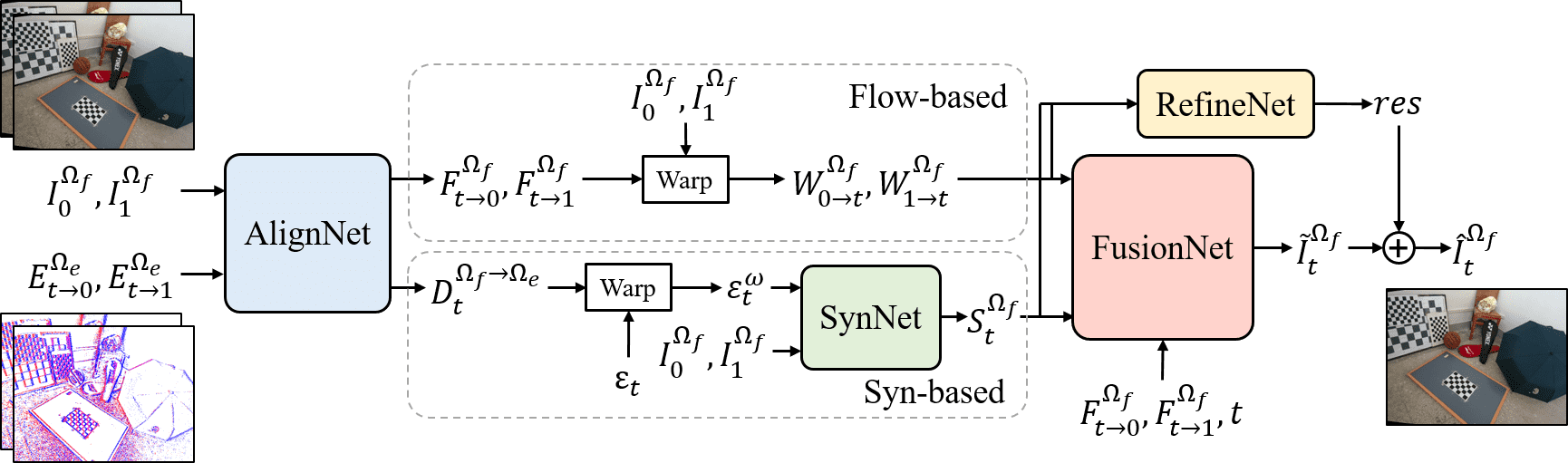}
   \caption{Overview of the proposed \mynetwork, which can be divided into two pathways, i.e., the \textit{Flow-based} and the \textit{Syn-based}, and has four subnets, \textit{AlignNet}, \textit{SynNet}, \textit{FusionNet}, and \textit{RefineNet}.}
   \label{Fig-overview}
\end{figure*}
Benefiting from extremely low latency, the event cameras can supplement the inter-frame information by emitting events asynchronously in response to the brightness change\cite{brandli2014240,gallego2020event}. Many works focus on video reconstruction solely with events, which can be seen as a type of VFI. These methods are mainly based
on RNN~\cite{rebecq2019high,zou2021learning,rebecq2019events} or GAN\cite{wang2019event,mostafavi2021learning}. However, it is still ill-posed since events only record the brightness change in the scenes, making it difficult to predict accurate brightness values from them.

Recently, event cameras have been adopted for high-quality VFI~\cite{tulyakov2021time,tulyakov2022time,he2022timereplayer,gao2022superfast,xiao2022eva,paikin2021efi}. Thanks to the extremely high temporal resolution, events can provide reliable motion prediction for dynamic scenes even with non-linear motions, resulting in better results than conventional F-VFI approaches.
Tulyakov \etal~\cite{tulyakov2021time} propose a network Time Lens that leverages events to compensate for inter-frame motions and textures in a hybrid manner, \ie, warping plus synthesis.
Further, Time Lens is extended to Time Lens++ where a motion spline estimator is introduced to predict reliable non-linear continuous flow from sparse events~\cite{tulyakov2022time}.
Instead of supervised learning, He~\etal~\cite{he2022timereplayer} design an unsupervised learning framework by applying cycle consistency to bridge the gap between synthesized and real-world events. In addition, Gao~\etal~\cite{gao2022superfast} propose an SNN-based fast-slow joint synthesis framework, \ie, SuperFast, for the high-speed E-VFI task. Xiao~\etal~\cite{xiao2022eva} analyze the drawbacks of existing methods and introduce a novel method named EVA$^2$ for E-VFI via cross-modal alignment and aggregation.

However, existing E-VFI approaches largely rely on synthetic data, restricted by per-pixel spatial alignment and ideal imaging without intense motion and sudden brightness change, which is difficult to fulfill in real-world applications since the events and frames are usually captured separately by an event camera and an intensity camera~\cite{tulyakov2021time,gehrig2021dsec}.
The misalignment between frames and events cannot be eliminated by global homography or stereo rectification, and inevitably produces cross-modal parallax and poses additional challenges to the VFI task, especially when dynamic scenes with complex motions and varying depths are encountered.

Therefore, our work focuses on handling the parallax between events and frames in \myname\, effectively utilizing events to compensate for inter-frame information and achieve high-quality video frame interpolation.

\subsection{Stereo Matching}
Stereo matching is the process of establishing pixel-to-pixel correspondences between two different views on the epipolar line. Many existing stereo matching methods are based on neural networks and leverage several advanced techniques to improve their performance, \eg, encoder-decoder architecture~\cite{zhang2019ga}, 3D convolution cost-volume module~\cite{kendall2017end,tulyakov2018practical,guo2022cvcnet,chen2023unambiguous}, adaptive sample aggregation module~\cite{xu2020aanet}, attention~\cite{xu2022attention}, transformer~\cite{li2021revisiting}, and unsupervised learning~\cite{uddin2022unsupervised}. However, frame-to-frame stereo matching methods are based on the assumption of high-quality imaging under ideal conditions, where there are no fast motion and high-dynamic-range scenes.

In recent works, event cameras with high-speed and high-dynamic-range imaging capabilities have been utilized to improve the performance of stereo-matching algorithms in various complex scenes.  Due to the fact that event and intensity cameras perceive the same light field, the edge information extracted from events and intensity images can be correlated to calculate the sparse disparity map~\cite{wang2021stereo,kim2022real}. Additionally, some learning-based methods are proposed to achieve dense disparity estimation. Specifically, Zou~\etal~\cite{zuo2021accurate} introduce an hourglass architecture with a pyramid attention module, extracting multi-scale features and performing fusion by using convolutional kernels of different sizes. Gu~\etal~\cite{gu2022self} propose a self-supervised learning framework, achieving data matching and disparity estimation by establishing gradient structure consistency between frames and events.

However, existing stereo-matching methods are limited by the frame rate of the input frame sequence.
In contrast, our method aims to estimate the corresponding disparity map during video frame interpolation, allowing the resulting disparity sequence to be decoupled from the frame rate of the input sequence. This enables a more comprehensive and effective representation of depth variations in the scene.

\section{Problem Formualtion}
Conventional Frame-based VFI (F-VFI) methods aim at reconstructing intermediate frames between two keyframes:
\begin{equation}\label{eq:f-vfi}
    I_t = \operatorname{F-VFI}(t; I_0, I_1),\quad t\in [0,1],
\end{equation}
where $I_{0}, I_{1}$ are input keyframes and $I_t$ is the target intermediate frame at a  normalized time $t \in [0,1]$. However, due to the lack of motion information between consecutive input keyframes, most Frame-based VFI (F-VFI) methods are built on simplified assumptions, \eg, linear motions~\cite{niklaus2020softmax,Lu_2022_CVPR,Niklaus_2018_CVPR,bao2019depth} or local movements~\cite{niklaus2017cvpr,niklaus2017iccv}, leading to performance degradation in real-world scenarios.

\begin{figure*}[t]
  \centering
   \includegraphics[width=\linewidth]{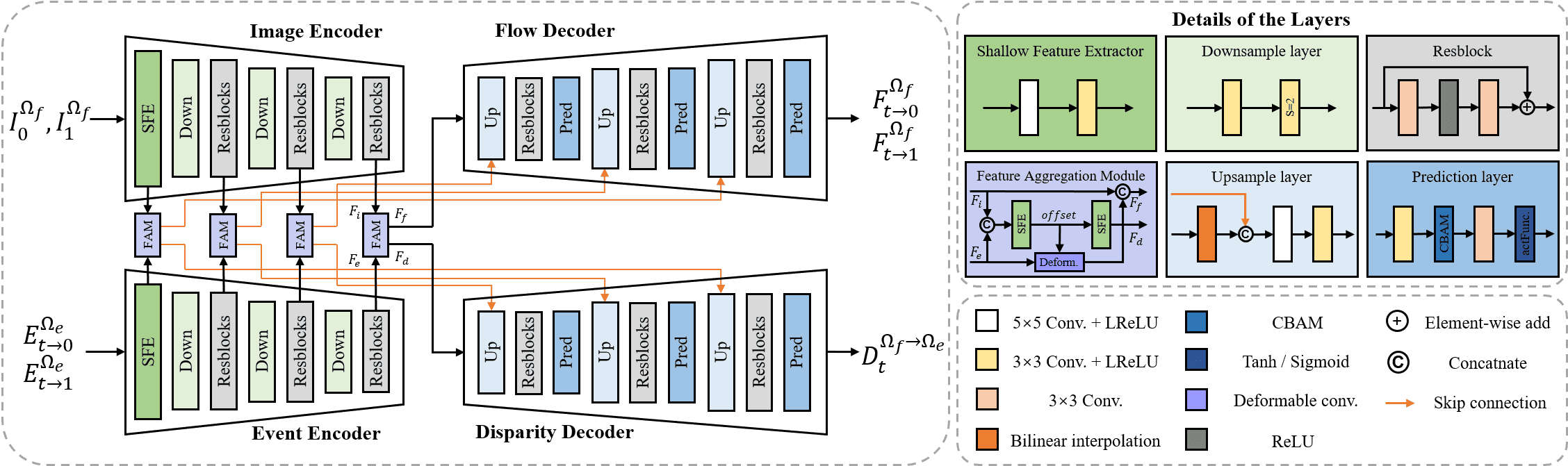}
   \caption{Structure of \textit{AlignNet}. We use different colors to represent different network layers, and the details of the layers are shown on the right side.}
   \label{Fig-alignnet}
\end{figure*}
To tackle this problem, E-VFI methods are proposed by predicting inter-frame motion information with events~\cite{tulyakov2021time,tulyakov2022time,he2022timereplayer,xiao2022eva,gao2022superfast}, achieving better interpolation performance than F-VFI methods.
They generate intermediate frames with the help of events triggered between input keyframes, thus can model motions close to the real trajectory:
\begin{equation}\label{eq:e-vfi}
    I_t^{\Omega_f} = \operatorname{E-VFI}(t; I_0^{\Omega_f}, I_1^{\Omega_f}; E_{0 \rightarrow 1}^{\Omega_f}),\quad t\in [0,1],
\end{equation}
where $\Omega_f$ represents the intensity camera's image plane, implying an assumption that the frames and events are aligned at the pixel level. However, this assumption is not always valid in existing stereo frame-event camera setups.
Parallax tends to exist between the two data, especially in foreground regions with small depths.
If we mix the unaligned events and frames directly for E-VFI, the parallax of the two data will bring burdens in motion estimation, leading to severe artifacts and distortion in the reconstruction result.

To tackle this problem, we introduce a novel framework for \myname\ with the misaligned event and frame data:
\begin{equation}\label{eq:se-vfi}
\begin{aligned}
    I_t^{\Omega_f}, \mathcal{D}_t^{\Omega_f \rightarrow \Omega_e} = \operatorname{\myname}(t; I_{0}^{\Omega_f}, I_{1}^{\Omega_f};E_{0 \rightarrow 1}^{\Omega_e}),\quad t \in [0,1]
\end{aligned}
\end{equation}
where $\Omega_f$ and $\Omega_e$ represent different image planes of intensity and event cameras respectively, and $\mathcal{D}_t^{\Omega_f \rightarrow \Omega_e}$ denotes the disparity from $\Omega_f$ to $\Omega_e$. The key to \myname\ is to deal with the parallax between events and frames by establishing a spatial data correlation, complementing the missing motion information between frames using events to realize the reconstruction of intermediate frames, and further estimating the disparity maps between events and frames from the spatial data correlation.

\section{Method}
In this section, we describe the proposed \mynetwork\ that can generate intermediate frames and disparities from misaligned events and frames.

\subsection{Pipeline Overview}
Our goal is to generate one or more intermediate frames $I_t^{\Omega_f}$ from consecutive keyframes $I_0^{\Omega_f}, I_1^{\Omega_f}$ in one view along with concurrent events $E_{0 \rightarrow 1}^{\Omega_e}$ in the other.
As shown in~\cref{Fig-overview}, \mynetwork\ contains four subnets, \textit{AlignNet}, \textit{SynNet}, \textit{FusionNet}, and \textit{RefineNet}. The main issue of \myname\ is to handle the parallax between the input events and frames. So we set AlignNet at first, which receives the origin data and mitigates the modality gap and spatial misalignment with the help of the key FAM. Specifically, we use FAM to establish the spatial correlation that is represented as the offset between inputs and obtain the aligned feature. Then the aligned feature and the offset are fed into two decoders to estimate bi-directional optical flow and disparity separately. Afterward, we split the VFI process into two pathways: the \textit{Flow-based} and the \textit{Syn-based}. The flow-based results demonstrate excellent performance in regions with small motions, while the syn-based results excel in regions with sufficient events.
Then these results are input to FusionNet to produce a high-quality output, which is further refined by RefineNet to improve the reconstruction quality.

\subsection{AlignNet}
Unlike conventional E-VFI methods, which usually concatenate frames and events as input, \myname\ focuses more on reducing the parallax effect in the data. Considering the modality discrepancy and the misalignment problem, AlignNet is based on an hourglass architecture~\cite{newell2016stacked} using two encoders and two decoders with similar structures but unshared weights as illustrated in \cref{Fig-alignnet}.

Assuming that we need to reconstruct the image at time $t$, we first divide the event stream into two parts and operate them by time shift and polarity reversal \cite{zhang2022unifying,tulyakov2021time} to get $E_{t\rightarrow 0}^{\Omega_e}, E_{t\rightarrow 1}^{\Omega_e}$,
and represent them by the voxel grid \cite{zhu2019unsupervised}.
The frames and events are initially fed into a Shallow Feature Extractor (SFE) to obtain full-scale features. Subsequently, they are passed through Downsample layers and Resblocks to obtain features at different scales.

However, the pivotal issue is the parallax between the inputs, so we design the FAM to tackle it. Inspired by \cite{dai2017deformable,chan2021understanding}, in FAM (see~\cref{Fig-alignnet}), we first concatenate the features $F_i$ and $F_e$, and then input them to an SFE to compute their spatial data correlation, which is represented as the offset. Since deformable convolution can sample flexible locations, we utilize the offset to guide the deformable convolution layer for feature alignment. We set the feature from the image encoder $F_i$ as the reference. Then the offset provides the data association between $F_i$ and $F_e$, guiding $F_e$ to align with $F_i$. Afterward, we concatenate the deformed feature with the input $F_i$ to obtain the aligned feature $F_f$ and pass the offset to another SFE to get the new feature $F_d$ that is used to estimate disparity.
The flow decoder and the disparity decoder differ only slightly in the prediction layers. At first, the CBAM~\cite{woo2018cbam} block is employed to enhance the features from both channel and spatial perspectives. Then the flow decoder uses Tanh as the activation function to be compatible with negative flow values, while the disparity decoder uses sigmoid since disparities are all positive. The features are fed into the two decoders that output bi-directional optical flows $F_{t \rightarrow 0}^{\Omega_f}, F_{t \rightarrow 1}^{\Omega_f}$ and the disparity $\mathcal{D}^{\Omega_f \rightarrow \Omega_e}_t$. In summary, the input and output of AlignNet are represented as:
\begin{equation}\label{eq:alignnet}
    F_{t \rightarrow 0}^{\Omega_f},F_{t \rightarrow 1}^{\Omega_f}, \mathcal{D}^{\Omega_f \rightarrow \Omega_e}_t = \operatorname{AlignNet}(I_0^{\Omega_f},I_1^{\Omega_f};E_{t \rightarrow 0}^{\Omega_e},E_{t \rightarrow 1}^{\Omega_e}).
\end{equation}

\subsection{SynNet}
\begin{figure}[t]
  \centering
   \includegraphics[width=0.85\linewidth]{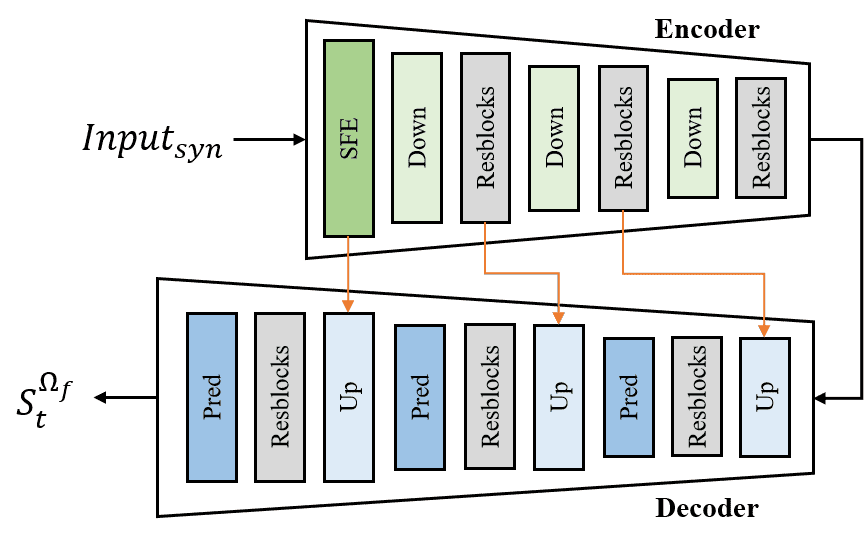}
   \caption{Architecture of \textit{SynNet}, which is a simple U-Net-like structure. The details of the network layers are described in~\cref{Fig-alignnet}
   .}
   \label{Fig-synnet}
\end{figure}
We divide our \mynetwork\ into two pathways, \ie, the flow-based and the syn-based. In the flow-based way, we utilize bi-directional flows to warp the boundary keyframes $I_0^{\Omega_f}, I_1^{\Omega_f}$ separately to the target time $t$ and get the warped images $W_{0 \rightarrow t}^{\Omega_f}$ and $W_{1 \rightarrow t}^{\Omega_f}$ as:
\begin{equation}\label{eq:flowwarp}
\begin{aligned}
        W_{0 \rightarrow t}^{\Omega_f} = \operatorname{Flow-Warp}(I_0^{\Omega_f}; F_{t \rightarrow 0}^{\Omega_f}),\\
        W_{1 \rightarrow t}^{\Omega_f}  = \operatorname{Flow-Warp}(I_1^{\Omega_f}; F_{t \rightarrow 1}^{\Omega_f}).
\end{aligned}
\end{equation}

However, the optical flow estimation relies on the assumption of brightness invariance, which can result in inaccuracies in regions where objects move rapidly. Thus we use the syn-based way to fix it.

Since events are emitted with the brightness change at the high temporal resolution, they can record complex motions and help reconstruct intermediate images. But the original events in the stereo camera setup are not aligned with the frames. So we first gather the events $e=\{x_i, y_i, t_i, p_i\}_{t=1}^N$ within a time window $\Delta t$ around the target time $t$ and then compress them into a tensor, which can be formulated as:
\begin{equation}\label{eq:ef}
    \varepsilon_t = \sum_{i \in \{i| t-\frac{\Delta t}{2}\le t_i \le t+\frac{\Delta t}{2} \}} p_i\delta(x-x_i,y-y_i).
\end{equation}

Subsequently, we use the previously learned disparity to warp the event tensors to get new tensors that match the spatial positions with the frames:
\begin{equation}\label{eq:dispwarp}
    \varepsilon_t^w = \operatorname{Disp-Warp}(\varepsilon_t; \mathcal{D}_t^{\Omega_f \rightarrow \Omega_e}).
\end{equation}
Then we concatenate the keyframes with the warped event tensors as the input ${Input}_{syn} = [I_0^{\Omega_f},I_1^{\Omega_f};\varepsilon_t^w]$ of the U-Net-like SynNet shown in \cref{Fig-synnet}, and finally we obtain a synthesized result:
\begin{equation}\label{eq:synnet}
    S_t^{\Omega_f} = \operatorname{SynNet}({Input}_{syn}).
\end{equation}

\subsection{FusionNet and RefineNet}
\begin{figure}[t]
  \centering
   \includegraphics[width=0.85\linewidth]{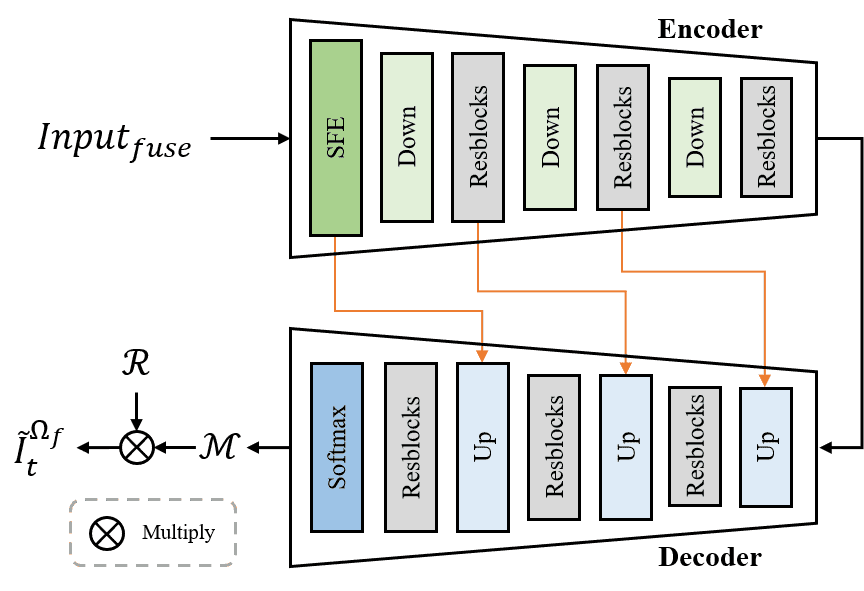}
   \caption{Architecture of \textit{FusionNet}, which outputs different attention maps to unify the warped and synthesized results together.}
   \label{Fig-fusionnet}
\end{figure}
The flow-based method performs well in regions with small motions and stable brightness, while the syn-based method relies more on regions with sufficient events and can handle complex scenes. Therefore, we use FuisonNet to combine the strengths of both methods and achieve better results. We use the warped and synthesized results together with the optical flow and target time $t$  as input, \ie, ${Input}_{fuse} = [W_{0 \rightarrow t}^{\Omega_f}, W_{1 \rightarrow t}^{\Omega_f}, S_t^{\Omega_f}, F_{t \rightarrow 0}^{\Omega_f}, F_{t \rightarrow 1}^{\Omega_f}, t]$, adjusting the proportion of each result by using the optical flow and target time as weights. FusionNet is shown in~\cref{Fig-fusionnet} and we use the Softmax function at the end to generate three attention maps $\mathcal{M} = \{ \mathcal{M}_0,\mathcal{M}_1,\mathcal{M}_2\}$ and then multiply $\mathcal{M}$ by the previous results $\mathcal{R} = \{W_{0 \rightarrow t}^{\Omega_f},W_{1 \rightarrow t}^{\Omega_f},S_t^{\Omega_f} \}$ to obtain the fused result $\tilde{I}_t^{\Omega_f}$:
\begin{equation}\label{eq:mask}
\begin{aligned}
        \mathcal{M} = \operatorname{FusionNet}({Input}_{fuse}),
\end{aligned}
\end{equation}
\begin{equation}\label{eq:fusion}
\begin{aligned}
        \tilde{I}_t^{\Omega_f} = \mathcal{M} \otimes \mathcal{R} = \mathcal{M}_0 \cdot W_{0 \rightarrow t}^{\Omega_f} + \mathcal{M}_1 \cdot W_{1 \rightarrow t}^{\Omega_f} + \mathcal{M}_2 \cdot S_t^{\Omega_f}.
\end{aligned}
\end{equation}

Finally, we use a Residual Dense Network (RDN)~\cite{zhang2018residual} as RefineNet, as shown in~\cref{Fig-overview} where the different results are concatenated and passed to it to learn a residual ${res}$, which is added to the fused result $\tilde{I}_t^{\Omega_f}$ to obtain the final result $\hat{I}_t^{\Omega_f}$, which can be formulated as:
\begin{equation}\label{eq:residual}
\begin{aligned}
        res =  \operatorname{RefineNet}&(W_{0 \rightarrow t}^{\Omega_f}, W_{1 \rightarrow t}^{\Omega_f}, S_t^{\Omega_f}),
\end{aligned}
\end{equation}
\begin{equation}\label{eq:refine}
\begin{aligned}
        \hat{I}_t^{\Omega_f} = &\tilde{I}_t^{\Omega_f} + res.
\end{aligned}
\end{equation}

\subsection{Loss Functions}
Our training loss comprises three parts: \textit{reconstruction loss} $\mathcal{L}_{rec}$, \textit{flow loss} $\mathcal{L}_{flow}$, and \textit{disparity loss} $\mathcal{L}_{disp}$.

The \textit{reconstruction loss} $\mathcal{L}_{rec}$ models the reconstruction quality of the intermediate frames. We denote the ground truth frames by $I_t^{gt}$ and utilize the $\mathcal{L}_1$ loss to evaluate the similarity between the reconstruction and the ground truth. To achieve better visual quality, we add the perceptual loss \cite{zhang2018unreasonable} to $\mathcal{L}_{rec}$ that is formulated as:
\begin{equation}\label{eq:loss_rec}
\begin{aligned}
    \mathcal{L}_{rec} =& \operatorname{\mathcal{L}_1}(S_t^{\Omega_f}, I_t^{gt}) + \operatorname{\mathcal{L}_1}(\tilde{I}_t^{\Omega_f}, I_t^{gt})+\operatorname{\mathcal{L}_1}(\hat{I}_t^{\Omega_f}, I_t^{gt})\\ &+ 0.1\operatorname{\mathcal{L}_{perceptual}}(\hat{I}_t^{\Omega_f}, I_t^{gt}).
\end{aligned}
\end{equation}

The \textit{flow loss} $\mathcal{L}_{flow}$ consists of the \textit{photometric loss} and the \textit{smoothness loss} used in \cite{Zhu-RSS-18}. The former aims to minimize the difference in intensity between the warped image and the ground truth, while the latter regularizes the output flow by minimizing the flow difference between adjacent pixels in the horizontal, vertical, and diagonal directions.
$\mathcal{L}_{flow}$ is formulated as:
\begin{equation}\label{eq:loss_flow}
\begin{aligned}
    \mathcal{L}_{flow} =
    &\operatorname{\mathcal{L}_{photometric}}(W_{0 \rightarrow t}^{\Omega_f}, I_t^{gt})+0.1\operatorname{\mathcal{L}_{smoothness}}(F_{t \rightarrow 0}^{\Omega_f})\\
    &+\operatorname{\mathcal{L}_{photometric}}(W_{1 \rightarrow t}^{\Omega_f}, I_t^{gt})+0.1\operatorname{\mathcal{L}_{smoothness}}(F_{t \rightarrow 1}^{\Omega_f}).
\end{aligned}
\end{equation}

The \textit{disparity loss} $\mathcal{L}_{disp}$ models the prediction quality of disparity. We denote the ground truth disparity by $\mathcal{D}_{t}^{gt}$, and first optimize the prediction by the smooth $\mathcal{L}_1$ loss since it is more robust, which is defined as:
\begin{equation}\label{eq:smooth_l1}
\begin{aligned}
\operatorname{smooth}_{L_{1}}(x) & = \left\{\begin{array}{ll}
0.5 x^{2} & \text { if }|x|<1, \\
|x|-0.5 & \text { otherwise. }
\end{array}\right.
\end{aligned}
\end{equation}
We also incorporate the edge-aware disparity smoothness loss $\mathcal{L}_{ds}$ used in \cite{godard2017unsupervised} to promote the local smoothness of disparities by computing the cost using the gradients of both disparities and frames, which is represented as:
\begin{equation}\label{eq:disp_smooth_loss}
\mathcal{L}_{ds} = \frac{1}{N}\sum_{i,j}|\partial_x \mathcal{D}_{t,i,j}^{\Omega_f \rightarrow \Omega_e}|e^{-||\partial_x I_{i,j}^{gt}||}+|\partial_y \mathcal{D}_{t,i,j}^{\Omega_f \rightarrow \Omega_e}|e^{-||\partial_y I_{i,j}^{gt}||},
\end{equation}
where $i, j$ are pixel coordinates, $x, y$ represent two directions, and $\partial D, \partial I$ are disparity gradients and image gradients.
The disparity loss can be expressed as:
\begin{equation}\label{eq:loss_disp}
\begin{aligned}
    \mathcal{L}_{disp} = \operatorname{smooth}_{L_{1}}(\mathcal{D}_t^{\Omega_f \rightarrow \Omega_e}, \mathcal{D}_{t}^{gt}) + 0.1\mathcal{L}_{ds}.
\end{aligned}
\end{equation}

The total training loss combines all of the above terms:
\begin{equation}\label{eq:loss_total}
\begin{aligned}
    \mathcal{L}_{total} &= \lambda_r\mathcal{L}_{rec} + \lambda_f\mathcal{L}_{flow} + \lambda_d\mathcal{L}_{disp}.
\end{aligned}
\end{equation}
In practice, we adjust the balancing factors for training on different datasets.

\begin{figure}[ht]
  \centering
   \includegraphics[width=0.7\linewidth]{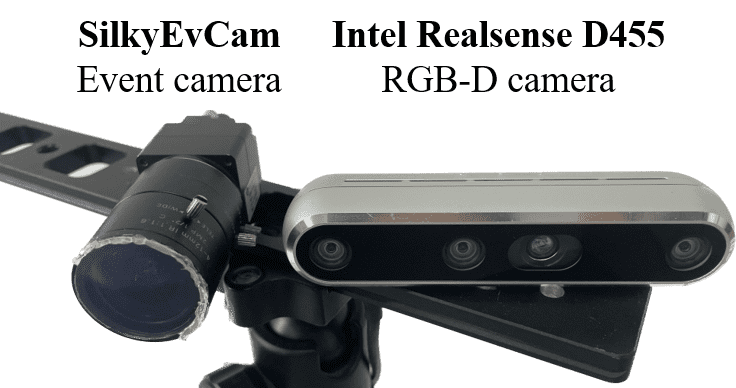}
   \caption{Illustration of the stereo event-intensity camera setup. It features a SilkyEvCam event camera (left) and an Intel Realsense D455 RGB-D camera (right). Both cameras are synchronized with a baseline of 8cm.}
   \label{Fig-SEID}
\end{figure}
\begin{table}[ht]
\small
\renewcommand{\arraystretch}{1.1}
\centering
\caption{Details of our \mydataset\ dataset. In total, the dataset consists of 34 sequences for both indoor and outdoor scenes.}
\vspace{-.5em}
\begin{tabular}{lllll}
\toprule[1.2pt] 
\textbf{Split} & \textbf{Subset} & \textbf{\#Sequences} & \textbf{\#Frames} & \textbf{\#Events (M)} \\ \hline
\textbf{Train} & Basketball & 6 & 2189 & 100.6\\
& Cars & 4 & 1516 & 152.6\\
& Checkerboard & 4 & 1189 & 119.1\\
& Indoor & 5 & 1736 & 132.2\\
& Pedestrians & 3 & 1310 & 141.8\\
& Square & 6 & 2433 & 277.3\\ \cline{3-5}
& & \textbf{28} & \textbf{10373} & \textbf{923.6}\\ \hline
\textbf{Test} & Basketball & 1 & 332 & 9.7\\
& Cars & 1 & 338 & 38.9\\
& Checkerboard & 1 & 298 & 20.6\\
& Indoor & 1 & 512 & 42.3\\
& Pedestrians & 1 & 241 & 34.7\\
& Square & 1 & 330 & 28.1\\ \cline{3-5}
& & \textbf{6} & \textbf{2051} & \textbf{174.3}\\  \hline
\textbf{Total} & & \textbf{34} & \textbf{12424} & \textbf{1097.8}\\
\toprule[1.2pt]
\end{tabular}
\label{tab:seid_info}
\end{table}
\section{Stereo Event-Intensity Dataset}
There are two public stereo event datasets, DSEC\cite{gehrig2021dsec} and MVSEC\cite{zhu2018multivehicle}. DSEC is a large-scale outdoor stereo event dataset especially for driving scenarios, while MVSEC captures a single indoor scene and multi-vehicle outdoor driving scenes. But for the VFI task, these two datasets have limited scene diversity and cannot effectively evaluate the algorithm's generalization ability across different scenes.
Additionally, these datasets record depths using LiDAR, which can only provide sparse depth values. Typically, only around $10\%$ of the depth values are valid, while the remaining $90\%$ are empty \cite{hu2022deep}. Therefore, the LiDAR data as the reference for VFI evaluation would have limited capability.

\begin{figure*}[t]
	\begin{minipage}[b]{0.49\linewidth}
		\centering
		\subfloat[Cars]{\includegraphics[width=0.49\linewidth]{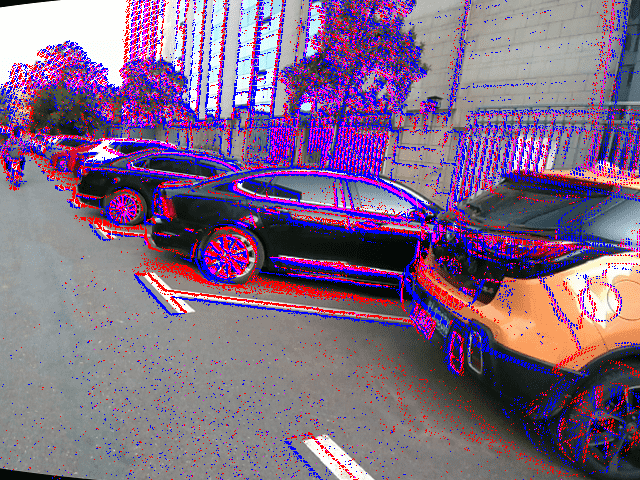}
        \includegraphics[width=0.49\linewidth]{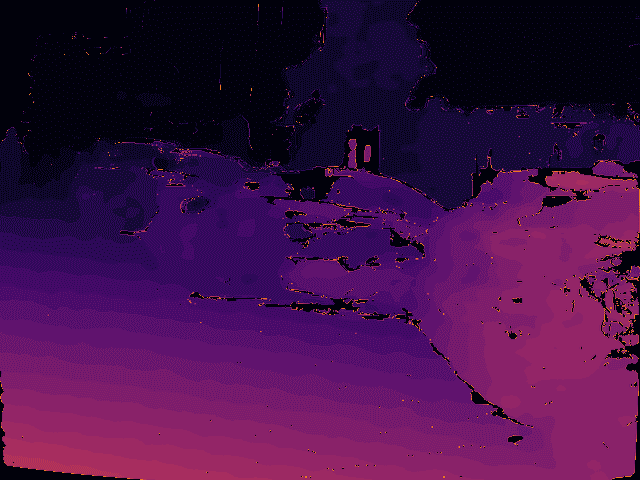}}
	\end{minipage}
     \begin{minipage}[b]{0.49\linewidth}
		\centering
		\subfloat[Indoor]{\includegraphics[width=0.49\linewidth]{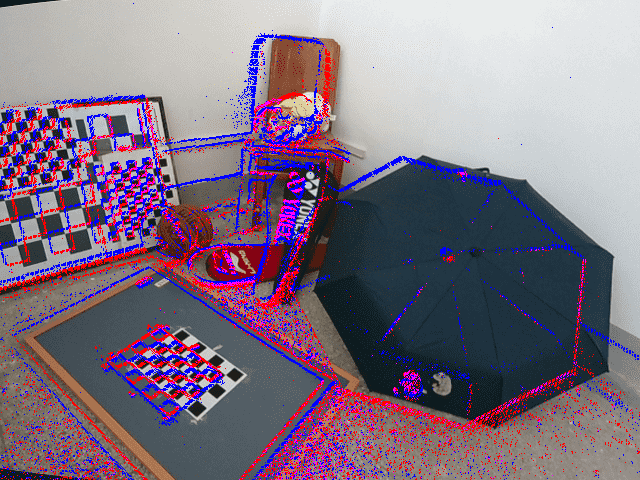}
        \includegraphics[width=0.49\linewidth]{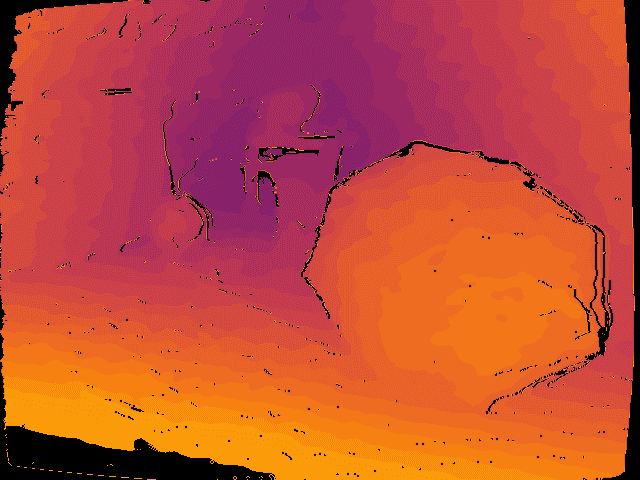}}
	\end{minipage}\\
	\begin{minipage}[b]{0.49\linewidth}
		\centering
		\subfloat[Pedestrians]{\includegraphics[width=0.49\linewidth]{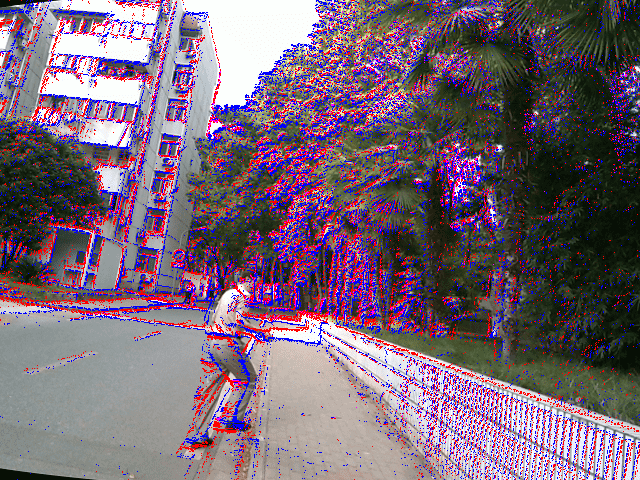}
        \includegraphics[width=0.49\linewidth]{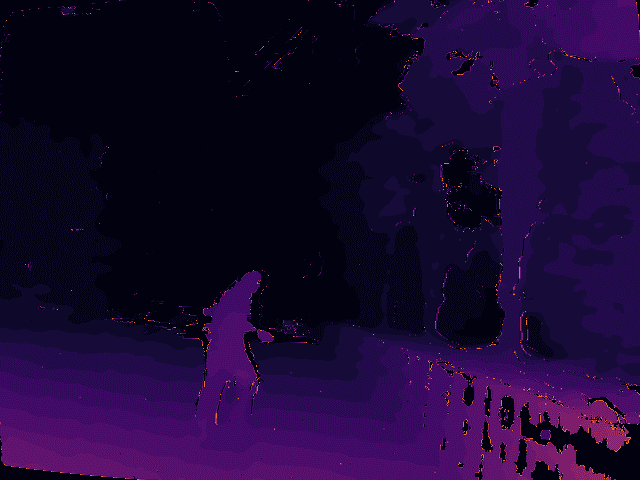}}
	\end{minipage}
    \begin{minipage}[b]{0.49\linewidth}
		\centering
		\subfloat[Square]{\includegraphics[width=0.49\linewidth]{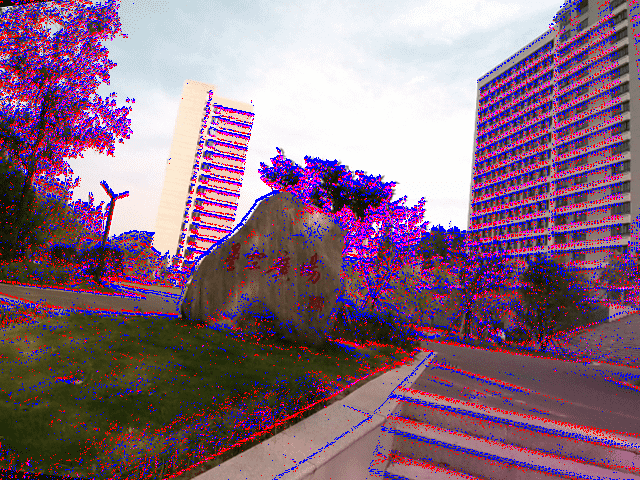}
        \includegraphics[width=0.49\linewidth]{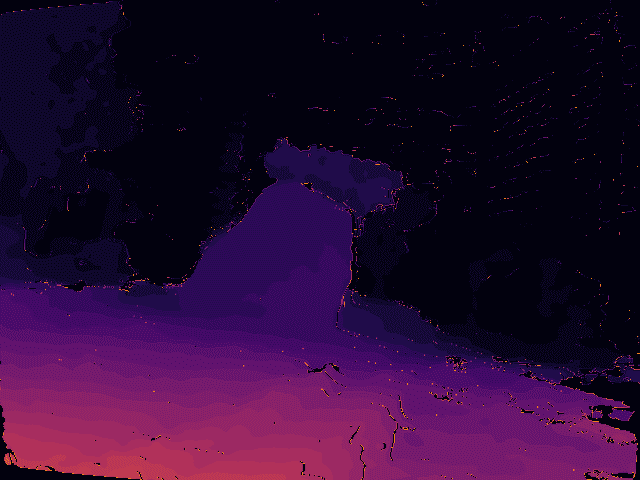}}
	\end{minipage}
	\caption{Illustration of example images rendered with events and their corresponding disparity images. Since we remove the depth values that are smaller than 1 meter to avoid errors during recording, there are holes in the disparity images.}
 \vspace{-4mm}
 \label{Fig-sample_seid}
\end{figure*}
\begin{figure*}[ht]
  \centering
  \begin{minipage}[b]{0.95\linewidth}
		\centering
        \subfloat[DSEC]{\includegraphics[width=0.33\linewidth]{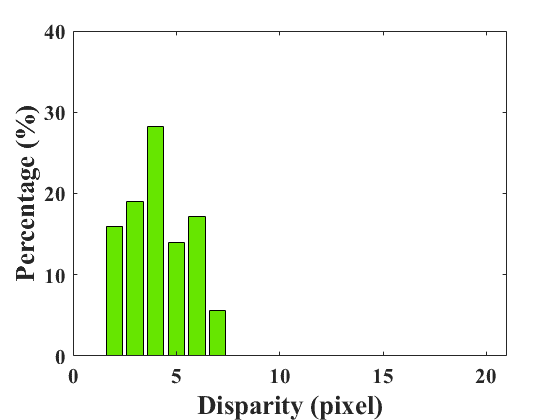}}
        \subfloat[MVSEC]{\includegraphics[width=0.33\linewidth]{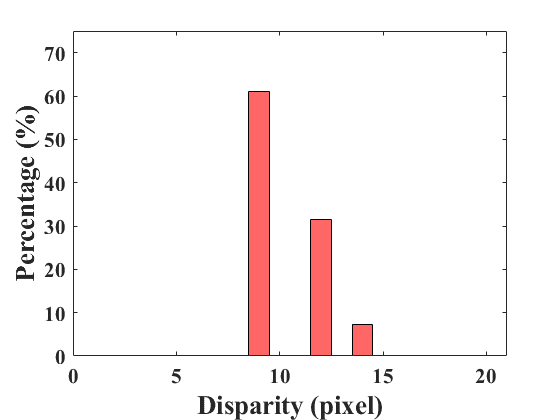}}
        \subfloat[SEID (Ours)]{\includegraphics[width=0.33\linewidth]{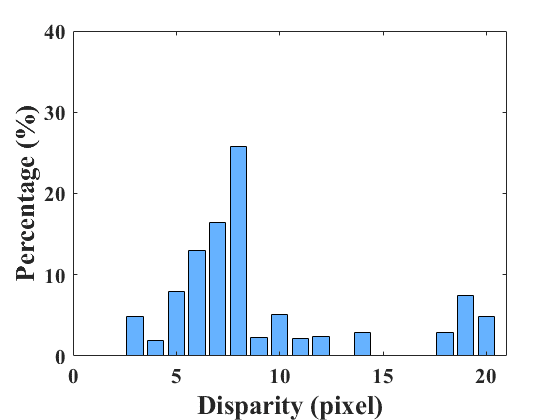}}
	\end{minipage}
   \caption{Illustration of the disparity distribution of each dataset. The disparity values of DSEC\cite{gehrig2021dsec} and MVSEC\cite{zhu2018multivehicle} are concentrated within a small range, indicating that the scenes they capture are similar and homogeneous. In contrast, our \mydataset\ dataset captures a wider range of scenes with richer variations in depth values, resulting in a broader disparity distribution.}
   \label{Fig-distribution}
   \vspace{-3mm}
\end{figure*}
To address this problem, we build a stereo visual acquisition system containing a SilkyEvCam event camera and an Intel Realsense D455 RGB-D camera with a baseline of 8cm as shown in ~\cref{Fig-SEID}. We collect a new \mydataset\ dataset that captures various indoor and outdoor scenes with varying depths and complex motions.
The RGB-D camera provides frames and depth maps that are synchronized in both temporal and spatial domains. We collect the data from different sensors through the ROS system and use the UTC clock of the control computer to timestamp the data for soft synchronization. In order to eliminate possible time delays, we manually calibrate the timestamps of events to achieve precise time synchronization.

To estimate the intrinsic and extrinsic parameters, we utilize the event-based image reconstruction algorithm E2VID~\cite{rebecq2019high} to generate high-quality images for calibration. In order to achieve good calibration results, we use the MATLAB calibration toolbox to calibrate the intrinsic parameters of the two cameras first, and then we use the obtained intrinsic parameters to calibrate the extrinsic parameters between the two cameras. Afterward, we employ the calibrated parameters to perform stereo rectification on the data, eliminating the deviations in the $y$ and $z$ directions. Furthermore, we remove the values of less than one meter to avoid invalid depth values captured by the RGB-D camera and then convert the depths to disparities using the calibrated parameters.

\begin{table}[t]
\centering
\small
\caption{Comparison of our \mydataset\ with the publicly available datasets MVSEC~\cite{zhu2018multivehicle} and DSEC~\cite{gehrig2021dsec}.}
\begin{tabular}{lcccc}
\toprule[1.2pt] 
 & FPS & Resolution & Color & Depth \\ \hline
MVSEC \cite{zhu2018multivehicle} & 32 & $346\times 260$ & \ding{55} & Sparse \\
DSEC \cite{gehrig2021dsec} & 20 & \bm{$640\times 480$} & \ding{51} & Sparse \\ \hline
SEID (Ours) & \textbf{60} & \bm{$640\times 480$} & \ding{51} & \textbf{Dense} \\ 
\toprule[1.2pt] 
\end{tabular}
\label{tab:seid}
\vspace{-2mm}
\end{table}
Our \mydataset\ dataset contains 34 indoor and outdoor sequences that are summarized in \cref{tab:seid_info}. The two cameras share a similar field of view (FoV), and both events and frames have a resolution of $640\times 480$. The dataset features frames at 60 FPS, which is higher than the previous stereo datasets as shown in \cref{tab:seid}. High spatial resolution helps capture richer texture details, while a high frame rate preserves more temporal and motion information, providing more valid references at various target timestamps for the VFI task. In addition, we use an RGB-D camera to capture the depths of the scenes, which provides dense disparity maps as shown in \cref{Fig-sample_seid} different from LiDAR data. This offers more reliable reference values for cross-modal stereo-matching tasks.
Our \mydataset\ dataset captures richer motions and varying depths, providing a more diverse range of depth variations and scene dynamics.
We calculate the average values for each disparity map and further compute the average disparity for each sequence. Then, we conduct a statistical analysis of the disparity distributions in the three datasets, as shown in \cref{Fig-distribution}. The DSEC and MVSEC datasets exhibit a concentration of disparity values within a narrow range, indicating their limited scene diversity. In contrast, our \mydataset\ dataset captures a wider range of scenes with richer variations in depths, resulting in a broader distribution of disparities. Therefore, \mydataset\ can be used to validate the impact of different disparity conditions on the performance of E-VFI. This enables us to fully exploit the potential of event cameras in stereo tasks and harness their capabilities.

\def\imgWidth{0.24\textwidth} 
\def\scc{(-1.9,-1.35)}
\def\ssxxsone{(-0.85,-0.15)} 
\def\ssyysone{(2.10, 0.80)} 
\def\ssizz{1.5cm} 
\def\ssmag{2}

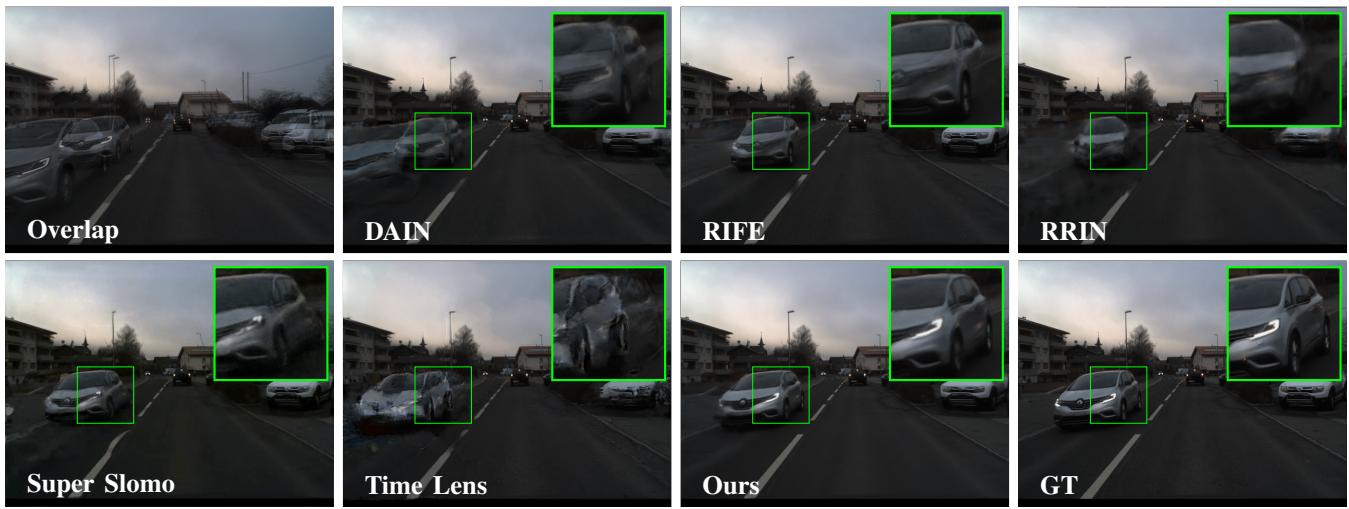
\begin{figure*}[!t] 
\centering
\tikzstyle{img} = [rectangle, minimum width=\imgWidth, draw=black]
        \begin{tikzpicture}[spy using outlines={green,magnification=\ssmag,size=\ssizz},inner sep=0]
            \node [align=center, img] {\includegraphics[width=\imgWidth]{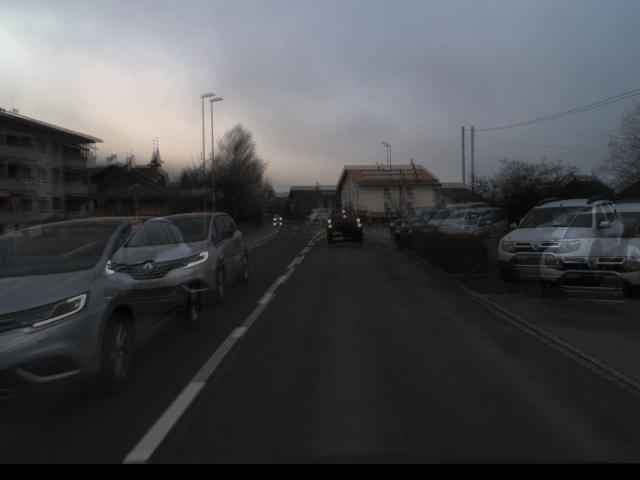}};
            \node [anchor=west] at \scc {\textcolor{white}{\bf Overlap}};
    	\end{tikzpicture}
		\begin{tikzpicture}[spy using outlines={green,magnification=\ssmag,size=\ssizz},inner sep=0]
            \node [align=center, img] {\includegraphics[width=\imgWidth]{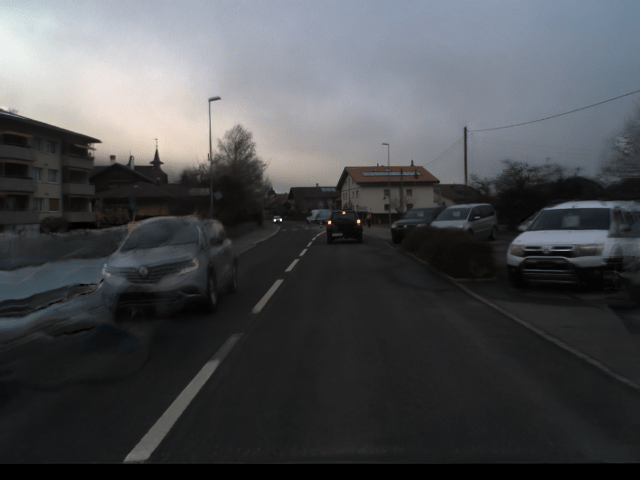}};
            \spy on \ssxxsone in node [left] at \ssyysone;
            \node [anchor=west] at \scc {\textcolor{white}{\bf DAIN}};
    	\end{tikzpicture}
        \begin{tikzpicture}[spy using outlines={green,magnification=\ssmag,size=\ssizz},inner sep=0]
            \node [align=center, img] {\includegraphics[width=\imgWidth]{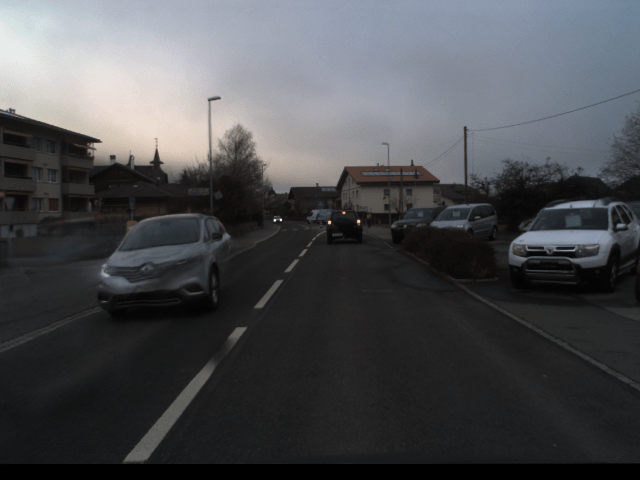}};
            \spy on \ssxxsone in node [left] at \ssyysone;
            \node [anchor=west] at \scc {\textcolor{white}{\bf RIFE}};
    	\end{tikzpicture}
		\begin{tikzpicture}[spy using outlines={green,magnification=\ssmag,size=\ssizz},inner sep=0]
            \node [align=center, img] {\includegraphics[width=\imgWidth]{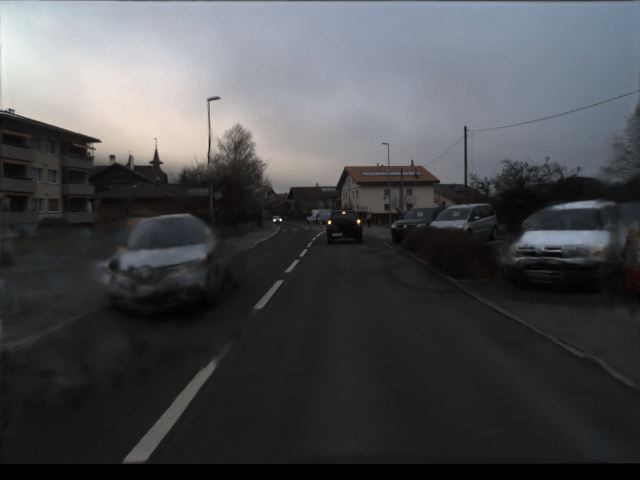}};
            \spy on \ssxxsone in node [left] at \ssyysone;
            \node [anchor=west] at \scc {\textcolor{white}{\bf RRIN}};
    	\end{tikzpicture}
\\
    \vspace{1mm}
		\begin{tikzpicture}[spy using outlines={green,magnification=\ssmag,size=\ssizz},inner sep=0]
            \node [align=center, img] {\includegraphics[width=\imgWidth]{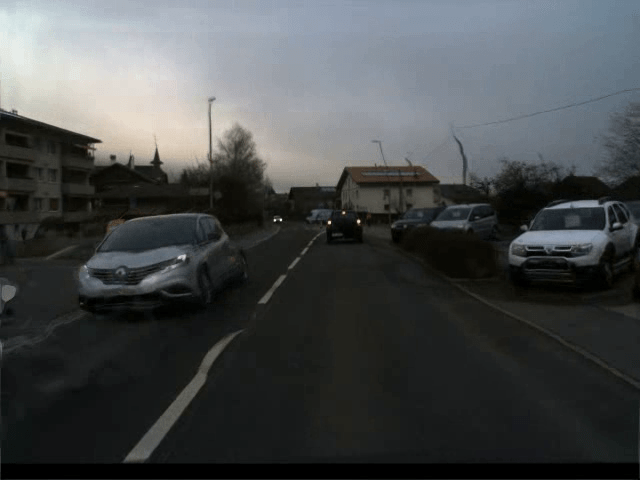}};
            \spy on \ssxxsone in node [left] at \ssyysone;
            \node [anchor=west] at \scc {\textcolor{white}{\bf Super Slomo}};
    	\end{tikzpicture}
		\begin{tikzpicture}[spy using outlines={green,magnification=\ssmag,size=\ssizz},inner sep=0]
            \node [align=center, img] {\includegraphics[width=\imgWidth]{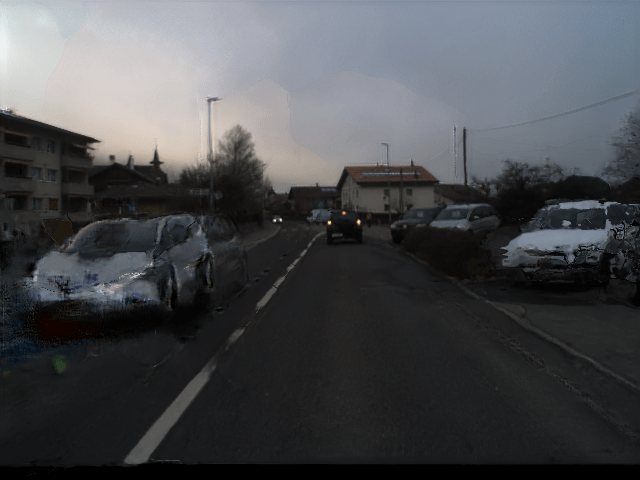}};
            \spy on \ssxxsone in node [left] at \ssyysone;
            \node [anchor=west] at \scc {\textcolor{white}{\bf Time Lens}};
    	\end{tikzpicture}
		\begin{tikzpicture}[spy using outlines={green,magnification=\ssmag,size=\ssizz},inner sep=0]
           \node [align=center, img] {\includegraphics[width=\imgWidth]{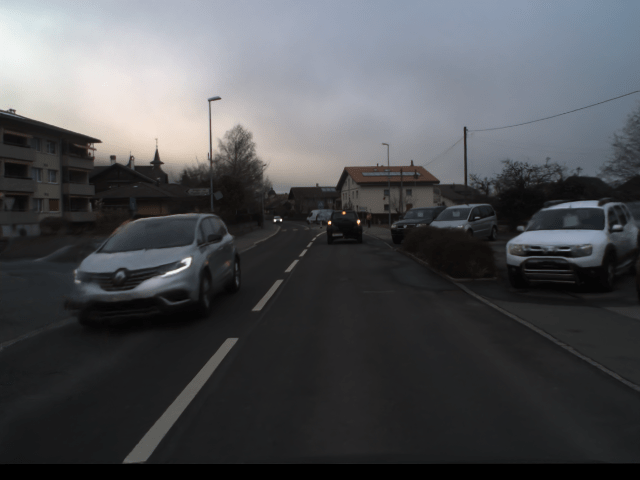}};
           \spy on \ssxxsone in node [left] at \ssyysone;
            \node [anchor=west] at \scc {\textcolor{white}{\bf Ours}};
    	\end{tikzpicture}
		\begin{tikzpicture}[spy using outlines={green,magnification=\ssmag,size=\ssizz},inner sep=0]
            \node [align=center, img] {\includegraphics[width=\imgWidth]{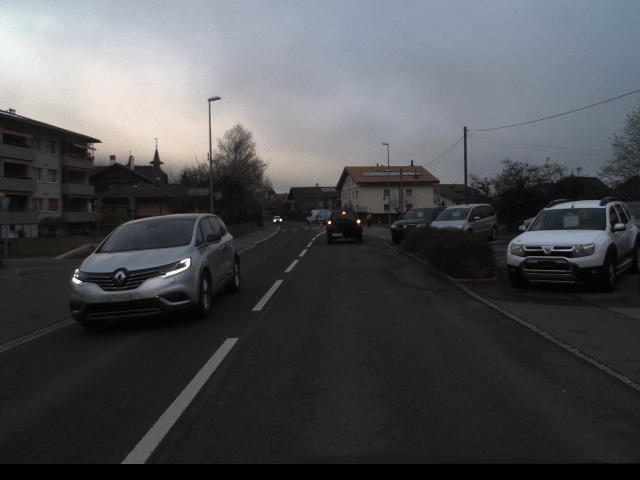}};
            \spy on \ssxxsone in node [left] at \ssyysone;
            \node [anchor=west] at \scc {\textcolor{white}{\bf GT}};
    	\end{tikzpicture}
	\caption{Qualitative comparisons of \mynetwork\ to state-of-the-art methods on the DSEC\cite{gehrig2021dsec} dataset. Zoom in on the details for a better view.}
	\label{Fig-dsec}
\end{figure*}

\def\imgWidth{0.24\textwidth} 
\def\scc{(-1.9,1.35)}
\def\ssxxsone{(-0.85,-1.30)} 
\def\ssyysone{(2.10, 0.85)} 
\def\ssizz{1.4cm} 
\def\ssmag{2}

\begin{figure*}[!t] 
\centering
\tikzstyle{img} = [rectangle, minimum width=\imgWidth, draw=black]
        \begin{tikzpicture}[spy using outlines={green,magnification=\ssmag,size=\ssizz},inner sep=0]
            \node [align=center, img] {\includegraphics[width=\imgWidth]{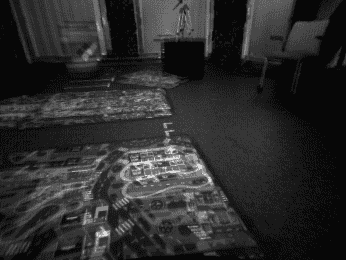}};
            \node [anchor=west] at \scc {\textcolor{white}{\bf Overlap}};
    	\end{tikzpicture}
		\begin{tikzpicture}[spy using outlines={green,magnification=\ssmag,size=\ssizz},inner sep=0]
            \node [align=center, img] {\includegraphics[width=\imgWidth]{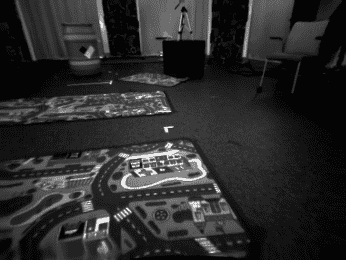}};
            \spy on \ssxxsone in node [left] at \ssyysone;
            \node [anchor=west] at \scc {\textcolor{white}{\bf DAIN}};
    	\end{tikzpicture}
        \begin{tikzpicture}[spy using outlines={green,magnification=\ssmag,size=\ssizz},inner sep=0]
            \node [align=center, img] {\includegraphics[width=\imgWidth]{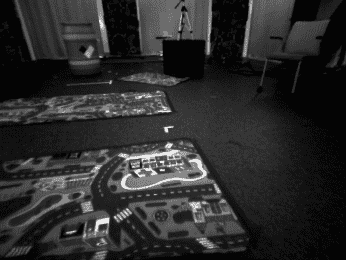}};
            \spy on \ssxxsone in node [left] at \ssyysone;
            \node [anchor=west] at \scc {\textcolor{white}{\bf RIFE}};
    	\end{tikzpicture}
		\begin{tikzpicture}[spy using outlines={green,magnification=\ssmag,size=\ssizz},inner sep=0]
            \node [align=center, img] {\includegraphics[width=\imgWidth]{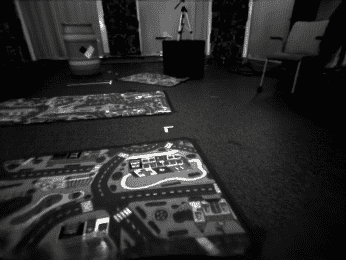}};
            \spy on \ssxxsone in node [left] at \ssyysone;
            \node [anchor=west] at \scc {\textcolor{white}{\bf RRIN}};
    	\end{tikzpicture}
\\
    \vspace{1mm}
		\begin{tikzpicture}[spy using outlines={green,magnification=\ssmag,size=\ssizz},inner sep=0]
            \node [align=center, img] {\includegraphics[width=\imgWidth]{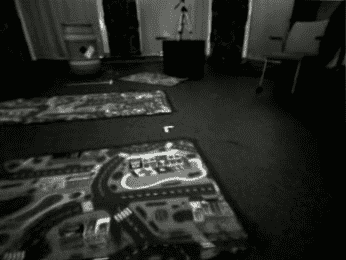}};
            \spy on \ssxxsone in node [left] at \ssyysone;
            \node [anchor=west] at \scc {\textcolor{white}{\bf Super Slomo}};
    	\end{tikzpicture}
		\begin{tikzpicture}[spy using outlines={green,magnification=\ssmag,size=\ssizz},inner sep=0]
            \node [align=center, img] {\includegraphics[width=\imgWidth]{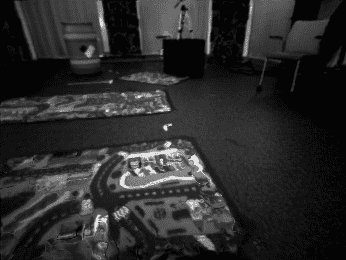}};
            \spy on \ssxxsone in node [left] at \ssyysone;
            \node [anchor=west] at \scc {\textcolor{white}{\bf Time Lens}};
    	\end{tikzpicture}
		\begin{tikzpicture}[spy using outlines={green,magnification=\ssmag,size=\ssizz},inner sep=0]
           \node [align=center, img] {\includegraphics[width=\imgWidth]{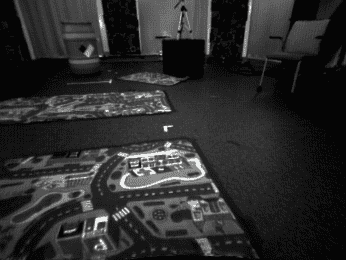}};
           \spy on \ssxxsone in node [left] at \ssyysone;
            \node [anchor=west] at \scc {\textcolor{white}{\bf Ours}};
    	\end{tikzpicture}
		\begin{tikzpicture}[spy using outlines={green,magnification=\ssmag,size=\ssizz},inner sep=0]
            \node [align=center, img] {\includegraphics[width=\imgWidth]{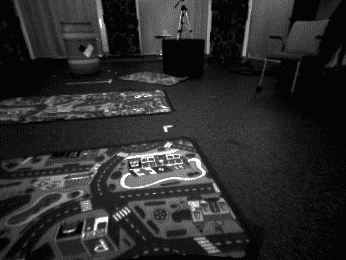}};
            \spy on \ssxxsone in node [left] at \ssyysone;
            \node [anchor=west] at \scc {\textcolor{white}{\bf GT}};
    	\end{tikzpicture}
	\caption{Qualitative comparisons of \mynetwork\ to state-of-the-art methods on the MVSEC\cite{zhu2018multivehicle} dataset. Zoom in on the details for a better view.}
	\label{Fig-mvsec}
\end{figure*}
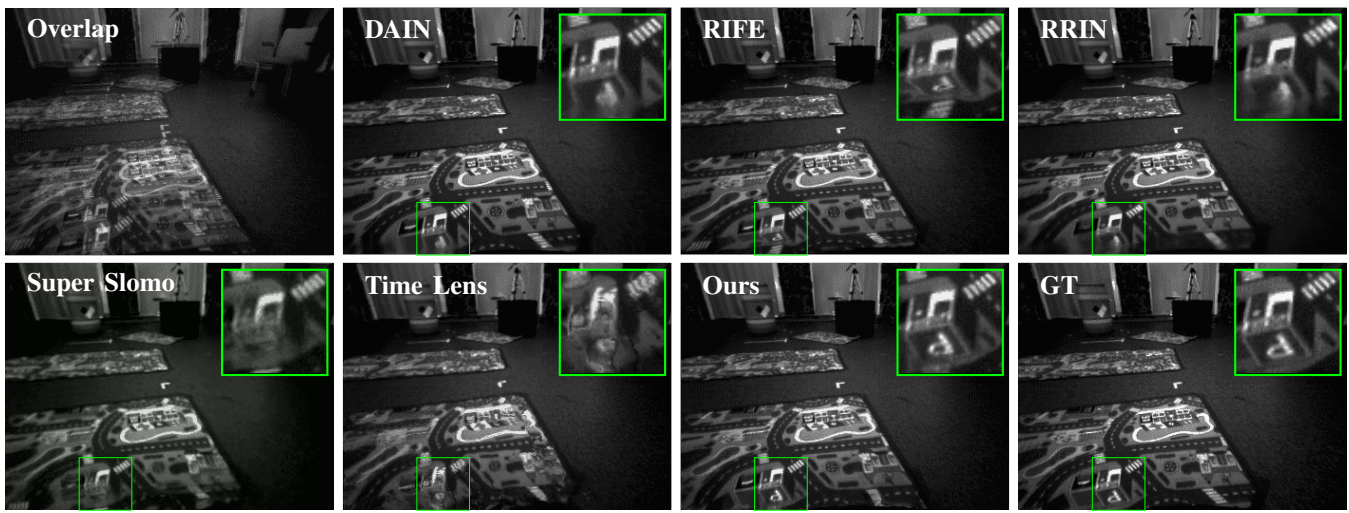

\section{Experiments and Analysis}
In this section, we evaluate and analyze the proposed \mynetwork. In \cref{experiment_settings}, we first describe the experimental settings, including datasets, training details, and compared methods. Then we compare
the VFI performance with the state-of-the-art in \cref{compare_VFI} and analyze the impact of disparity on the E-VFI task. After that, we evaluate the performance of stereo matching with existing cross-modal stereo matching methods in \cref{compare_SM}. Finally, the importance of the subnets in our \mynetwork\ is discussed in \cref{ablation}.

\subsection{Experimental Settings}\label{experiment_settings}
\subsubsection{Datasets}
We use DSEC\cite{gehrig2021dsec}, MVSEC\cite{zhu2018multivehicle}, and our \mydataset\ datasets for training and evaluation.

\textbf{\textit{DSEC.}}
We select 23 daylight event-frame sequences for training and 10 for testing, using the data from the left intensity camera and the right event camera. Both events and frames are rectified to eliminate distortions and deviations, and the frames are resized to 640 $\times $ 480 pixels to keep the same resolution as the events.

\textbf{\textit{MVSEC.}}
We use the {\it ``indoor$\_$flying''} scene data to verify the performance of our model. We choose 3 sequences as the training set and 1 sequence as the testing set, also with the left frames and right events. The extrinsic parameters are used to convert the provided depths to disparities, both frames and events are undistorted and rectified with the calibration parameters.

\textbf{\textit{SEID (Ours).}}
We divide \mydataset\ into two parts, \ie, the training set containing 28 sequences and the testing set containing 6 sequences. Each part consists of six subsets as illustrated in \cref{tab:seid_info}. The events, frames, and disparities are undistorted and rectified using the calibration parameters.

\subsubsection{Training Details}
We implement the proposed \mynetwork\ in Pytorch \cite{paszke2019pytorch} and train three models separately on DSEC, MVSEC, and \mydataset\ datasets. To enhance the robustness of \mynetwork, we augment the data by randomly cropping the samples into $224\times 224$ patches. It is trained using the Adam optimizer \cite{kingma2014adam} with $\beta_1=0.9$ and $\beta_2=0.999$ for 100 epochs.
The learning rate is initialized at $3\times 10^{-4}$ and decayed by 0.8 every 10 epochs. As different datasets record different scenes with varying disparities, we adjust the balancing parameters in \cref{eq:loss_total} for training.
We set $[\lambda_r,~ \lambda_f,~ \lambda_d] = [2, 0.005, 0.008]$ for training on DSEC, $[\lambda_r,~ \lambda_f,~ \lambda_d] = [1, 0.01, 0.001]$ for MVSEC, and $[\lambda_r,~ \lambda_f,~ \lambda_d] = [2, 0.002, 0.005]$ for \mydataset.
The model is trained for 100 epochs on each dataset, using 2 NVIDIA TITAN RTX3090 GPUs.

\begin{table*}[ht]
\centering
\small
\caption{Quantitative comparisons of the proposed \mynetwork\ to state-of-the-art methods on the DSEC\cite{gehrig2021dsec}, MVSEC\cite{zhu2018multivehicle}, and our \mydataset\ datasets. We compute PSNR and SSIM on the reconstructed frames.}
\vspace{-3mm}
\begin{tabular}{lcccccccccc}
\toprule[1.2pt] 
 Method & Event & & PSNR $\uparrow$ & SSIM $\uparrow$ & & PSNR $\uparrow$ & SSIM $\uparrow$ & & PSNR $\uparrow$ & SSIM $\uparrow$ \\ \hline
\textbf{\textit{DSEC}} & & & \multicolumn{2}{c}{1-frame skip} & & \multicolumn{2}{c}{3-frame skip} & & \multicolumn{2}{c}{5-frame skip} \\ \hline
DAIN\cite{bao2019depth} & \ding{55} & & 30.36 & 0.8977 & & 27.72 & 0.8535 & & \underline{25.96} & \underline{0.8147} \\
RIFE\cite{huang2022real} & \ding{55} & & \underline{32.24} & 0.9008 & & \underline{28.58} & \underline{0.8541} & & 25.75 & 0.7969 \\
RRIN\cite{li2020video} & \ding{55} & & 32.15 & \textbf{0.9172} & & 27.25 & 0.8451 & & 24.99 & 0.7950 \\
Super Slomo\cite{jiang2018super} & \ding{55} & & 28.45 & 0.8507 & & 26.31 & 0.8101 & & 24.76 & 0.7726 \\
Time Lens\cite{tulyakov2021time} & \ding{51} & & 29.36 & 0.8661 & & 26.47 & 0.8144 & & 24.77 & 0.7735 \\
\mynetwork\ (Ours) & \ding{51} & & \textbf{32.97} & \underline{0.9141} & & \textbf{30.29} & \textbf{0.8834} & & \textbf{26.69} & \textbf{0.8186}\\
\hline
\textbf{\textit{MVSEC}} & & & \multicolumn{2}{c}{3-frame skip} & & \multicolumn{2}{c}{5-frame skip} & & \multicolumn{2}{c}{7-frame skip} \\ \hline
DAIN\cite{bao2019depth} & \ding{55} & & 32.40 & 0.9124 & & 30.87 & \underline{0.8814} & & 29.62 & \underline{0.8512} \\
RIFE\cite{huang2022real} & \ding{55} & & \underline{35.09} & 0.9159 & & \underline{32.08} & 0.8648 & & \underline{31.45} & 0.8440 \\
RRIN\cite{li2020video} & \ding{55} & & 33.57 & \underline{0.9168} & & 31.40 & 0.8725 & & 29.83 & 0.8336 \\
Super Slomo\cite{jiang2018super} & \ding{55} & & 28.61 & 0.7843 & & 27.82 & 0.7611 & & 27.05 & 0.7383 \\
Time Lens\cite{tulyakov2021time} & \ding{51} & & 31.28 & 0.8692 & & 29.81 & 0.8341 & & 28.70 & 0.8049 \\
\mynetwork\ (Ours) & \ding{51} & & \textbf{35.23} & \textbf{0.9305} & & \textbf{33.79} & \textbf{0.9053} & & \textbf{32.04} & \textbf{0.8741}\\ \hline
\textbf{\textit{\mydataset}} & & & \multicolumn{2}{c}{5-frame skip} & & \multicolumn{2}{c}{7-frame skip} & & \multicolumn{2}{c}{9-frame skip} \\ \hline
 DAIN\cite{bao2019depth} & \ding{55} & & 26.38 & 0.8418 & & 24.84 & 0.7959 & & 23.50 & 0.7538 \\
 RIFE\cite{huang2022real} & \ding{55} & & \underline{27.53} & \underline{0.8550} & & \underline{26.29} & \underline{0.8187} & & \underline{24.00} & 0.7569 \\
 RRIN\cite{li2020video} & \ding{55} & & 26.32 & 0.8470 & & 24.00 & 0.7831 & & 22.34 & 0.7307 \\
 Super Slomo\cite{jiang2018super} & \ding{55} & & 26.09 & 0.8353 & & 24.22 & 0.7862 & & 22.67 & 0.7408 \\
 Time Lens\cite{tulyakov2021time} & \ding{51} & & 25.91 & 0.8252 & & 24.46 & 0.7895 & & 23.32 & \underline{0.7602} \\
 \mynetwork\ (Ours) & \ding{51} & & \textbf{29.27} & \textbf{0.8881} & & \textbf{27.51} & \textbf{0.8533} & & \textbf{25.83} & \textbf{0.8138}\\ 
 \toprule[1.2pt] 
\end{tabular}
\label{tab:quantitative}
\end{table*}
\subsubsection{Benchmark}
We compare \mynetwork\ with one open-sourced E-VFI approach, \ie, Time Lens\cite{tulyakov2021time} and four state-of-the-art F-VFI methods, DAIN\cite{bao2019depth}, RIFE\cite{huang2022real}, RRIN\cite{li2020video}, and Super Slomo\cite{jiang2018super}. The metrics of Peak Signal to Noise Ratio (PSNR) and Structural SIMilarity (SSIM)\cite{wang2004image} are used for quantitative evaluation. The higher the values of PSNR and SSIM, the better the performance.

Since the datasets have different frame rates, we adaptively evaluate each with different skip values to ensure similar inter-frame motion durations.
Specifically, we implement the experiments under the setting of 1-, 3-, and 5-frame skips for sequences in DSEC, 3-, 5-, and 7-frame skips in MVSEC, and 5-, 7-, and 9-frame skips in \mydataset.

\subsection{Comparisons of Video Frame Interpolation}\label{compare_VFI}
In this subsection, comparisons of our \mynetwork\ to the state-of-the-art methods are made qualitatively and quantitatively in multi-frame reconstruction. After that, the superiority and robustness of \mynetwork\ compared to other E-VFI methods in dealing with disparity changes are also verified.

\subsubsection{Results of F-VFI and E-VFI}
In this part, we analyze the limitations of the F-VFI methods and identify the challenges encountered by the E-VFI methods when handling stereo event-frame data. Finally, we highlight the effectiveness of our method. Qualitative and quantitative results are provided to show the superiority of our \mynetwork.

\textbf{\textit{Qualitative comparisons.}}
We select two samples from the DSEC and MVSEC datasets to compare the visual performance of each VFI algorithm as illustrated in \cref{Fig-dsec,Fig-mvsec}.

\def\cimwidth{0.185}
\def\zuoxia{(-3.95,-0.25)}
\def\youshang{(-1.70,1.45)}
\def\zuox{(0.55,-0.25)}
\def\yous{(2.8,1.45)}
\def\ssizz{1.2cm}
\def\ssmag{2}

\begin{figure}[ht]
\footnotesize
	\centering
 \begin{minipage}[t]{\linewidth}
    		\centering
      \captionsetup[subfloat]{labelsep=none,format=plain,labelformat=empty}
                \begin{tikzpicture}[spy using outlines={rectangle,green,magnification=\ssmag,size=\ssizz},inner sep=0]
				\node {
                \subfloat[Frame 0]{\includegraphics[width=0.5\linewidth]{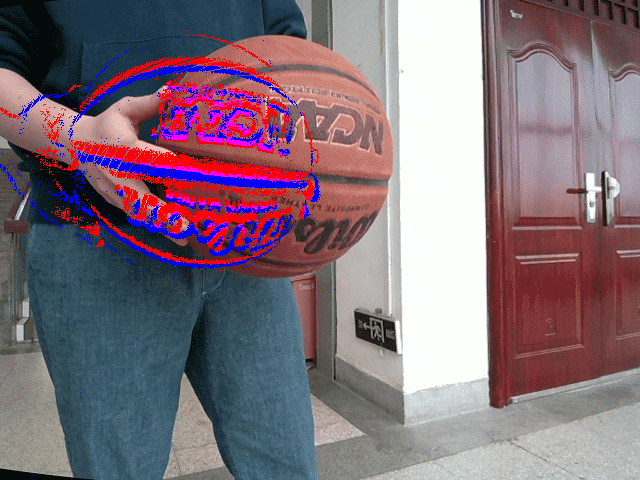}}
                \subfloat[Frame 1]{\includegraphics[width=0.5\linewidth]{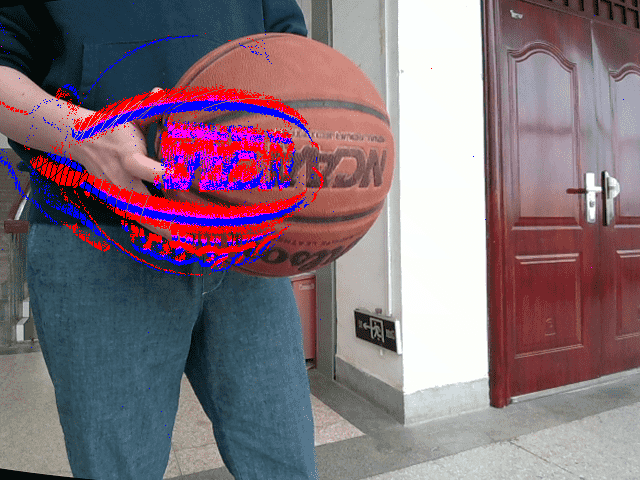}}
                };
				\draw[line width=1pt,green] \zuoxia rectangle \youshang;
                \draw[line width=1pt,green] \zuox rectangle \yous;
				\end{tikzpicture}
    	\end{minipage}
\vspace{-7mm}
\begin{minipage}[]{\linewidth}
            \vspace{-2.5mm}
    		\centering
      \captionsetup[subfloat]{labelsep=none,format=plain,labelformat=empty}
      \begin{tikzpicture}[inner sep=0]
            \node [label={[label distance=0.27cm,text depth=-1ex,rotate=90]right: \textcolor{black}{\scriptsize {RIFE}}}] at (15,15) {};
            \end{tikzpicture}
      \subfloat[]{\includegraphics[width=\cimwidth\textwidth,trim={80 200 240 40},clip]{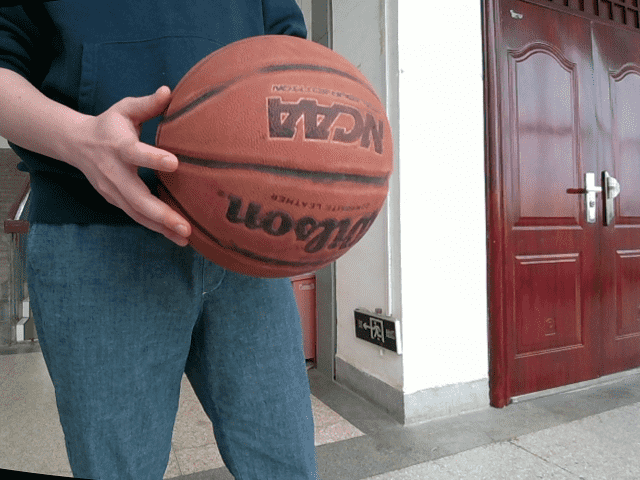}}\hfill
      \subfloat[]{\includegraphics[width=\cimwidth\textwidth,trim={80 200 240 40},clip]{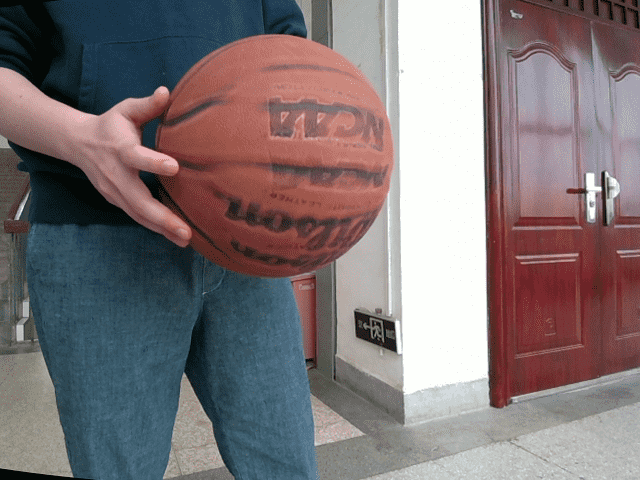}}\hfill
	   \subfloat[]{\includegraphics[width=\cimwidth\textwidth,trim={80 200 240 40},clip]{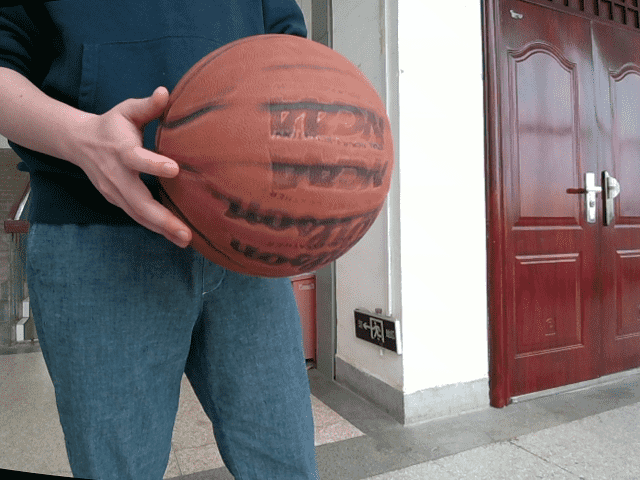}}\hfill
	   \subfloat[]{\includegraphics[width=\cimwidth\textwidth,trim={80 200 240 40},clip]{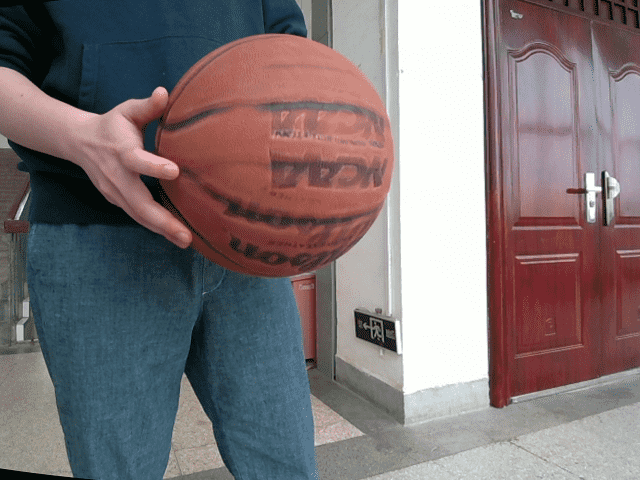}}\hfill
	   \subfloat[]{\includegraphics[width=\cimwidth\textwidth,trim={80 200 240 40},clip]{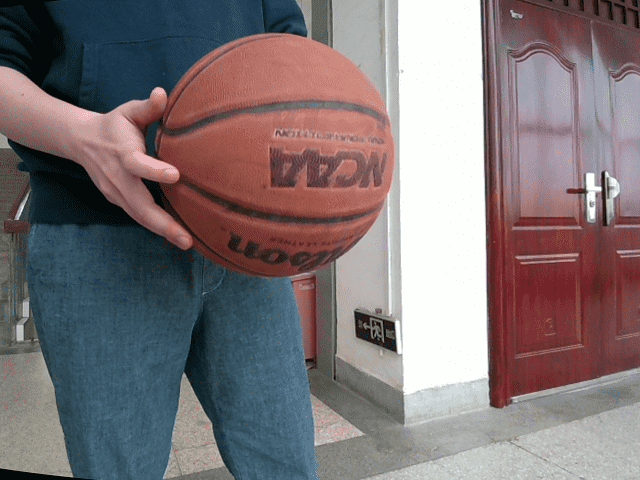}}\hfill
    \end{minipage}
    \vspace{-7mm}
    \begin{minipage}[]{\linewidth}
    		\centering
      \captionsetup[subfloat]{labelsep=none,format=plain,labelformat=empty}
      \begin{tikzpicture}[inner sep=0]
            \node [label={[label distance=0.05cm,text depth=-1ex,rotate=90]right: \textcolor{black}{\scriptsize {Time Lens}}}] at (15,15) {};
            \end{tikzpicture}
      \subfloat[]{\includegraphics[width=\cimwidth\textwidth,trim={80 200 240 40},clip]{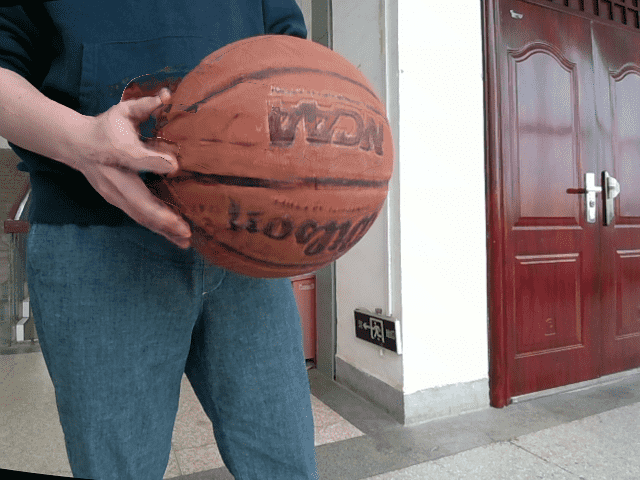}}\hfill
      \subfloat[]{\includegraphics[width=\cimwidth\textwidth,trim={80 200 240 40},clip]{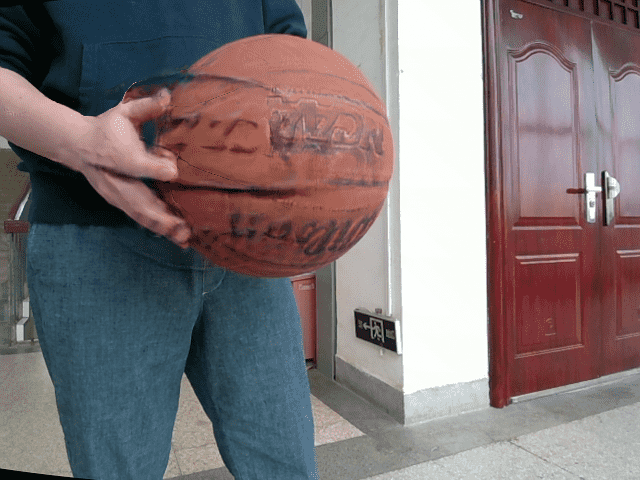}}\hfill
	   \subfloat[]{\includegraphics[width=\cimwidth\textwidth,trim={80 200 240 40},clip]{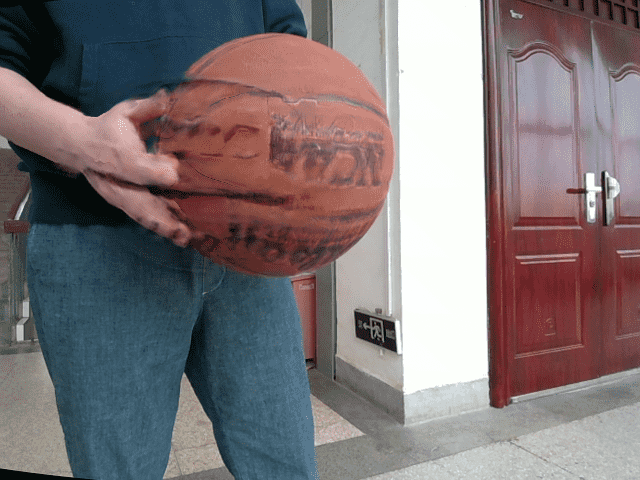}}\hfill
	   \subfloat[]{\includegraphics[width=\cimwidth\textwidth,trim={80 200 240 40},clip]{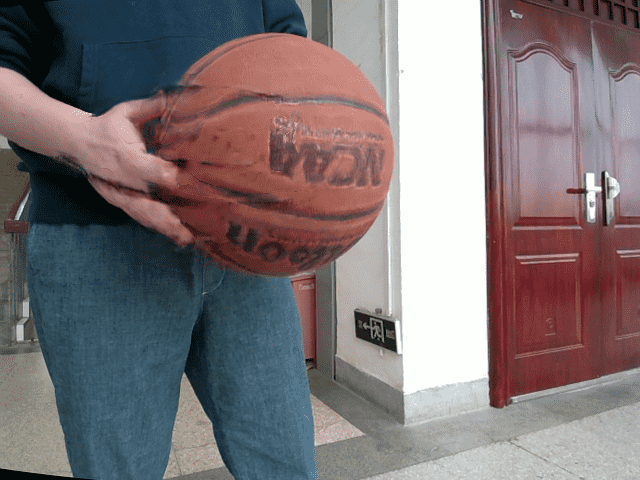}}\hfill
	   \subfloat[]{\includegraphics[width=\cimwidth\textwidth,trim={80 200 240 40},clip]{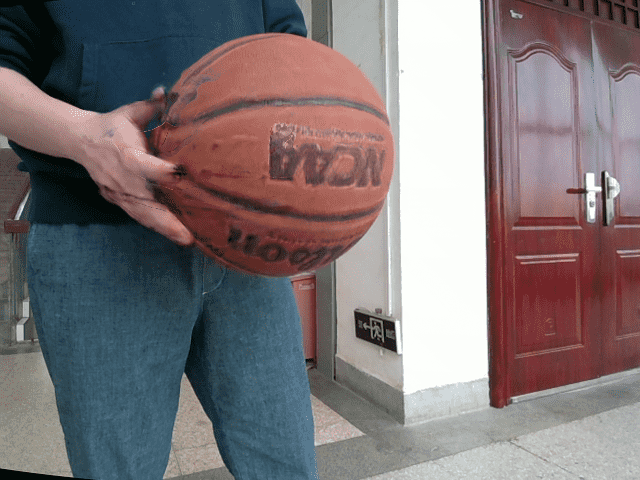}}\hfill
    \end{minipage}
    \vspace{-7mm}
    \begin{minipage}[]{\linewidth}
    		\centering
      \captionsetup[subfloat]{labelsep=none,format=plain,labelformat=empty}
      \begin{tikzpicture}[inner sep=0]
            \node [label={[label distance=0.33cm,text depth=-1ex,rotate=90]right: \textcolor{black}{\scriptsize {Ours}}}] at (15,15) {};
            \end{tikzpicture}
      \subfloat[]{\includegraphics[width=\cimwidth\textwidth,trim={80 200 240 40},clip]{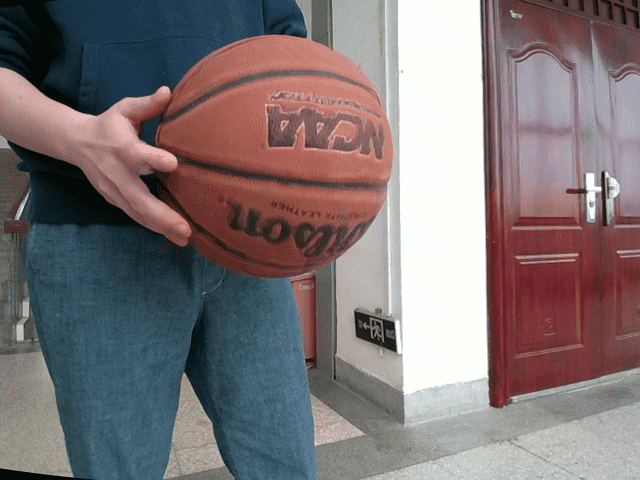}}\hfill
      \subfloat[]{\includegraphics[width=\cimwidth\textwidth,trim={80 200 240 40},clip]{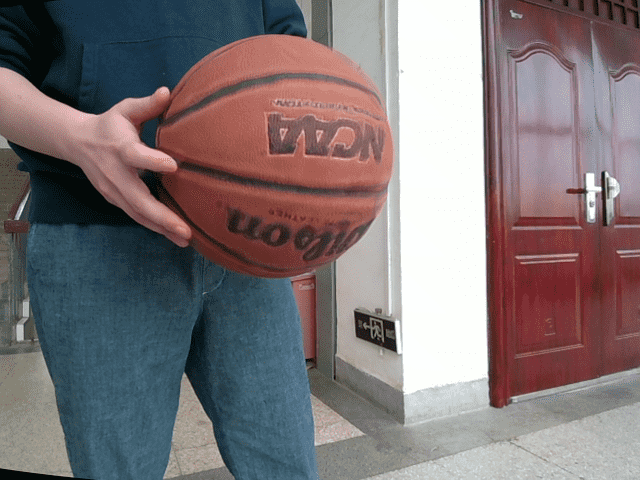}}\hfill
	   \subfloat[]{\includegraphics[width=\cimwidth\textwidth,trim={80 200 240 40},clip]{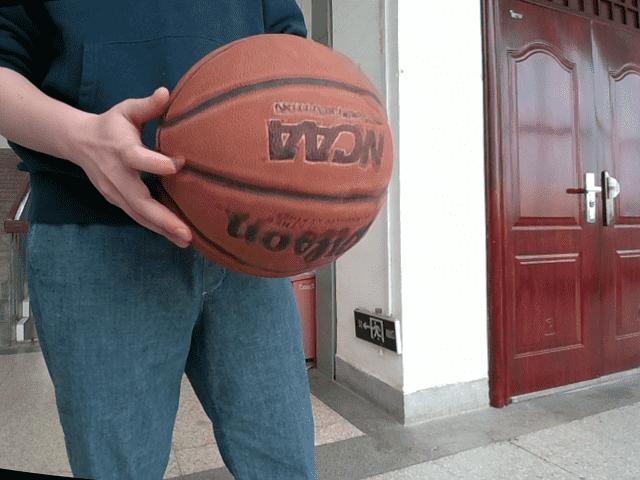}}\hfill
	   \subfloat[]{\includegraphics[width=\cimwidth\textwidth,trim={80 200 240 40},clip]{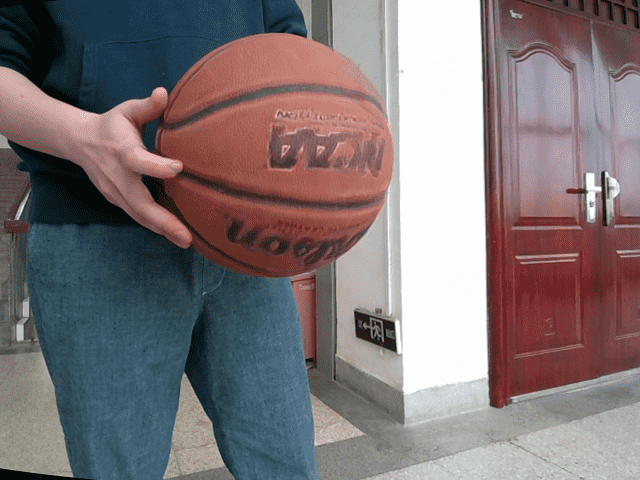}}\hfill
	   \subfloat[]{\includegraphics[width=\cimwidth\textwidth,trim={80 200 240 40},clip]{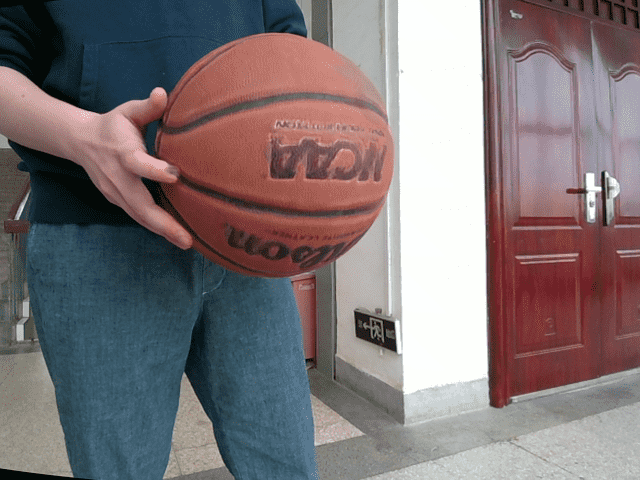}}\hfill
    \end{minipage}
    \vspace{-7mm}\begin{minipage}[]{\linewidth}
    		\centering
      \captionsetup[subfloat]{labelsep=none,format=plain,labelformat=empty}
      \begin{tikzpicture}[inner sep=0]
            \node [label={[label distance=0.42cm,text depth=-1ex,rotate=90]right: \textcolor{black}{\scriptsize {GT}}}] at (15,15) {};
            \end{tikzpicture}
      \subfloat[1st]{\includegraphics[width=\cimwidth\textwidth,trim={80 200 240 40},clip]{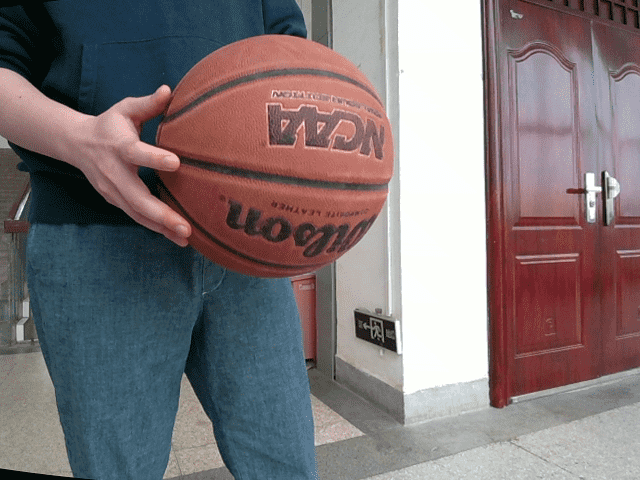}}\hfill
      \subfloat[2nd]{\includegraphics[width=\cimwidth\textwidth,trim={80 200 240 40},clip]{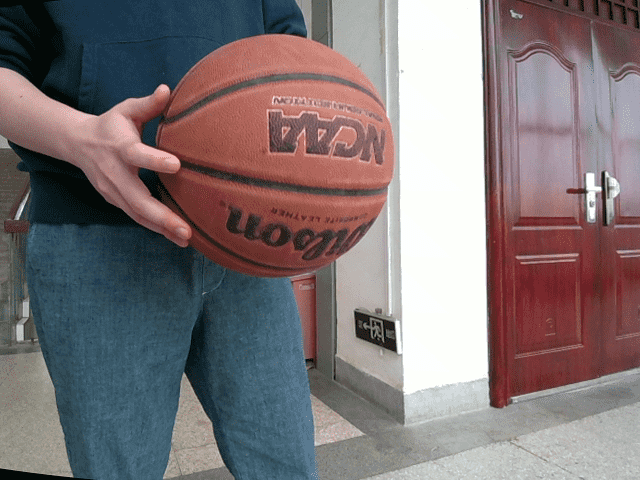}}\hfill
	   \subfloat[3rd]{\includegraphics[width=\cimwidth\textwidth,trim={80 200 240 40},clip]{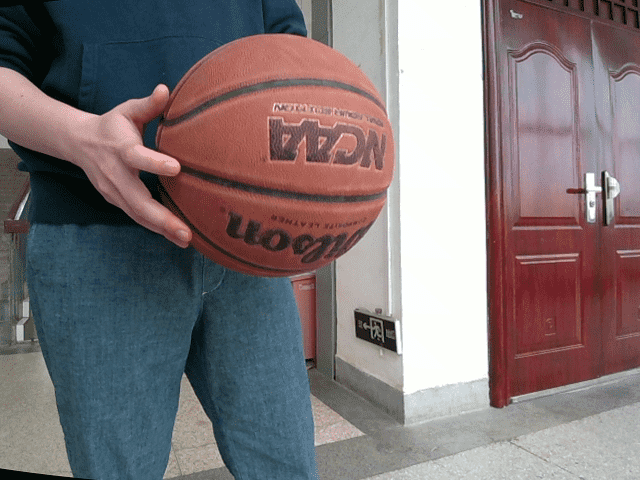}}\hfill
	   \subfloat[4th]{\includegraphics[width=\cimwidth\textwidth,trim={80 200 240 40},clip]{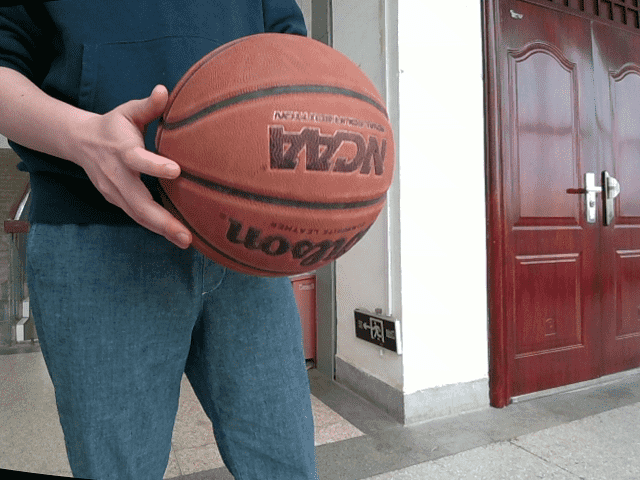}}\hfill
	   \subfloat[5th]{\includegraphics[width=\cimwidth\textwidth,trim={80 200 240 40},clip]{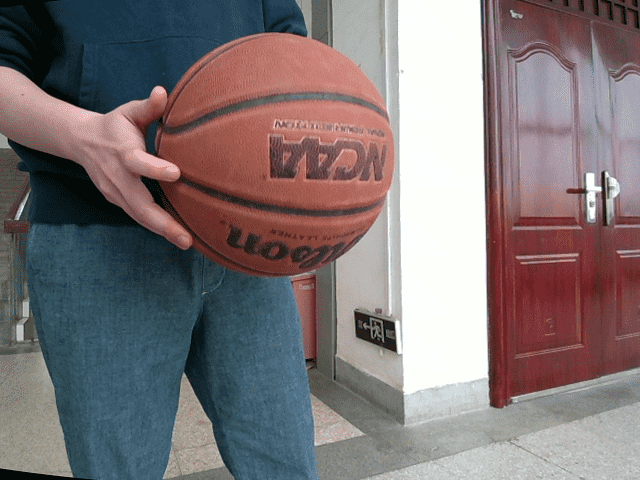}}\hfill
    
        \vspace{1mm}\vspace{7mm}

    \vspace{0.3mm}
    \end{minipage}

	\caption{Comparison of continuous frame reconstruction results by RIFE\cite{huang2022real}, Time Lens\cite{tulyakov2021time} and the proposed \mynetwork\ on \mydataset. Our \mynetwork\ achieves the best visual quality.}
	\label{Fig-seid1}
\end{figure}
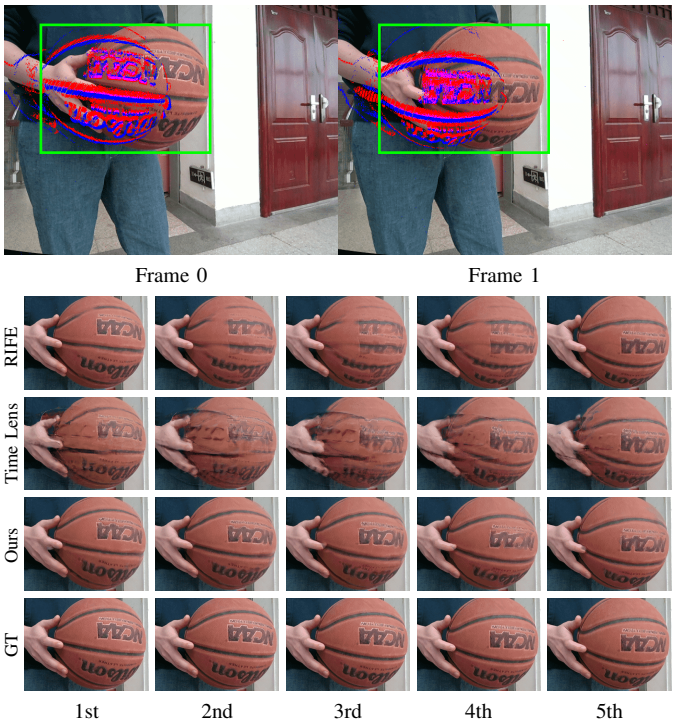
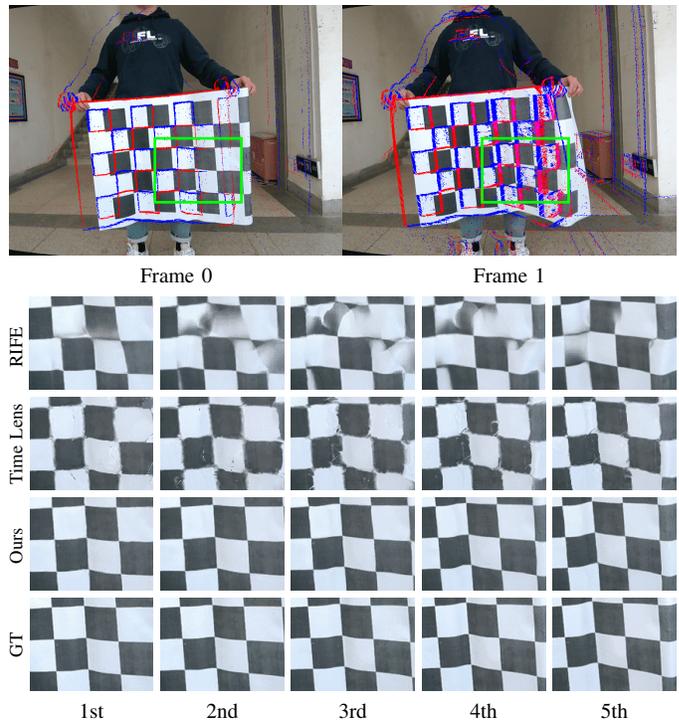
\def\cimwidth{0.185}
\def\zuoxia{(-2.5,-0.91)}
\def\youshang{(-1.35,-0.06)}
\def\zuox{(1.85,-0.91)}
\def\yous{(3,-0.06)}
\def\ssizz{1.2cm}
\def\ssmag{2}

\begin{figure}[ht]
\footnotesize
	\centering
 \begin{minipage}[t]{\linewidth}
    		\centering
      \captionsetup[subfloat]{labelsep=none,format=plain,labelformat=empty}
                \begin{tikzpicture}[spy using outlines={rectangle,green,magnification=\ssmag,size=\ssizz},inner sep=0]
				\node {
                \subfloat[Frame 0]{\includegraphics[width=0.5\linewidth]{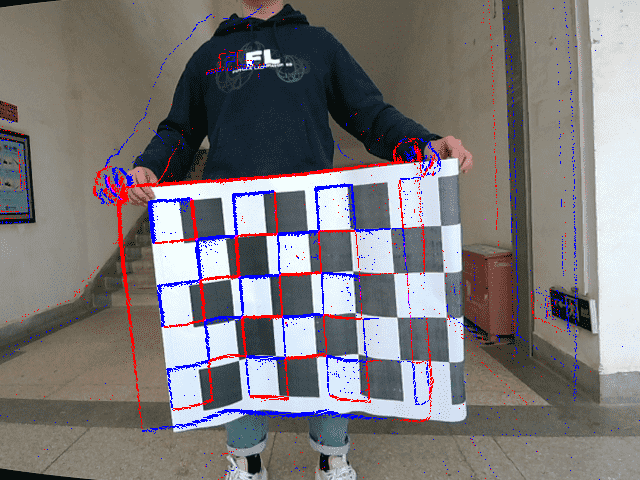}}
                \subfloat[Frame 1]{\includegraphics[width=0.5\linewidth]{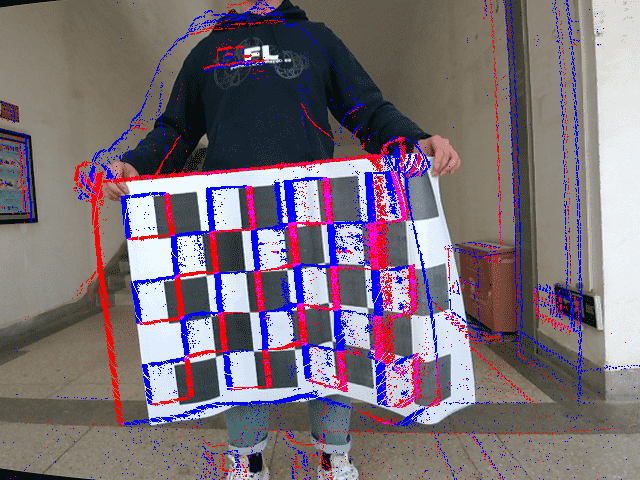}}
                };
				\draw[line width=1pt,green] \zuoxia rectangle \youshang;
                \draw[line width=1pt,green] \zuox rectangle \yous;
				\end{tikzpicture}
    	\end{minipage}
\vspace{-7mm}
\begin{minipage}[]{\linewidth}
            \vspace{-2.5mm}
    		\centering
      \captionsetup[subfloat]{labelsep=none,format=plain,labelformat=empty}
      \begin{tikzpicture}[inner sep=0]
            \node [label={[label distance=0.27cm,text depth=-1ex,rotate=90]right: \textcolor{black}{\scriptsize {RIFE}}}] at (15,15) {};
            \end{tikzpicture}
      \subfloat[]{\includegraphics[width=\cimwidth\textwidth,trim={280 100 200 260},clip]{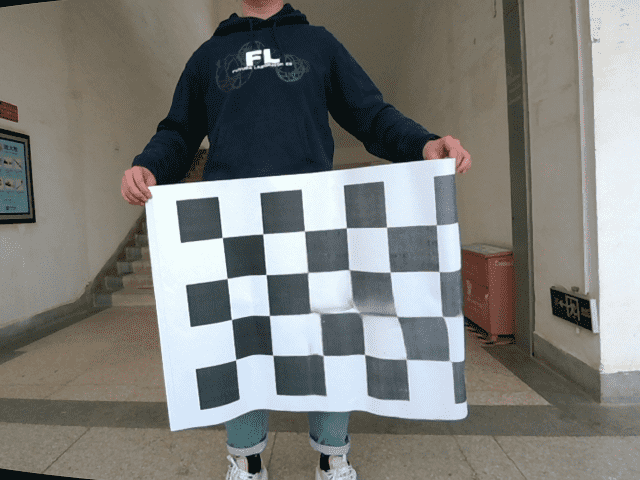}}\hfill
      \subfloat[]{\includegraphics[width=\cimwidth\textwidth,trim={280 100 200 260},clip]{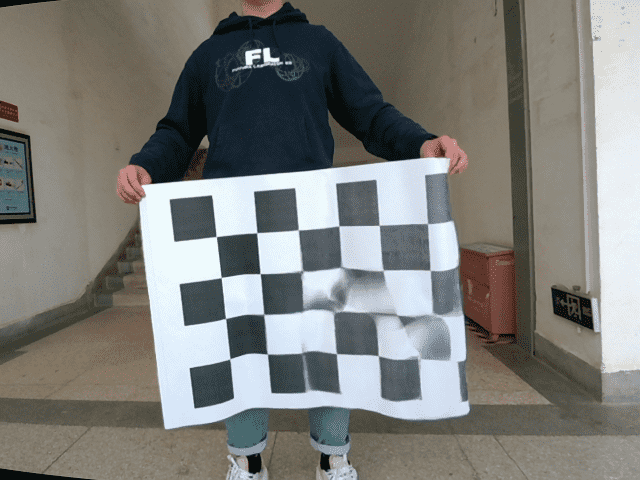}}\hfill
	   \subfloat[]{\includegraphics[width=\cimwidth\textwidth,trim={280 100 200 260},clip]{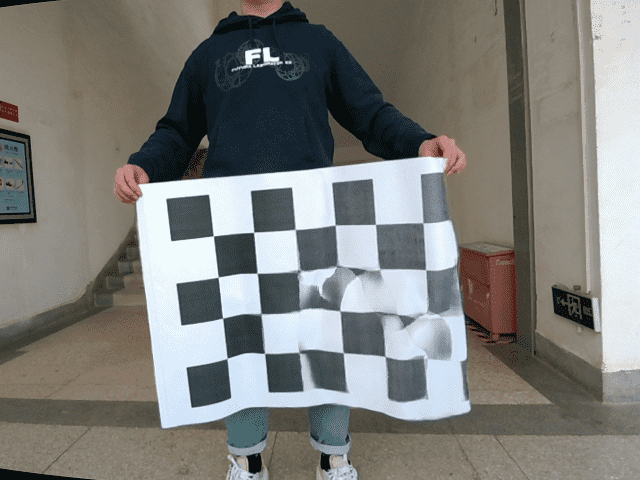}}\hfill
	   \subfloat[]{\includegraphics[width=\cimwidth\textwidth,trim={280 100 200 260},clip]{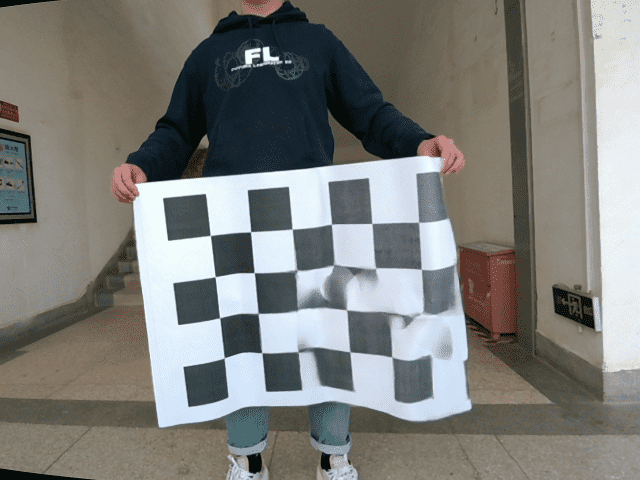}}\hfill
	   \subfloat[]{\includegraphics[width=\cimwidth\textwidth,trim={280 100 200 260},clip]{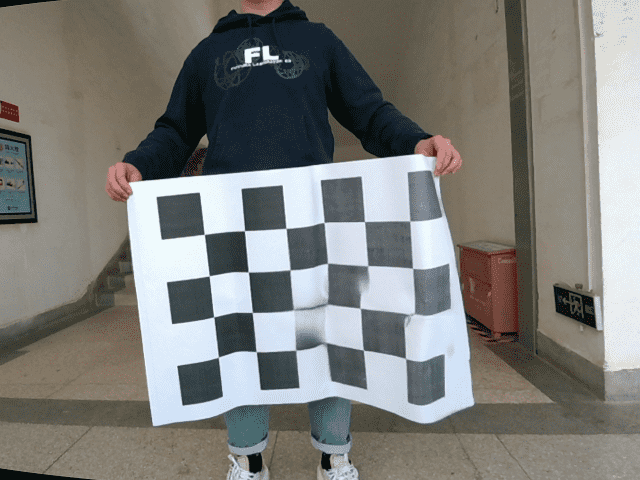}}\hfill
    \end{minipage}
    \vspace{-7mm}
    \begin{minipage}[]{\linewidth}
    		\centering
      \captionsetup[subfloat]{labelsep=none,format=plain,labelformat=empty}
      \begin{tikzpicture}[inner sep=0]
            \node [label={[label distance=0.05cm,text depth=-1ex,rotate=90]right: \textcolor{black}{\scriptsize {Time Lens}}}] at (15,15) {};
            \end{tikzpicture}
      \subfloat[]{\includegraphics[width=\cimwidth\textwidth,trim={280 100 200 260},clip]{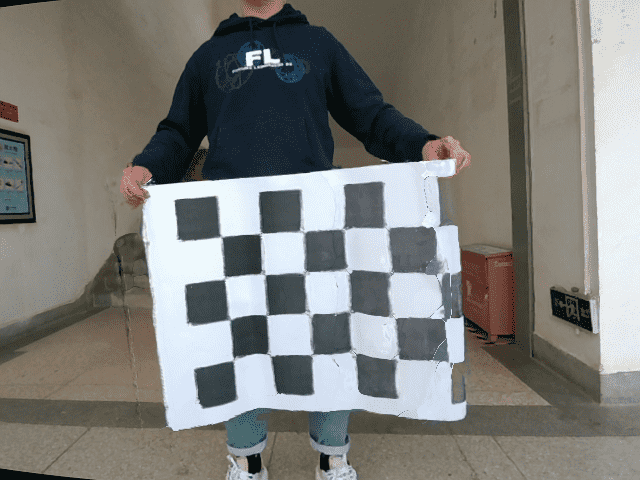}}\hfill
      \subfloat[]{\includegraphics[width=\cimwidth\textwidth,trim={280 100 200 260},clip]{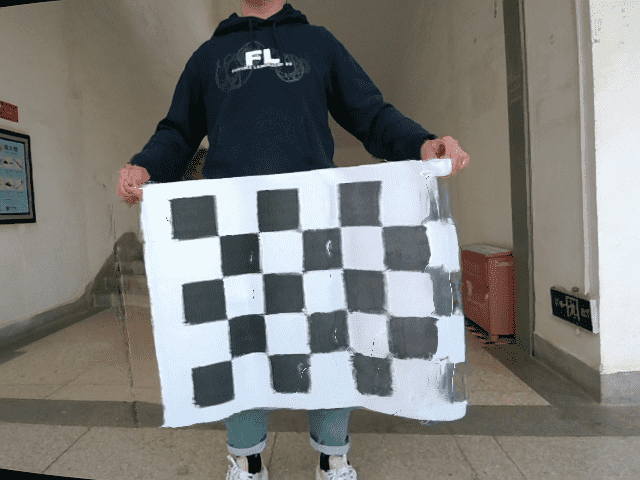}}\hfill
	   \subfloat[]{\includegraphics[width=\cimwidth\textwidth,trim={280 100 200 260},clip]{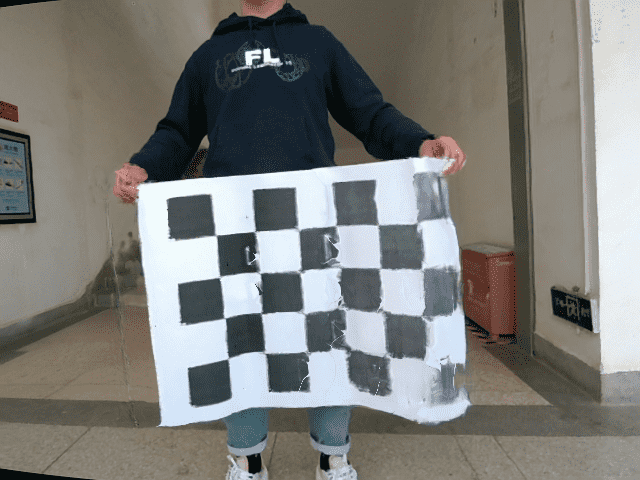}}\hfill
	   \subfloat[]{\includegraphics[width=\cimwidth\textwidth,trim={280 100 200 260},clip]{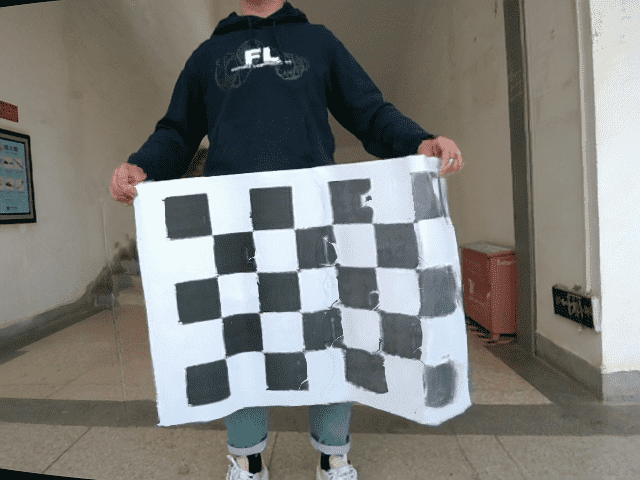}}\hfill
	   \subfloat[]{\includegraphics[width=\cimwidth\textwidth,trim={280 100 200 260},clip]{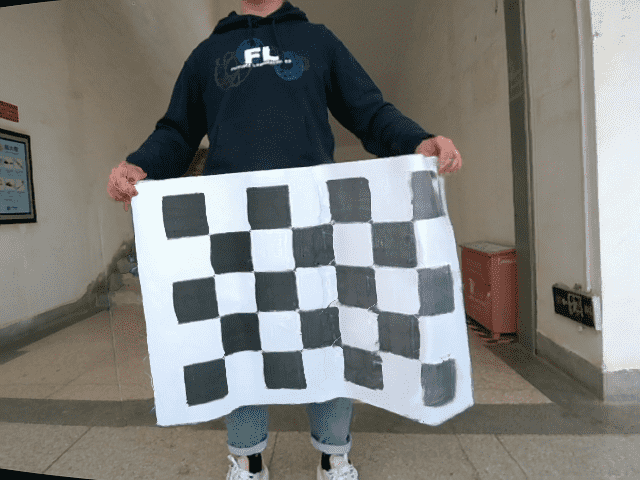}}\hfill
    \end{minipage}
    \vspace{-7mm}
    \begin{minipage}[]{\linewidth}
    		\centering
      \captionsetup[subfloat]{labelsep=none,format=plain,labelformat=empty}
      \begin{tikzpicture}[inner sep=0]
            \node [label={[label distance=0.33cm,text depth=-1ex,rotate=90]right: \textcolor{black}{\scriptsize {Ours}}}] at (15,15) {};
            \end{tikzpicture}
      \subfloat[]{\includegraphics[width=\cimwidth\textwidth,trim={280 100 200 260},clip]{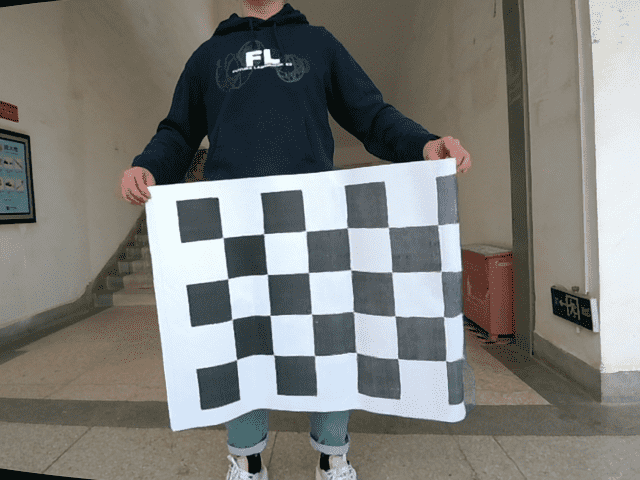}}\hfill
      \subfloat[]{\includegraphics[width=\cimwidth\textwidth,trim={280 100 200 260},clip]{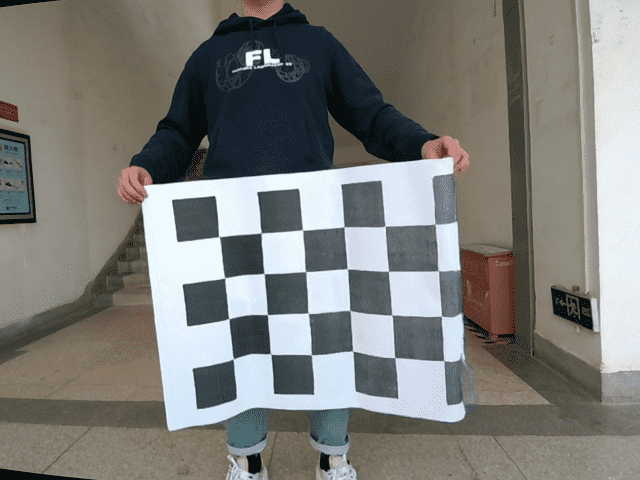}}\hfill
	   \subfloat[]{\includegraphics[width=\cimwidth\textwidth,trim={280 100 200 260},clip]{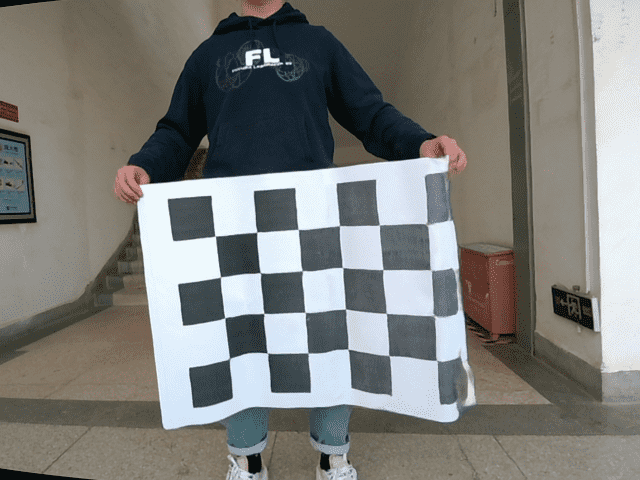}}\hfill
	   \subfloat[]{\includegraphics[width=\cimwidth\textwidth,trim={280 100 200 260},clip]{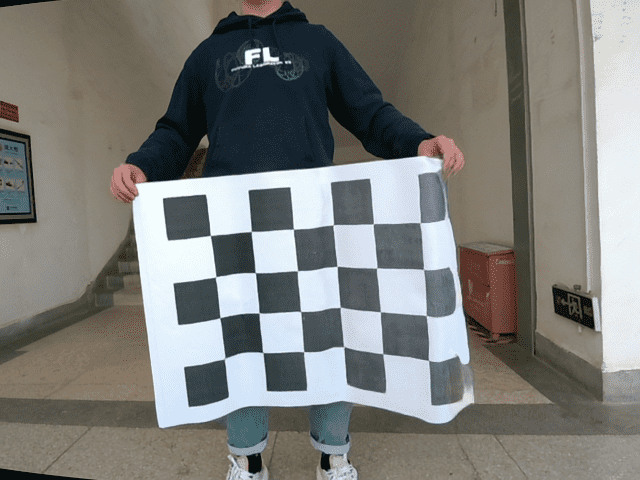}}\hfill
	   \subfloat[]{\includegraphics[width=\cimwidth\textwidth,trim={280 100 200 260},clip]{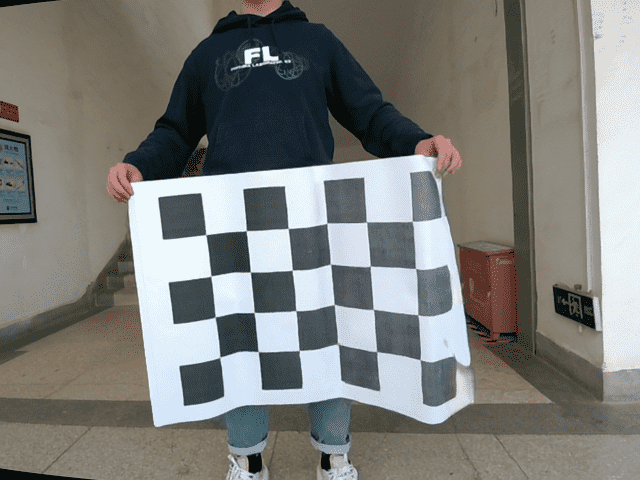}}\hfill
    \end{minipage}
    \vspace{-7mm}\begin{minipage}[]{\linewidth}
    		\centering
      \captionsetup[subfloat]{labelsep=none,format=plain,labelformat=empty}
      \begin{tikzpicture}[inner sep=0]
            \node [label={[label distance=0.42cm,text depth=-1ex,rotate=90]right: \textcolor{black}{\scriptsize {GT}}}] at (15,15) {};
            \end{tikzpicture}
      \subfloat[1st]{\includegraphics[width=\cimwidth\textwidth,trim={280 100 200 260},clip]{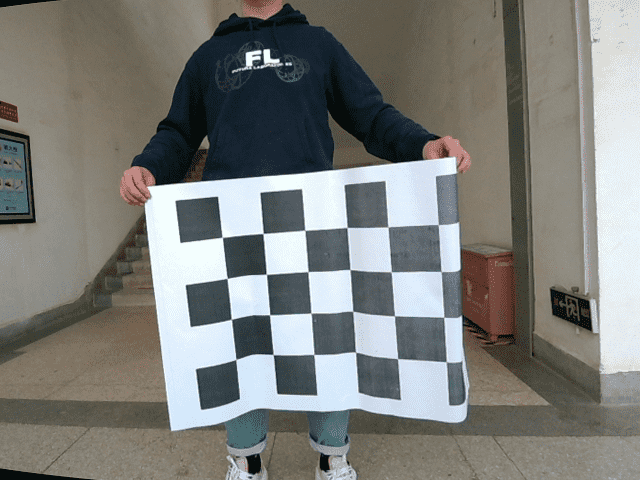}}\hfill
      \subfloat[2nd]{\includegraphics[width=\cimwidth\textwidth,trim={280 100 200 260},clip]{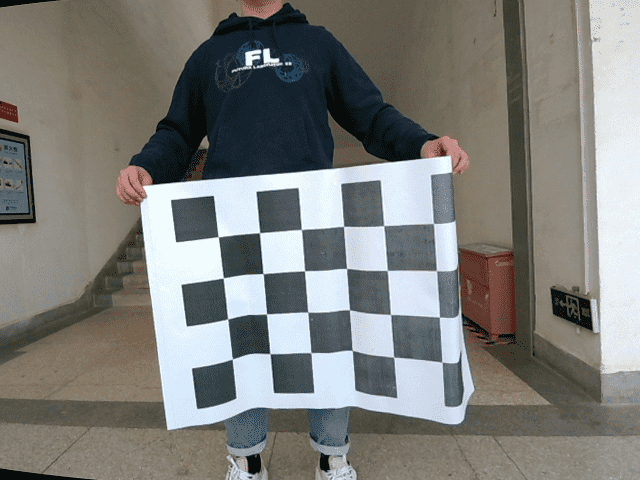}}\hfill
	   \subfloat[3rd]{\includegraphics[width=\cimwidth\textwidth,trim={280 100 200 260},clip]{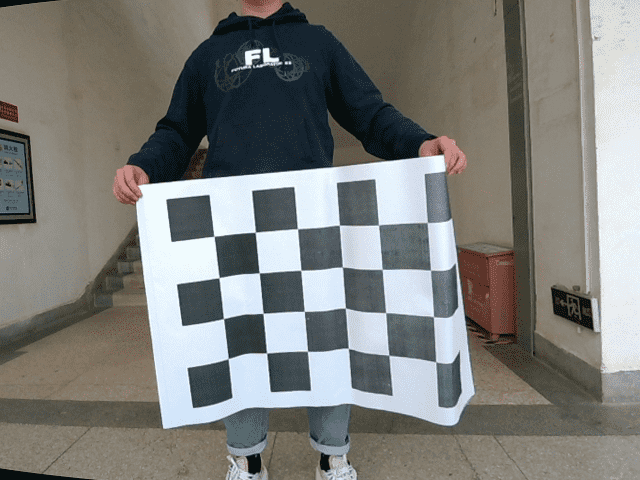}}\hfill
	   \subfloat[4th]{\includegraphics[width=\cimwidth\textwidth,trim={280 100 200 260},clip]{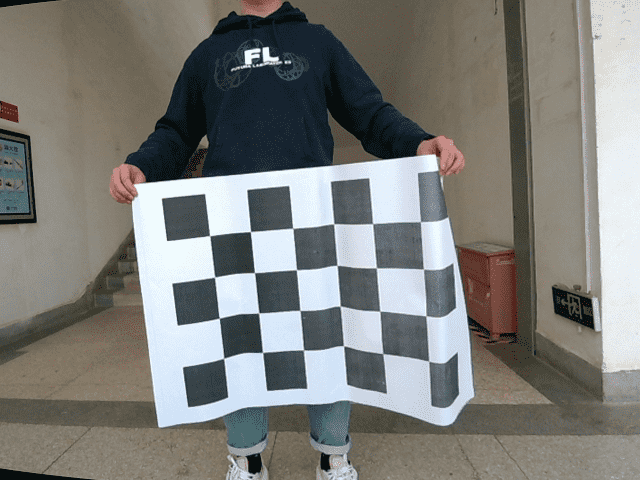}}\hfill
	   \subfloat[5th]{\includegraphics[width=\cimwidth\textwidth,trim={280 100 200 260},clip]{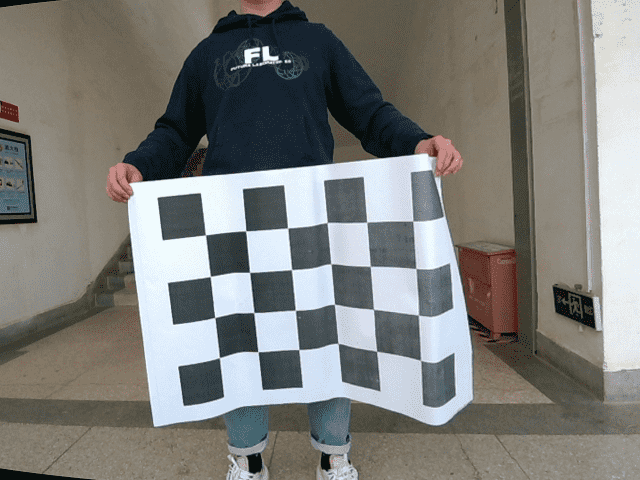}}\hfill
    
        \vspace{1mm}\vspace{7mm}

    \vspace{0.3mm}
    \end{minipage}

	\caption{Comparison of continuous frame reconstruction results by RIFE\cite{huang2022real}, Time Lens\cite{tulyakov2021time} and the proposed \mynetwork\ on \mydataset. Our \mynetwork\ achieves the best visual quality.}
	\label{Fig-seid2}
\end{figure}
\begin{figure*}[ht]
	\begin{minipage}[b]{0.50\linewidth}
		\centering
		\subfloat[MVSEC]{\includegraphics[width=0.50\linewidth]{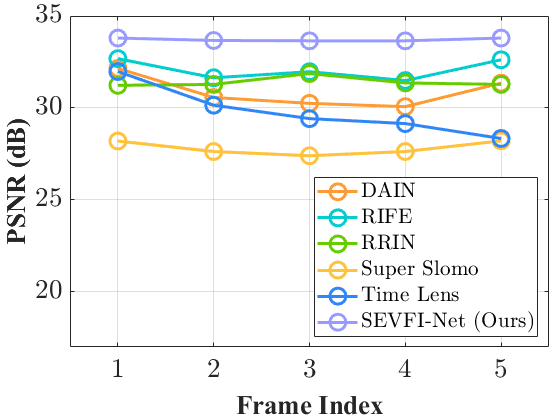}
        \includegraphics[width=0.50\linewidth]{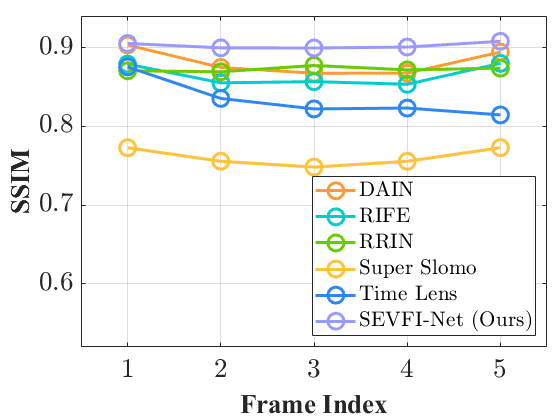}}
	\end{minipage}
    \begin{minipage}[b]{0.50\linewidth}
		\centering
		\subfloat[SEID (Ours)]{\includegraphics[width=0.50\linewidth]{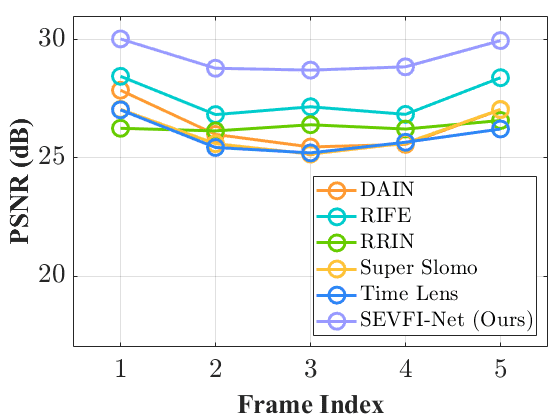}
        \includegraphics[width=0.50\linewidth]{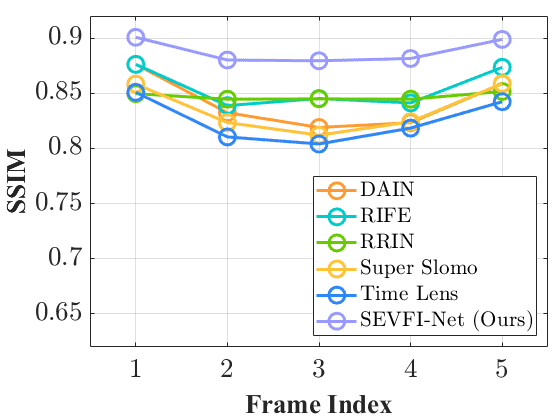}}
	\end{minipage}
	\caption{The ``Rope plot" shows the capability of continuous frame reconstruction. We perform frame prediction on the MVSEC\cite{zhu2018multivehicle} and our \mydataset\ datasets by generating 5 intermediate frames. We then calculate the average PSNR/SSIM at each frame index separately. The proposed \mynetwork\ achieves the highest metrics at each frame index.}
 \label{Fig-ropeplot}
\end{figure*}
As displayed in \cref{Fig-dsec,Fig-mvsec}, the interpolated frames of the F-VFI algorithms fail in dealing with fast and complex motions, leading to blurred or distorted edges in the results.
Although they can roughly reconstruct the scene, they estimate inter-frame motion using the input keyframes and are only good at handling simple linear motion. When the scene becomes complex or the motion is intense, the F-VFI algorithms often fail and produce unrealistic structures and textures. This can result in significant degradation of the interpolated frames.

Compared to F-VFI methods, existing E-VFI methods can estimate a more precise inter-frame motion model with the help of events. However, the cross-modal misalignment in stereo camera setups severely disturbs their performance and causes significant distortions and artifacts during the frame interpolation process as illustrated in \cref{Fig-dsec,Fig-mvsec}.
This is because existing E-VFI algorithms often rely on the assumption of pixel-level alignment between events and frames, which poses challenges in the context of stereo camera setups where capturing data from different modalities and aligning them effectively can be difficult.

In contrast, our proposed \mynetwork\ achieves superior visual performance.
Compared to existing methods, our method specifically addresses the challenges posed by the stereo camera setups and leverages the misalignment between different data.
Benefiting from the multiple encoders and feature aggregation modules, our \mynetwork\ can process events and frames separately, enabling us to effectively handle biases and discrepancies between the modalities and achieve data fusion, leading to the best visual performance.

\begin{figure}[ht]
 \centering
   \vspace{-3mm}
	\begin{minipage}[b]{\linewidth}
		\centering
		\subfloat[PSNR]{\includegraphics[width=0.50\linewidth]{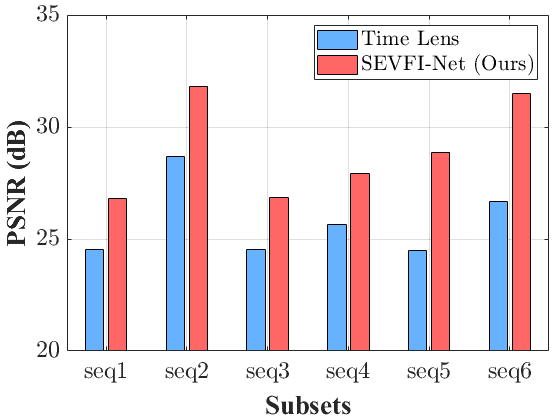}}
        \subfloat[Disparity \& Improvement]{\includegraphics[width=0.50\linewidth]{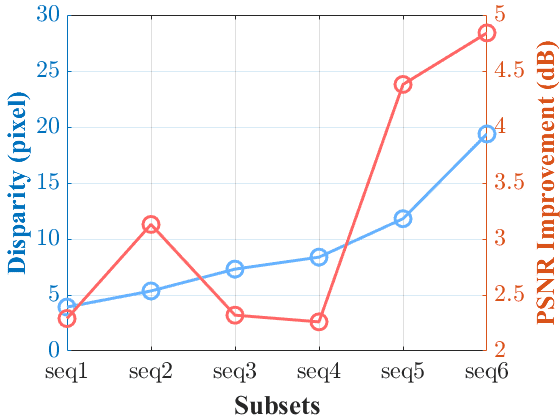}}
	\end{minipage}
   \caption{Illustration of the impact of disparity changes on the E-VFI algorithms. (a) It shows the PSNR of \mynetwork\ and Time Lens\cite{tulyakov2021time}. \mynetwork\ outperforms Time Lens over each subset; (b) It shows the average disparity of each subset along with the corresponding PSNR improvement between our \mynetwork\ and Time Lens. As the average disparity increases, the improvement in PSNR also shows an overall increasing trend. This indicates that our method is more robust in scenes with disparity variations.
   }
   \label{Fig-difference}
\end{figure}
\textbf{\textit{Quantitative comparisons.}}
Quantitative comparisons are given in \cref{tab:quantitative}.
We first notice that Time Lens, despite its utilization of events to estimate motion information and the integration of synthesis-based and flow-based approaches, exhibits significant performance degradation compared to F-VFI methods when dealing with real-world stereo datasets. This degradation can be attributed to the absence of effective data alignment in Time Lens.
We then compare our \mynetwork\ with Time Lens. \mynetwork\ achieves significant improvements over Time Lens, with up to 3.98 dB and 0.0712 enhancements in terms of PSNR and SSIM, respectively.
As shown in~\cref{Fig-dsec,Fig-mvsec}, \mynetwork\ showcases superior reconstruction results with fewer artifacts and distortions compared to Time Lens.
This demonstrates the effectiveness of our method in mitigating distractions caused by the cross-modal disparity.
Moreover, we compare the proposed \mynetwork\ with other F-VFI methods not affected by the parallax. The metrics illustrate that the \mynetwork\ outperforms these state-of-the-art F-VFI methods, which means \mynetwork\ removes the artifacts caused by the cross-modal parallax, achieving precise non-linear motion estimation from misaligned events and frames.

\subsubsection{Results of Multi-frame Reconstruction}
In this part, we compare the performance of our \mynetwork\ with the existing methods in multi-frame reconstruction. We provide qualitative and quantitative results to show the advantages of our method.

\mynetwork\ exhibits unique advantages when it comes to handling non-linear motion, as shown in \cref{Fig-seid1,Fig-seid2}, giving the best continuous multi-frame reconstruction and exhibiting excellent ability in reconstructing dynamic scenes, with the motion patterns closely resembling the ground truth. We also conduct experiments of sequence interpolation on MVSEC and our \mydataset\ and calculate the average PSNR and SSIM at each frame index separately as shown in \cref{Fig-ropeplot}, the proposed \mynetwork\ achieves the highest metrics at each frame index.
This is due to the high temporal resolution of events enables us to effectively model non-linear and fast-moving motions and generate interpolated frames that are visually appealing while preserving the natural motion characteristics of the scene.

\subsubsection{Results of Changing Disparity}
In addition, we also evaluate the impact of disparity on E-VFI results as illustrated in \cref{Fig-difference}. We perform experiments on our \mydataset\ dataset with the 5-frame skip for each subset, seq1-seq6 in the subfigures represent [``Pedestrians", ``Basketball", ``Cars", ``Square", ``Checkerboard", ``Indoor"] in our \mydataset\ respectively.
And we calculate the PSNR improvement of our method compared to Time Lens under different disparity conditions, to demonstrate its stability and robustness when dealing with disparity changes. The data misalignment leads to performance degradation for Time Lens, especially when the average disparity value increases. In contrast, our algorithm maintains a relatively stable performance across different disparity values.
Therefore, as the average disparity increases, the improvement in PSNR also shows an overall increasing trend, indicating that it is more stable and robust in scenes with disparity variations.

\hspace*{\fill}

Overall, our method not only addresses the challenges of the stereo camera setup and misaligned modalities but also achieves the best results in terms of frame interpolation and maintains robustness even in the presence of large depth variations.

\subsection{Comparisons of Stereo Matching}\label{compare_SM}
In this subsection, we evaluate the stereo-matching performance of our method with existing cross-modal stereo-matching methods. The qualitative and quantitative experiments demonstrate the superiority of our method.

As formulated in~\cref{eq:alignnet}, our \mynetwork\ is capable of not only producing interpolated frames but also estimating the corresponding disparity results. To validate its effectiveness in disparity estimation, we conduct experiments on the DSEC\cite{gehrig2021dsec} with a 3-frame skip setting and on the MVSEC\cite{zhu2018multivehicle} and \mydataset\ with a 5-frame skip setting.

\begin{table*}[th]
\centering
\small
\caption{Quantitative comparisons of the quality of disparities on DSEC~\cite{gehrig2021dsec}, MVSEC\cite{zhu2018multivehicle} and \mydataset\ datasets. We compute averaged end-point-error (EPE) and 1-pixel, 2-pixel, 3-pixel error.}
\begin{tabular}{lcccccccc}
\toprule[1.2pt] 
{Methods} && EPE (px)$\downarrow$ && $>1$px ($\%$)$\downarrow$ && $>2$px ($\%$)$\downarrow$ && $>3$px ($\%$)$\downarrow$ \\\hline
\textbf{\textit{DSEC}} &&&&&&&&\\\hline
HSM\cite{kim2022real} + RIFE\cite{huang2022real} & & \underline{12.83} && \underline{77.39} && \underline{59.80} && \underline{49.49} \\
HSM\cite{kim2022real} + Time Lens\cite{tulyakov2021time} & & 13.94 && 80.40 && 63.96 && 53.77 \\
SSIE\cite{gu2022self} + RIFE\cite{huang2022real} & & 16.42 && 96.79 && 93.59 && 90.40 \\
SSIE\cite{gu2022self} + Time Lens\cite{tulyakov2021time} & & 16.44 && 96.79 && 93.59 && 90.40 \\
\mynetwork\ (Ours) & & \textbf{4.10} && \textbf{65.27} && \textbf{45.98} && \textbf{35.69} \\\hline
\textbf{\textit{MVSEC}} &&&&&&&&\\\hline
HSM\cite{kim2022real} + RIFE\cite{huang2022real} & & 18.45 && \underline{74.05} && \underline{61.35} && \underline{56.09} \\
HSM\cite{kim2022real} + Time Lens\cite{tulyakov2021time} & & 19.54 && 80.96 && 68.25 && 61.63 \\
SSIE\cite{gu2022self} + RIFE\cite{huang2022real} & & \underline{13.05} && 94.59 && 89.35 && 84.32 \\ 
SSIE\cite{gu2022self} + Time Lens\cite{tulyakov2021time} & & 13.12 && 94.63 && 89.42 && 84.42\\
\mynetwork\ (Ours) & & \textbf{2.62} && \textbf{55.38} && \textbf{36.09} && \textbf{24.22} \\\hline
\textbf{\textit{\mydataset}} &&&&&&&&\\\hline 
HSM\cite{kim2022real} + RIFE\cite{huang2022real} & & 21.96 && 91.80 && 85.64 && 79.60\\
HSM\cite{kim2022real} + Time Lens\cite{tulyakov2021time} & & 21.97 && \underline{91.67} && \underline{85.52} && \underline{79.49} \\
SSIE\cite{gu2022self} + RIFE\cite{huang2022real} & & 14.00 && 93.42 && 87.35 && 81.62 \\
SSIE\cite{gu2022self} + Time Lens\cite{tulyakov2021time} & & \underline{13.94} && 93.27 && 87.11 && 81.37 \\
\mynetwork\ (Ours) & & \textbf{4.32} && \textbf{65.76} && \textbf{52.34} && \textbf{42.47} \\
\toprule[1.2pt] 
\end{tabular}
\label{tab:disp}
\end{table*}
\def\imwidth{0.12}
\begin{figure*}[ht]
\footnotesize
\centering
\captionsetup[subfloat]{labelsep=none,format=plain,labelformat=empty}
	\begin{minipage}[]{\linewidth}
		\flushright
  \begin{tikzpicture}[inner sep=0]
            \node [label={[label distance=0.45cm,text depth=0ex,rotate=90]right: \textcolor{black}{\scriptsize {DSEC}}}] at (15,15) {};
            \end{tikzpicture}
		\subfloat[]{\includegraphics[width=\imwidth\linewidth]{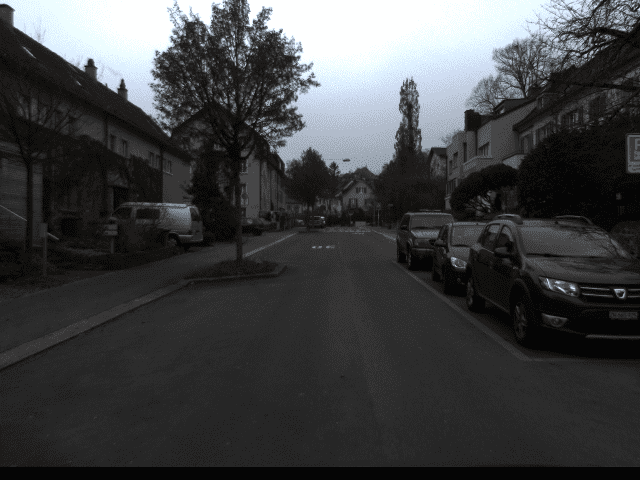}}\hfill
        \subfloat[]{\includegraphics[width=\imwidth\linewidth]{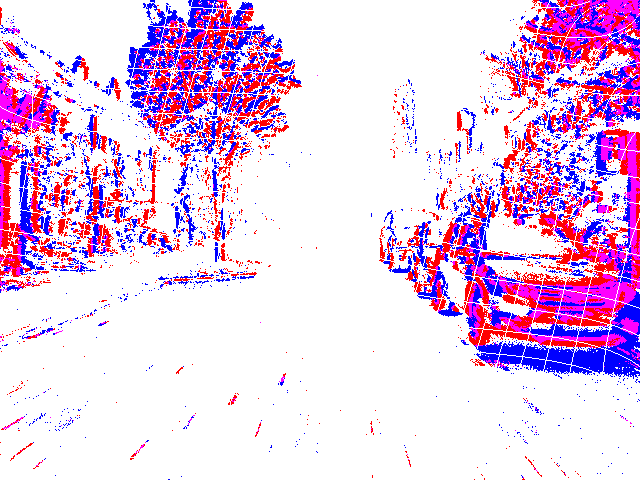}}\hfill
        \subfloat[]{\includegraphics[width=\imwidth\linewidth]{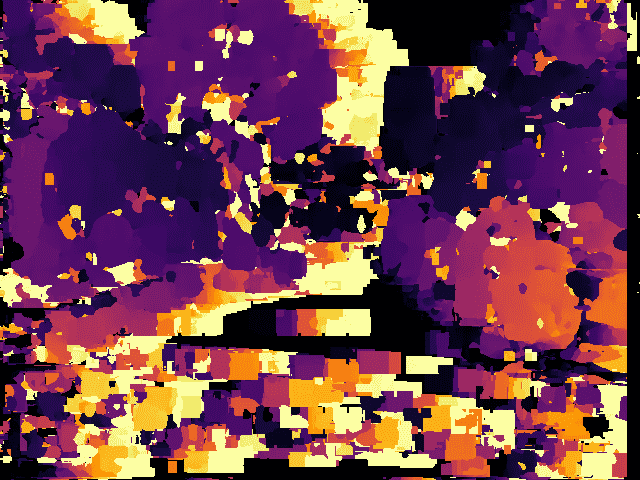}}\hfill
        \subfloat[]{\includegraphics[width=\imwidth\linewidth]{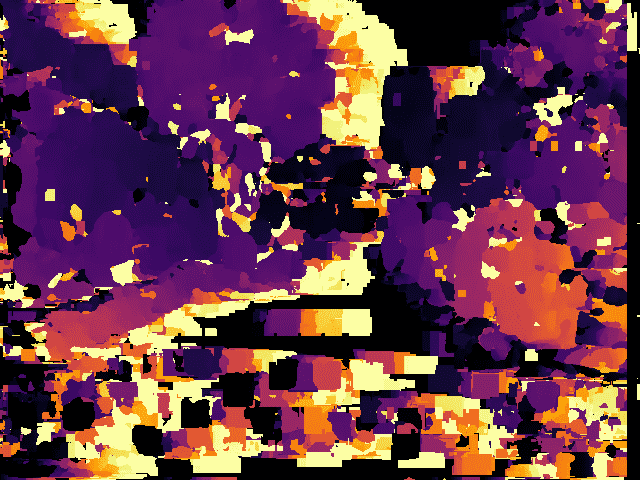}}\hfill
        \subfloat[]{\includegraphics[width=\imwidth\linewidth]{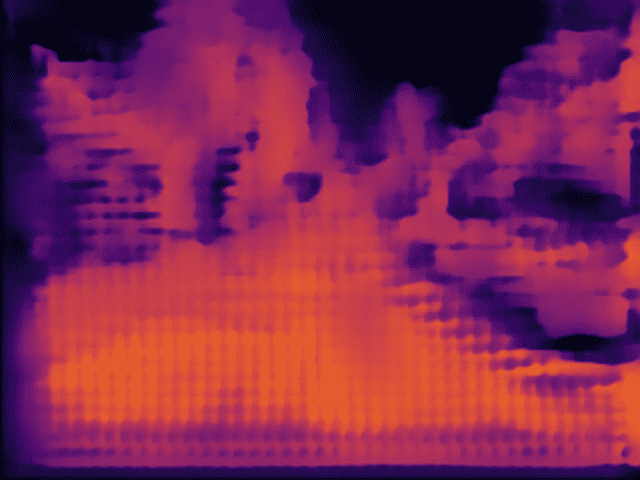}}\hfill
        \subfloat[]{\includegraphics[width=\imwidth\linewidth]{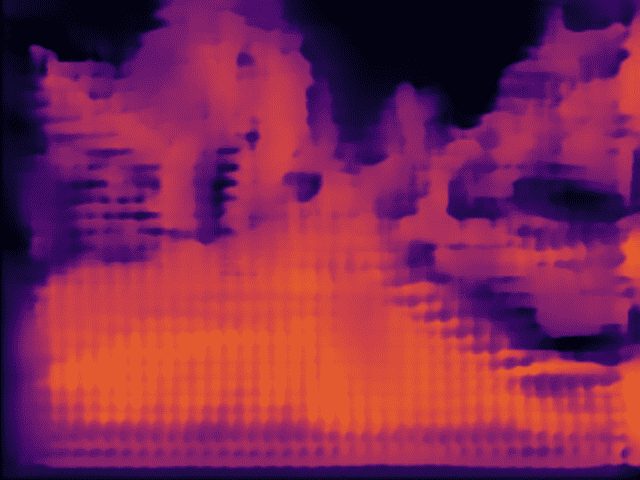}}\hfill
        \subfloat[]{\includegraphics[width=\imwidth\linewidth]{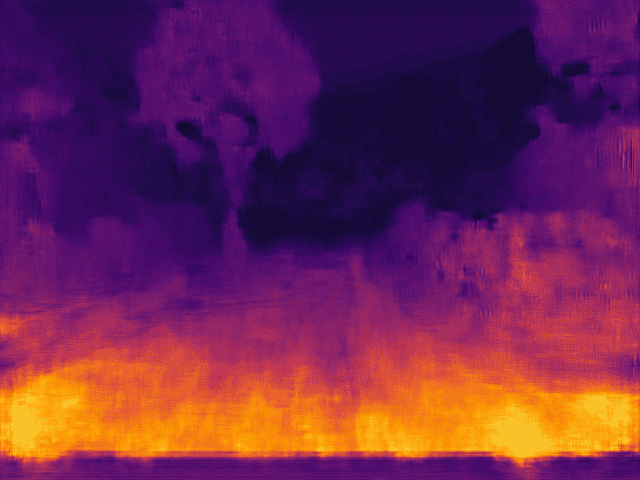}}\hfill
        \subfloat[]{\includegraphics[width=\imwidth\linewidth]{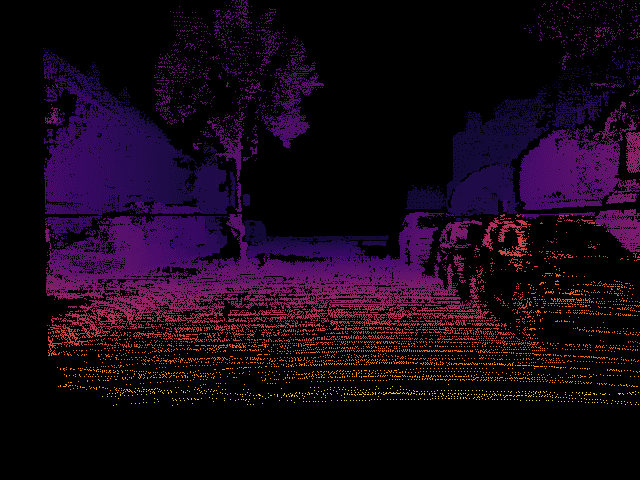}}
        \vspace{-7.5mm}
	\end{minipage}\\ \hfill
	\begin{minipage}[]{\linewidth}
		\flushright
  \begin{tikzpicture}[inner sep=0]
            \node [label={[label distance=0.41cm,text depth=0ex,rotate=90]right: \textcolor{black}{\scriptsize {MVSEC}}}] at (15,15) {};
            \end{tikzpicture}
		\subfloat[]{\includegraphics[width=\imwidth\linewidth]{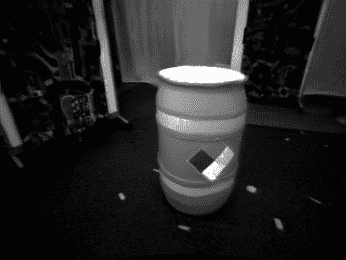}}\hfill
        \subfloat[]{\includegraphics[width=\imwidth\linewidth]{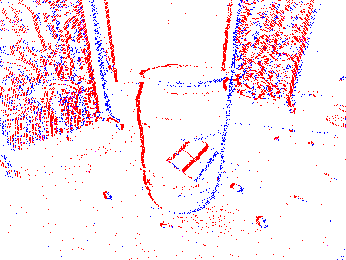}}\hfill
        \subfloat[]{\includegraphics[width=\imwidth\linewidth]{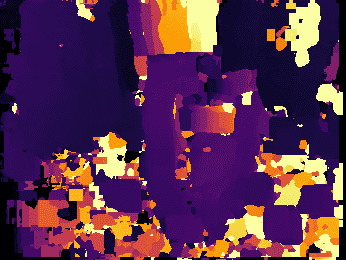}}\hfill
        \subfloat[]{\includegraphics[width=\imwidth\linewidth]{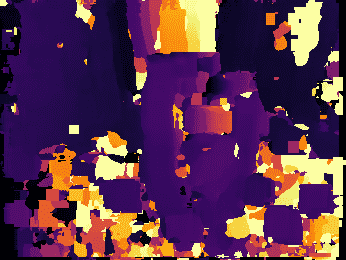}}\hfill
        \subfloat[]{\includegraphics[width=\imwidth\linewidth]{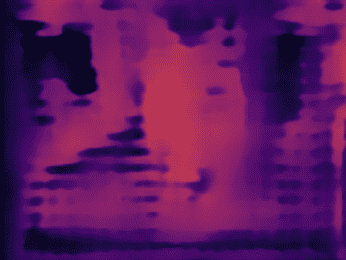}}\hfill
        \subfloat[]{\includegraphics[width=\imwidth\linewidth]{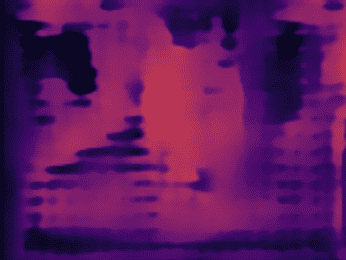}}\hfill
        \subfloat[]{\includegraphics[width=\imwidth\linewidth]{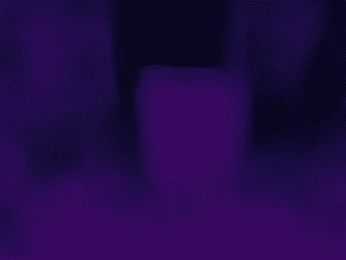}}\hfill
        \subfloat[]{\includegraphics[width=\imwidth\linewidth]{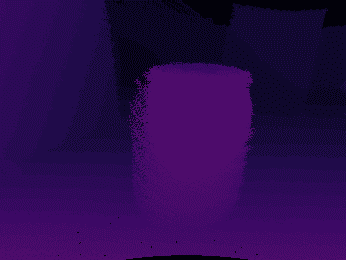}}
        \vspace{-7.5mm}
	\end{minipage}\\ \hfill
    \begin{minipage}[]{\linewidth}
		\flushright
  \begin{tikzpicture}[inner sep=0]
            \node [label={[label distance=0.50cm,text depth=0ex,rotate=90]right: \textcolor{black}{\scriptsize {SEID}}}] at (15,15) {};
            \end{tikzpicture}
		\subfloat[Image]{\includegraphics[width=\imwidth\linewidth]{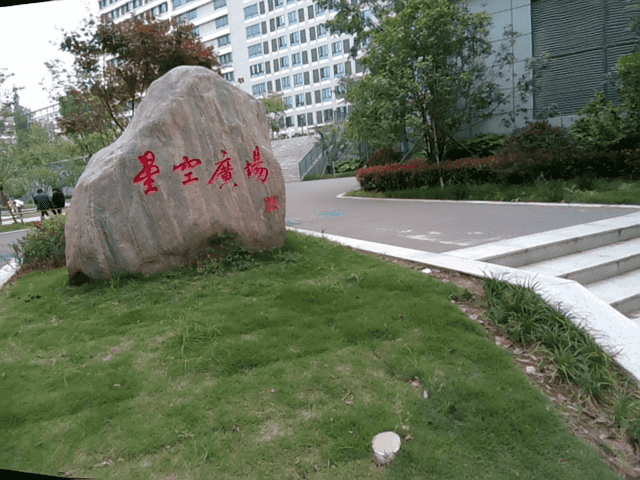}}\hfill
        \subfloat[Event]{\includegraphics[width=\imwidth\linewidth]{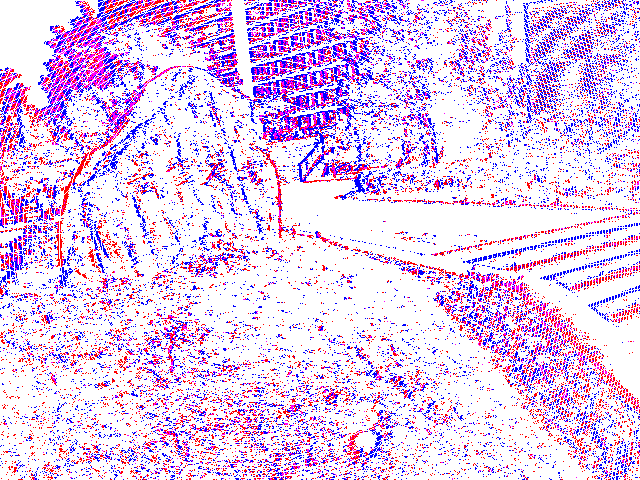}}\hfill
        \subfloat[HSM + RIFE]{\includegraphics[width=\imwidth\linewidth]{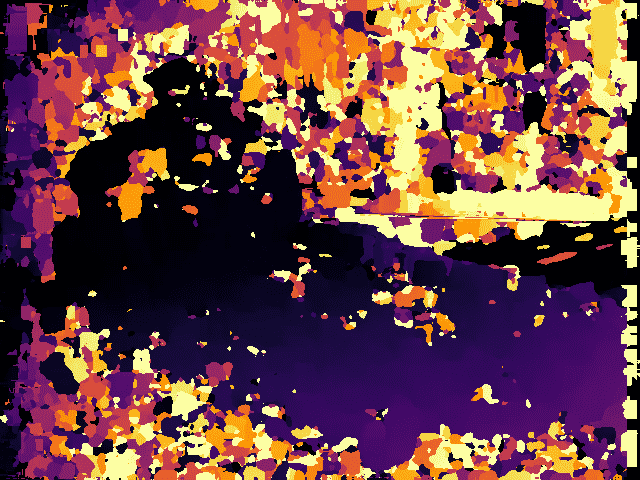}}\hfill
        \subfloat[HSM + Time Lens]{\includegraphics[width=\imwidth\linewidth]{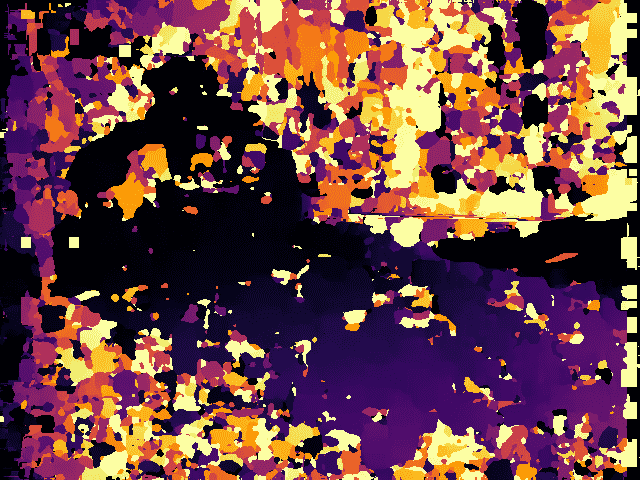}}\hfill
        \subfloat[SSIE + RIFE]{\includegraphics[width=\imwidth\linewidth]{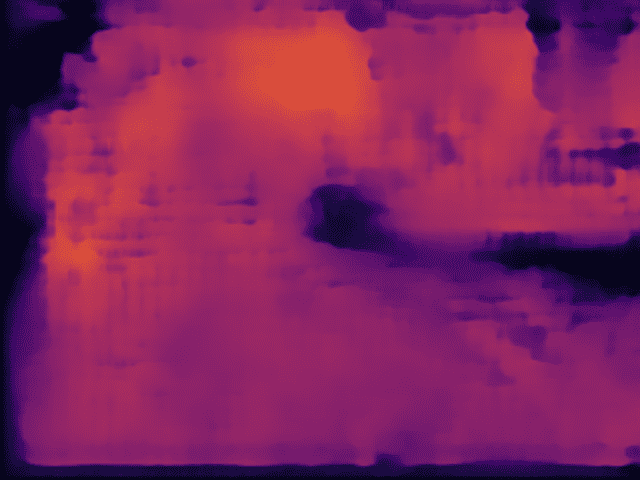}}\hfill
        \subfloat[SSIE + Time Lens]{\includegraphics[width=\imwidth\linewidth]{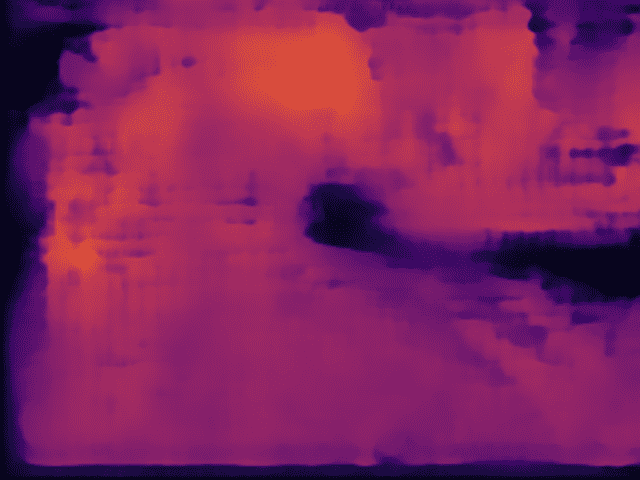}}\hfill
        \subfloat[\mynetwork\ (Ours)]{\includegraphics[width=\imwidth\linewidth]{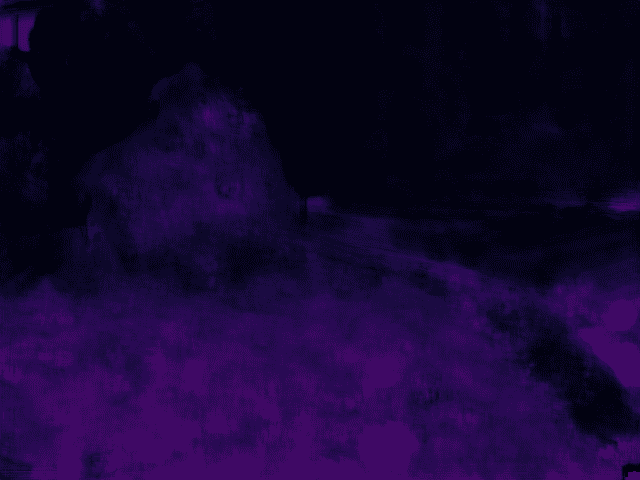}}\hfill
        \subfloat[GT]{\includegraphics[width=\imwidth\linewidth]{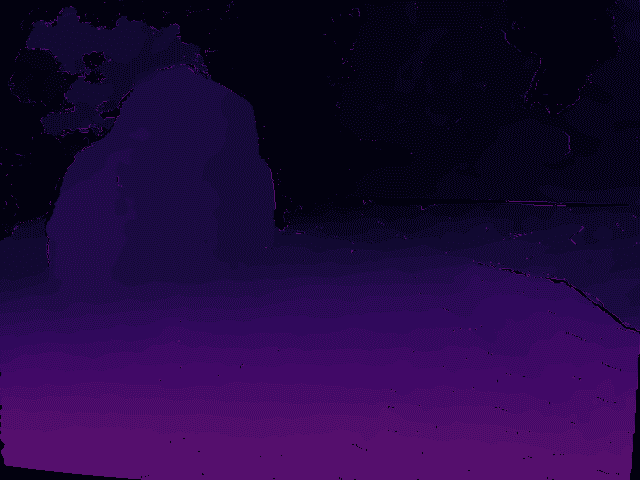}}\hfill
	\end{minipage}
	\caption{Illustration of the qualitative results of disparities on DSEC\cite{gehrig2021dsec} (1st row) MVSEC\cite{zhu2018multivehicle} (2nd row) and \mydataset\ (3rd row). The first column is the interpolated frame by our \mynetwork, the middle five columns are the disparity results predicted by different methods, and the last column is the ground truth.}
 \label{Fig-disparity}
\end{figure*}
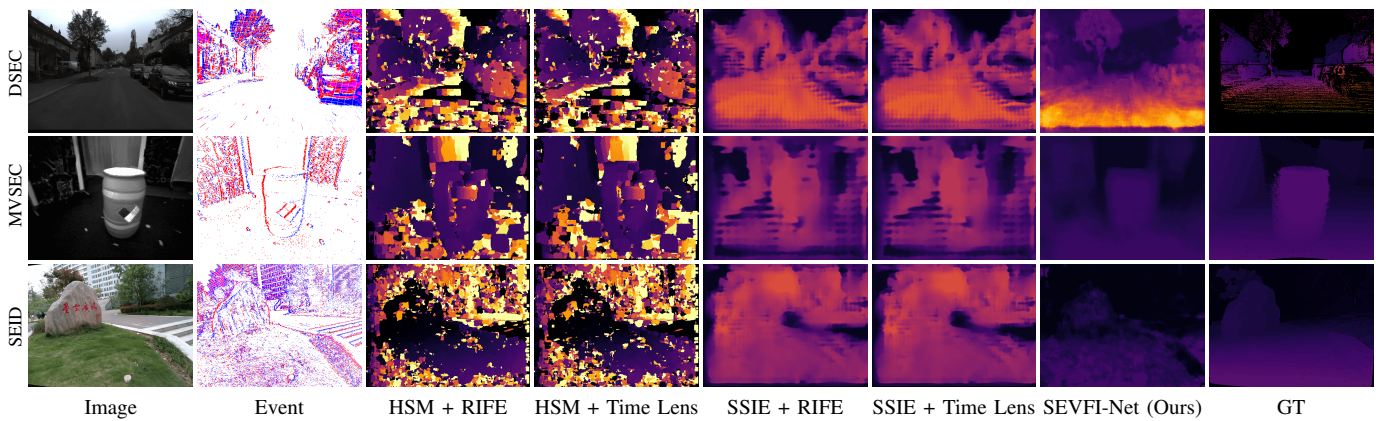
To the best of our knowledge, our \mynetwork\ is the first work that combines stereo matching with video frame interpolation.
In our experimental setup, we only have two keyframes and the event stream between them, which cannot be directly applied to existing event-frame stereo-matching algorithms. Therefore, we first choose two representative VFI algorithms, \ie, RIFE\cite{huang2022real} and Time Lens\cite{tulyakov2021time} to generate intermediate frames and then input them along with the event stream into two existing stereo matching algorithms, \ie, HSM\cite{kim2022real} and SSIE\cite{gu2022self}, to compare their performance with our method.

We utilize the widely-used metrics end-point error (EPE) and 1-pixel, 2-pixel, 3-pixel error, where EPE is the mean disparity error in pixels and the $c$-pixel error is the average percentage of the pixel whose EPE is larger than $c$ pixels.
Compared to the other methods, ours achieves the best visualization performance and gains an $86.6\%$ decrease in EPE and $71.3\%$ decrease in the 3-pixel error as shown in \cref{tab:disp,Fig-disparity}.
This is because the combination of the VFI and stereo-matching methods can be influenced by the quality of frame interpolation. The accuracy and smoothness of the interpolated frames play a crucial role in achieving high-quality results. Additionally, existing cross-modal stereo-matching algorithms are typically designed based on the assumption of static scenes and moving cameras, and may struggle with complex scenes, leading to significant performance degradation.

In contrast, our \mynetwork\ excels at matching events and images effectively, allowing for the recovery of accurate depth information in complex scenes. Importantly, our method overcomes the limitations of traditional methods constrained by frame rates by outputting disparities through interpolation. This approach enables a more flexible and continuous representation of depth information, which greatly enhances the overall performance of the system. The metrics shown in \cref{tab:disp} demonstrate that our \mynetwork\ outperforms existing methods by a large margin.

\begin{table*}[th]
\centering
\small
\caption{Quantitative comparisons of the outputs of each subnetwork of the proposed \mynetwork\ on the DSEC\cite{gehrig2021dsec}, MVSEC\cite{zhu2018multivehicle} and our \mydataset\ datasets.}
\begin{tabular}{cccccccccc}
\toprule[1.2pt] 
\multirow{2}*{Results} && \multicolumn{2}{c}{DSEC}                                &&  \multicolumn{2}{c}{MVSEC} && \multicolumn{2}{c}{\mydataset\ (Ours)}
 \\\cline{3-4} \cline{6-7} \cline{9-10}
&& PSNR $\uparrow$ & SSIM $\uparrow$ && PSNR $\uparrow$ & SSIM $\uparrow$ && PSNR $\uparrow$ & SSIM $\uparrow$\\ \hline
Flow-based $W_{0 \rightarrow t}^{\Omega_f}$  && 28.24 & 0.8464 && 32.21 & 0.8691 && 27.16 & 0.8426\\
Flow-based $W_{1 \rightarrow t}^{\Omega_f}$  && 27.77 & 0.8428 && 31.86 & 0.8587 && 26.87 & 0.8333\\
Syn-based $S_t^{\Omega_f}$ && 26.04 & 0.7988 && 28.51 & 0.8008 && 23.54 & 0.7345\\
Fused result $\tilde{I}_t^{\Omega_f}$ && \underline{29.75} & \underline{0.8722} && \underline{32.81} & \underline{0.8834} && \underline{28.73} & \underline{0.8711}\\
Refined result $\hat{I}_t^{\Omega_f}$ && \textbf{30.29} & \textbf{0.8834} && \textbf{33.79} & \textbf{0.9053} && \textbf{29.27} & \textbf{0.8881}\\
\toprule[1.2pt] 
\end{tabular}
\label{tab:ablation}
\end{table*}
\def\cimwidth{0.16}
\def\zuoxia{(2,-0.28)}
\def\youshang{(2.64,0.72)}

\begin{figure*}[ht]
\footnotesize
	\centering

 \begin{minipage}[]{0.49\linewidth}
    		\centering
      \captionsetup[subfloat]{labelsep=none,format=plain,labelformat=empty}
                \begin{tikzpicture}[spy using outlines={rectangle,green,magnification=\ssmag,size=\ssizz},inner sep=0]
				\node {
                \subfloat[(a) An outdoor scene\label{abl}]{\includegraphics[width=\linewidth]{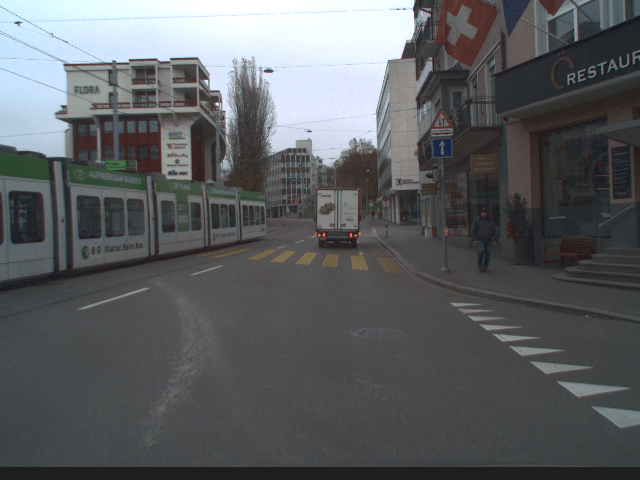}}
                };
				\draw[line width=1pt,green] \zuoxia rectangle \youshang;
				\end{tikzpicture}
    	\end{minipage}
 \begin{minipage}[]{0.49\linewidth}
      \flushright
      \captionsetup[subfloat]{labelsep=none,format=plain,labelformat=empty}
      \begin{tikzpicture}[inner sep=0]
            \node [label={[label distance=0.85cm,text depth=0.5ex,rotate=90]right: \textcolor{black}{\scriptsize {1st}}}] at (15,15) {};
            \end{tikzpicture}
      \subfloat[\label{abl:a}]{\includegraphics[width=\cimwidth\textwidth,trim={465 205 130 205},clip]{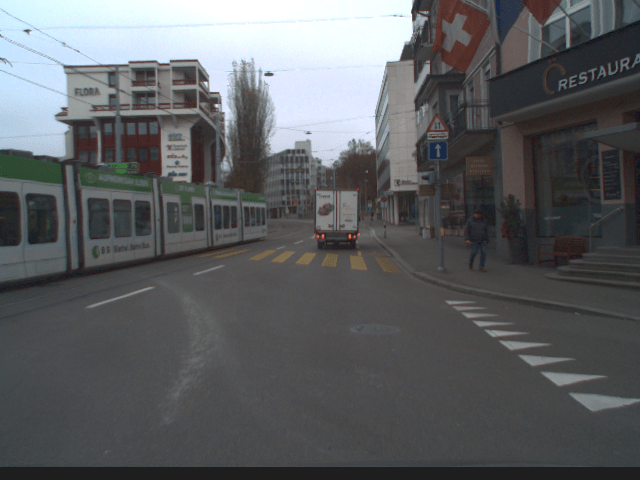}}\hfill
      \subfloat[\label{abl:b}]{\includegraphics[width=\cimwidth\textwidth,trim={465 205 130 205},clip]{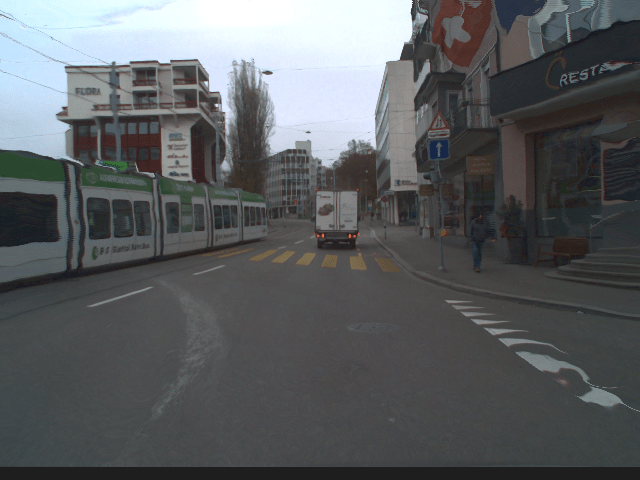}}\hfill
	   \subfloat[\label{abl:c}]{\includegraphics[width=\cimwidth\textwidth,trim={465 205 130 205},clip]{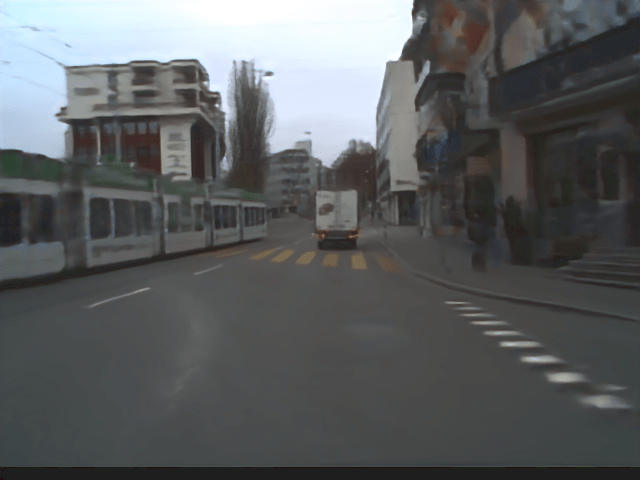}}\hfill
      \subfloat[\label{abl:d}]{\includegraphics[width=\cimwidth\textwidth,trim={465 205 130 205},clip]{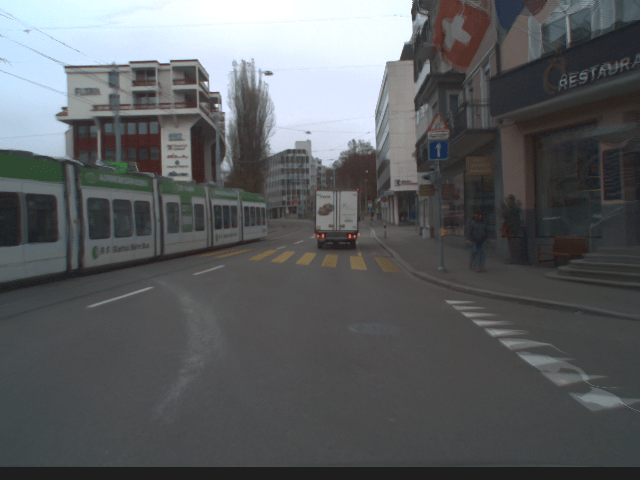}}\hfill
	   \subfloat[\label{abl:e}]{\includegraphics[width=\cimwidth\textwidth,trim={465 205 130 205},clip]{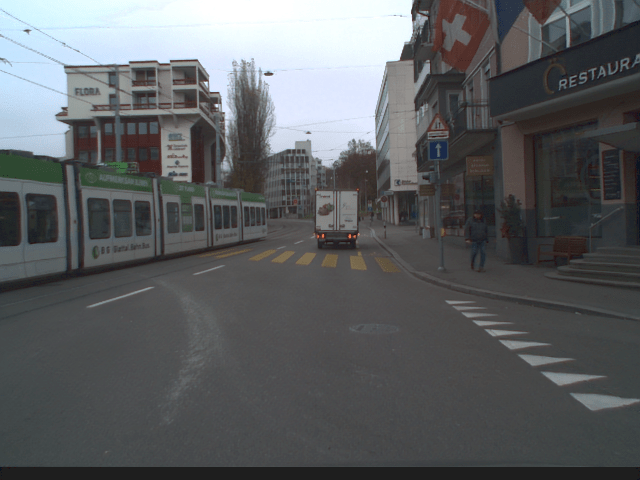}}\hfill
	   \subfloat[\label{abl:f}]{\includegraphics[width=\cimwidth\textwidth,trim={465 205 130 205},clip]{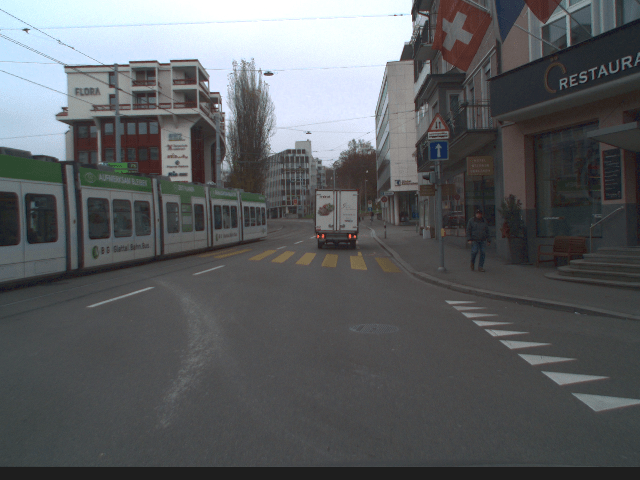}}\hfill
    \vspace{-10mm}
      \flushright
      \captionsetup[subfloat]{labelsep=none,format=plain,labelformat=empty}
      \begin{tikzpicture}[inner sep=0]
            \node [label={[label distance=0.85cm,text depth=0.5ex,rotate=90]right: \textcolor{black}{\scriptsize {2nd}}}] at (15,15) {};
            \end{tikzpicture}
      \subfloat[]{\includegraphics[width=\cimwidth\textwidth,trim={465 205 130 205},clip]{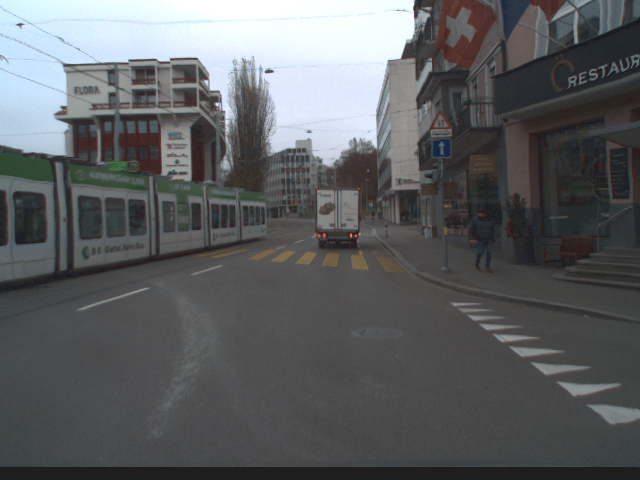}}\hfill
      \subfloat[]{\includegraphics[width=\cimwidth\textwidth,trim={465 205 130 205},clip]{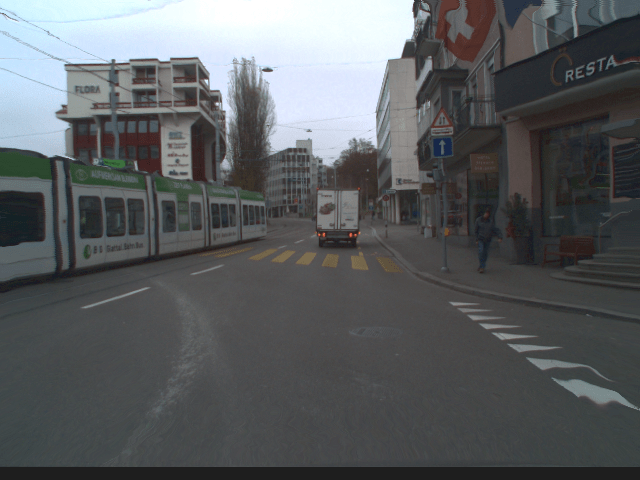}}\hfill
	   \subfloat[]{\includegraphics[width=\cimwidth\textwidth,trim={465 205 130 205},clip]{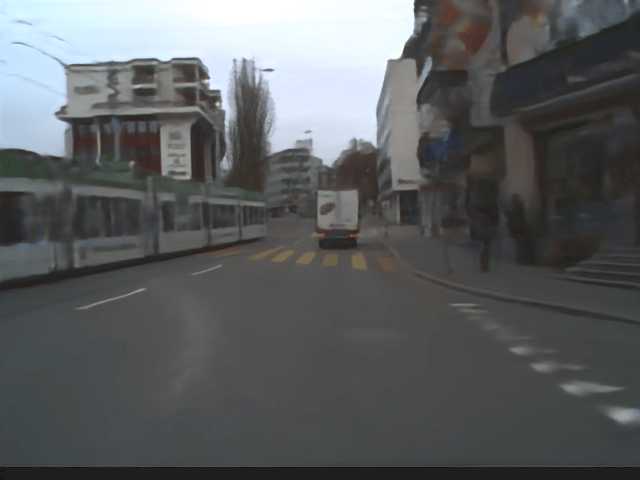}}\hfill
      \subfloat[]{\includegraphics[width=\cimwidth\textwidth,trim={465 205 130 205},clip]{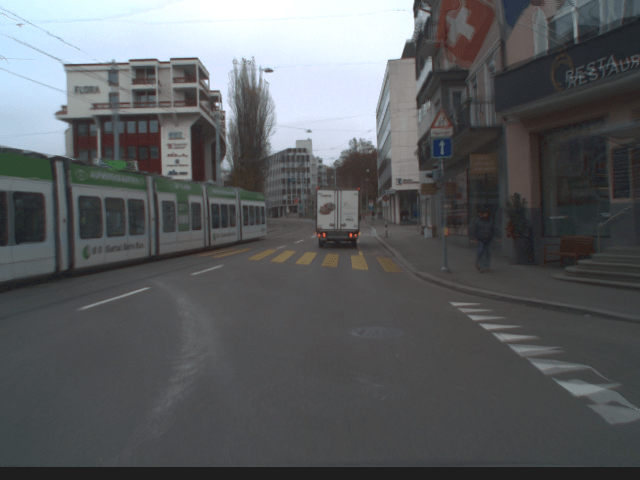}}\hfill
	   \subfloat[]{\includegraphics[width=\cimwidth\textwidth,trim={465 205 130 205},clip]{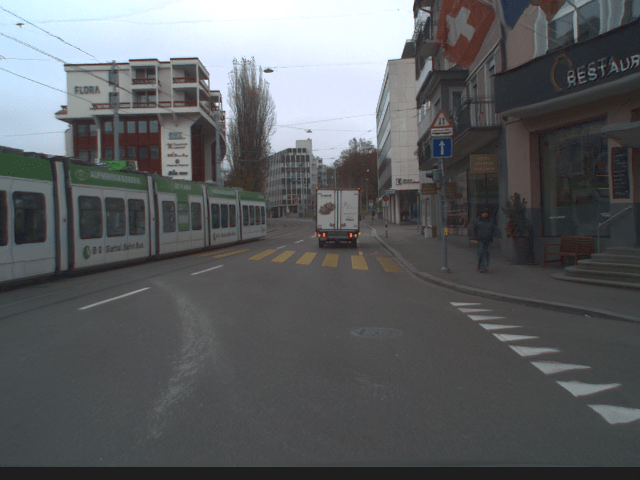}}\hfill
	   \subfloat[]{\includegraphics[width=\cimwidth\textwidth,trim={465 205 130 205},clip]{figs/ablation/86/_GT.png}}\hfill
    \vspace{-10mm}
      \flushright
      \captionsetup[subfloat]{labelsep=none,format=plain,labelformat=empty}
            \begin{tikzpicture}[inner sep=0]
            \node [label={[label distance=0.85cm,text depth=0.5ex,rotate=90]right: \textcolor{black}{\scriptsize {3rd}}}] at (15,15) {};
            \end{tikzpicture}
      \subfloat[(b) $W_{0 \rightarrow t}^{\Omega_f}$]{\includegraphics[width=\cimwidth\textwidth,trim={465 205 130 205},clip]{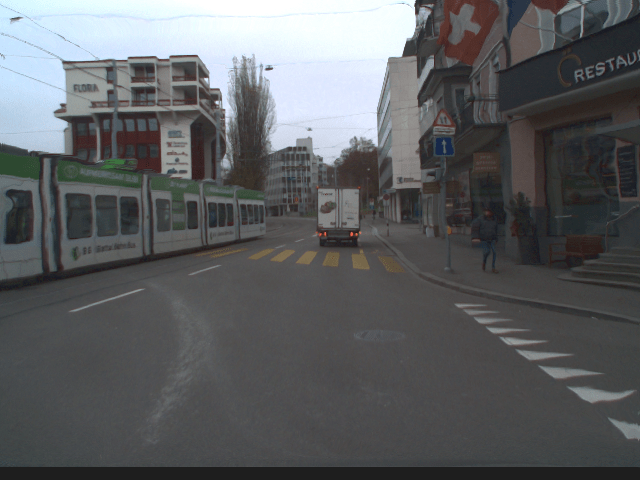}}\hfill
      \subfloat[(c) $W_{1 \rightarrow t}^{\Omega_f}$]{\includegraphics[width=\cimwidth\textwidth,trim={465 205 130 205},clip]{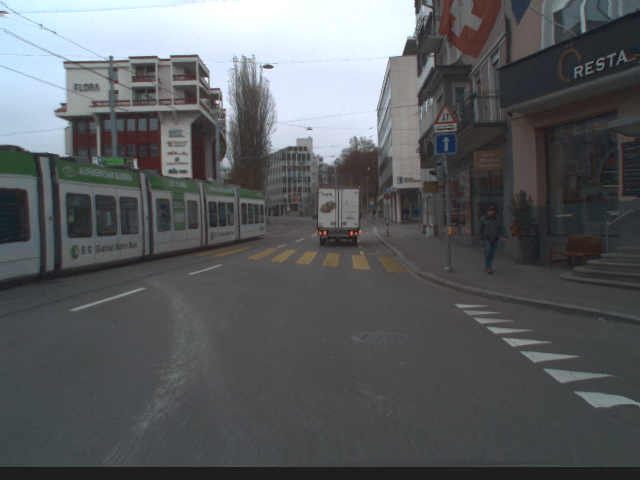}}\hfill
	   \subfloat[(d) $S_t^{\Omega_f}$]{\includegraphics[width=\cimwidth\textwidth,trim={465 205 130 205},clip]{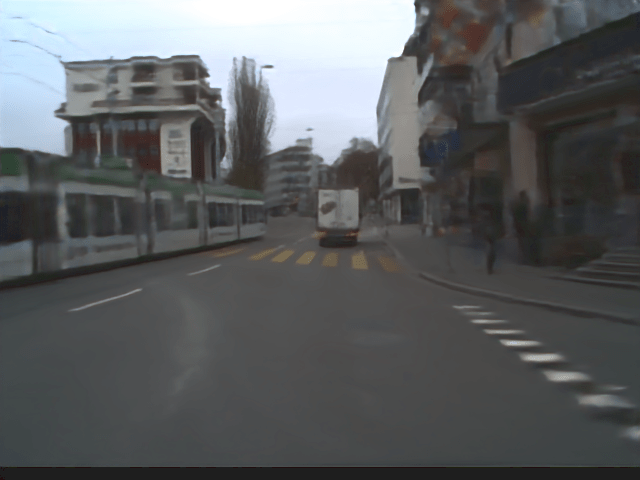}}\hfill
      \subfloat[(e) $\tilde{I}_t^{\Omega_f}$]{\includegraphics[width=\cimwidth\textwidth,trim={465 205 130 205},clip]{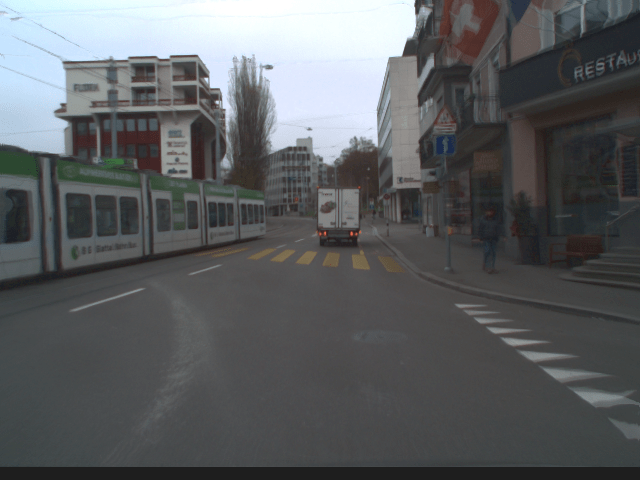}}\hfill
	   \subfloat[(f) $\hat{I}_t^{\Omega_f}$]{\includegraphics[width=\cimwidth\textwidth,trim={465 205 130 205},clip]{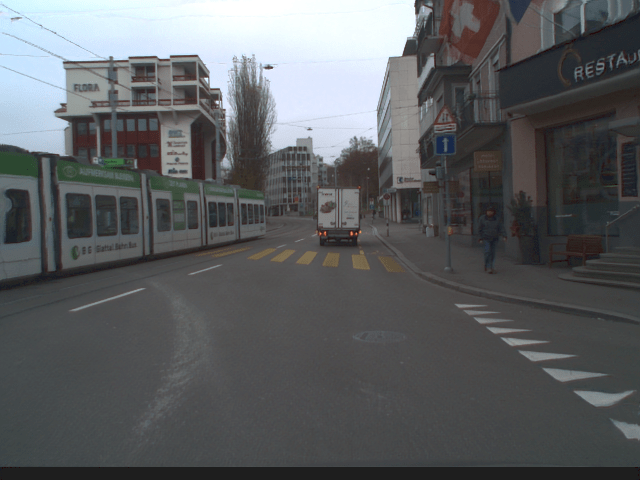}}\hfill
	   \subfloat[\vspace{-4mm}(g) GT]{\includegraphics[width=\cimwidth\textwidth,trim={465 205 130 205},clip]{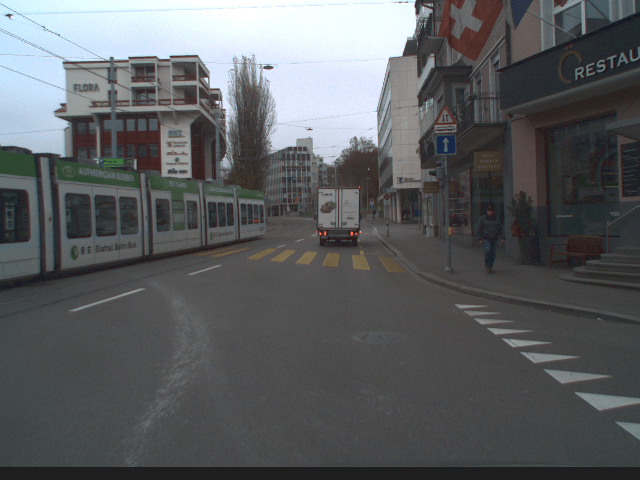}}
    \\
    \vspace{2mm}
    \end{minipage}

     \vspace{-2mm}
	\caption{Qualitative comparisons of the outputs of each subnetwork of the proposed \mynetwork\ on the DSEC dataset.}
	\label{fig-ablation}
 \vspace{-1mm}
\end{figure*}

\subsection{Ablation Study}\label{ablation}
In this section, we investigate the contribution of each subnetwork in our \mynetwork\ to the interpolation results. As illustrated in \cref{tab:ablation,fig-ablation}, we compare the flow-based results, \ie, $W_{0 \rightarrow t}^{\Omega_f}, W_{1 \rightarrow t}^{\Omega_f}$, the syn-based result $S_t^{\Omega_f}$, the fused result $\tilde{I}_t^{\Omega_f}$, and the final refined result $\hat{I}_t^{\Omega_f}$ on the DSEC, MVSEC, and \mydataset\ datasets.

As displayed in \cref{abl:a,abl:b}, the flow-based results bring artifacts and distortions when the target time is far from input keyframes,
while the syn-based results can roughly reconstruct inter-frame motion as shown in \cref{abl:c}.
However, due to the limited information in the warped event data, which only includes the surrounding information of the target frame, it is challenging to effectively incorporate the underlying temporal information and achieve a high level of detail in the reconstruction.

The results in \cref{tab:ablation} demonstrate that the utilization of attention-based fusion maps allows us to harness the advantages of both pathways and enhance the quality of the reconstruction. We observe an average improvement of 2.41 dB in terms of PSNR, and the visual quality is also enhanced as illustrated in \cref{abl:d}.
Furthermore, to improve the fused results by addressing artifacts and inconsistencies, we design RefineNet at the end, which computes the residual maps to capture the differences between the fused result and the ground truth. \cref{tab:ablation} shows an improvement of 0.69 dB in terms of PSNR and \cref{abl:e} achieves the best visualization quality.

\section{Conclusion}
In this paper, we tackle the stereo event-based video frame interpolation task with a novel network named \mynetwork.
Considering the problems of spatial misalignment and modality differences present in stereo camera setups, we design a core FAM, establishing spatial correspondence and achieving data fusion using features extracted from events and frames separately.
Additionally, we utilize attention maps to combine flow-based and syn-based results for better fusion. Our \mynetwork\ can generate high-quality intermediate frames and cross-modal disparities with spatially misaligned events and frames.
We also build a stereo visual acquisition system and collect a new \mydataset\ containing diverse complex scenes.
Extensive experiments demonstrate that the proposed \mynetwork\ establishes state-of-the-art performance on real-world stereo event-intensity datasets.

\bibliographystyle{IEEEtran}

\vfill

\end{document}